%% file: neurips_data_2022.tex
\documentclass{article}

\usepackage[final]{neurips_data_2022}

\usepackage[utf8]{inputenc} 
\usepackage[T1]{fontenc}    
\usepackage[numbers]{natbib}
\usepackage{hyperref}       
\usepackage{url}            
\usepackage{booktabs}       
\usepackage{amsfonts}       
\usepackage{nicefrac}       
\usepackage{microtype}      
\usepackage{xcolor}         
\usepackage[pdftex]{graphicx}
\usepackage{subfigure}
\usepackage{multirow}
\usepackage{enumitem}
\usepackage{algorithmic}
\usepackage{algorithm}
\usepackage{caption}
\usepackage{amsmath}
\usepackage{amssymb}
\usepackage{mathtools}
\usepackage{amsthm}
\usepackage[bb=boondox]{mathalfa}
\usepackage{wrapfig}
\usepackage[figuresright]{rotating}
\usepackage{chngpage}

\theoremstyle{plain}

\theoremstyle{definition}

\theoremstyle{remark}

\title{The Surprising Effectiveness of PPO in Cooperative Multi-Agent Games}

\author{Chao Yu$^{1\sharp}$\thanks{Equal Contribution. $^{\flat}$ Equal Advising. }, 
Akash Velu$^{2\natural*}$, 
Eugene Vinitsky$^{2\flat}$, 
Jiaxuan Gao$^{1}$, \\
\textbf{
Yu Wang$^{1\flat}$, 
Alexandre Bayen$^{2}$,  
Yi Wu$^{13\flat}$}
\\
$^1$ Tsinghua University $^2$ University of California, Berkeley $^3$ Shanghai Qi Zhi Institute\\
$^{\sharp}$\texttt{zoeyuchao@gmail.com}, $^{\natural}$\texttt{akashvelu@berkeley.edu} \\
}

\begin{document}

\maketitle

\vspace{-4mm}
\begin{abstract}
Proximal Policy Optimization (PPO) is a ubiquitous on-policy reinforcement learning algorithm but is significantly less utilized than off-policy learning algorithms in multi-agent settings. This is often due to the belief that PPO is significantly less sample efficient than off-policy methods in multi-agent systems. In this work, we carefully study the performance of PPO in cooperative multi-agent settings. We show that PPO-based multi-agent algorithms achieve surprisingly strong performance in four popular multi-agent testbeds: the particle-world environments, the StarCraft multi-agent challenge, Google Research Football, and the Hanabi challenge, with minimal hyperparameter tuning and without any domain-specific algorithmic modifications or architectures. Importantly, compared to competitive off-policy methods, PPO often achieves competitive or superior results in both final returns and sample efficiency. Finally, through ablation studies, we analyze implementation and hyperparameter factors that are critical to PPO's empirical performance, and give concrete practical suggestions regarding these factors. Our results show that when using these practices, simple PPO-based methods can be a strong baseline in cooperative multi-agent reinforcement learning. Source code is released at \url{https://github.com/marlbenchmark/on-policy}.
\end{abstract}

\vspace{-4mm}
\section{Introduction}
\vspace{-3mm}

Recent advances in reinforcement learning (RL) and multi-agent reinforcement learning (MARL) have led to a great deal of progress in creating artificial agents which can cooperate to solve tasks: DeepMind's AlphaStar surpassed professional-level performance in the StarCraft II~\cite{vinyals2019grandmaster}, OpenAI Five defeated the world-champion in Dota II~\cite{DBLP:journals/corr/abs-1912-06680}, and OpenAI demonstrated the emergence of human-like tool-use agent behaviors via multi-agent learning~\cite{baker2020emergent}. These notable successes were driven largely by on-policy RL algorithms such as IMPALA~\cite{espeholt2018impala} and PPO~\cite{PPO,DBLP:journals/corr/abs-1912-06680} which were often coupled with distributed training systems to utilize massive amounts of parallelism and compute. In the aforementioned works, tens of thousands of CPU cores and hundreds of GPUs were utilized to collect and train on an extraordinary volume of training samples. This is in contrast to recent academic progress and literature in MARL which has largely focused developing off-policy learning frameworks such as MADDPG~\cite{lowe2017multi} and value-decomposed Q-learning~\cite{sunehag2018value, pmlr-v80-rashid18a}; methods in these frameworks have yielded state-of-the-art results on a wide range of multi-agent benchmarks~\cite{wang2021qplex, wang2021rode}. 


In this work, we revisit the use of \emph{Proximal Policy Optimization} (PPO) -- an on-policy algorithm\footnote{Technically, PPO adopts off-policy corrections for sample-reuse. However, unlike off-policy methods, PPO does not utilize a replay buffer to train on samples collected throughout training.} popular in single-agent RL but under-utilized in recent MARL literature -- in multi-agent settings. We hypothesize that the relative lack of PPO in multi-agent settings can be attributed to two related factors: first, the belief that PPO is less sample-efficient than off-policy methods and is correspondingly less useful in resource-constrained settings, and second, the fact that common implementation and hyperparameter tuning practices when using PPO in single-agent settings often do not yield strong performance when transferred to multi-agent settings.

We conduct a comprehensive empirical study to examine the performance of PPO on four popular cooperative multi-agent benchmarks: the multi-agent particle world environments (MPE)~\citep{lowe2017multi}, the StarCraft multi-agent challenge (SMAC) ~\citep{starcraft}, Google Research Football (GRF)~\cite{kurach2020google} and the Hanabi challenge~\citep{bard2020hanabi}. We first show that when compared to off-policy baselines, PPO achieves strong task performance and competitive sample-efficiency. We then identify five implementation factors and hyperparameters which are particularly important for PPO's performance, offer concrete suggestions about these configuring factors, and provide intuition as to why these suggestions hold. 

Our aim in this work is \emph{not} to propose a novel MARL algorithm, but instead to empirically demonstrate that with simple modifications, PPO can achieve strong performance in a wide variety of cooperative multi-agent settings. We additionally believe that our suggestions will assist practitioners in achieving competitive results with PPO.

Our contributions are summarized as follows: 
\vspace{-\topsep}
\begin{itemize}
    \setlength{\parskip}{0pt} \setlength{\itemsep}{0pt plus 1pt}
    \item We demonstrate that PPO, without any domain-specific algorithmic changes or architectures and with minimal tuning, achieves final performances competitive to off-policy methods on four multi-agent cooperative benchmarks. 
    \item We demonstrate that PPO obtains these strong results while using a comparable number of samples to many off-policy methods. 
    \item We identify and analyze five implementation and hyperparameter factors that govern the practical performance of PPO in these settings, and offer concrete suggestions as to best practices regarding these factors. 
\end{itemize} 
\vspace{-\topsep}

\section{Related Works} \label{Sec: related} 
\vspace{-3mm}
MARL algorithms generally fall between two frameworks: centralized and decentralized learning. Centralized methods~\cite{claus1998dynamics} directly learn a single policy to produce the joint actions of all agents. In decentralized learning~\cite{littman1994markov}, each agent optimizes its reward independently; these methods can tackle general-sum games but may suffer from instability even in simple matrix games~\cite{foerster2017learning}. \emph{Centralized training and decentralized execution (CTDE)} algorithms fall in between these two frameworks. Several past CTDE methods~\cite{lowe2017multi, foerster2017counterfactual} adopt actor-critic structures and learn a centralized critic which takes global information as input. Value-decomposition (VD) methods are another class of CTDE algorithms which represent the joint Q-function as a function of agents' local Q-functions~\cite{sunehag2018value, pmlr-v80-rashid18a, son2019qtran}
and have established state of the art results in popular MARL benchmarks \cite{wang2021rode, wang2021qplex}.

In single-agent continuous control tasks~\cite{duan2016benchmarking}, advances in off-policy methods such as SAC~\cite{haarnoja2018soft} led to a consensus that despite their early success, policy gradient (PG) algorithms such as PPO are less sample efficient than off-policy methods.
Similar conclusions have been drawn in multi-agent domains: \cite{papoudakis2021benchmarking} report that multi-agent PG methods such as COMA are outperformed by MADDPG and QMix~\cite{pmlr-v80-rashid18a} by a clear margin in the particle-world environment~\cite{mordatch2017emergence} and the StarCraft multi-agent challenge~\cite{starcraft}.

The use of PPO in multi-agent domains is studied by several concurrent works. \cite{dewitt2020independent} empirically show that decentralized, independent PPO (IPPO) can achieve high success rates in several hard SMAC maps -- however, the reported IPPO results remain overall worse than QMix, and the study is limited to SMAC. \cite{papoudakis2021benchmarking} perform a broad benchmark of various MARL algorithms and note that PPO-based methods often perform competitively to other methods. Our work, on the other hand, focuses on PPO and analyzes its performance on a more comprehensive set of cooperative multi-agent benchmarks. We show PPO achieves strong results in the vast majority of tasks and also identify and analyze different implementation and hyperparameter factors of PPO which are influential to its performance multi-agent domains; to the best of our knowledge, these factors have not been studied to this extent in past work, particularly in multi-agent contexts.

Our empirical analysis of PPO's implementation and hyperparameter factors in multi-agent settings is similar to the studies of policy-gradient methods in single-agent RL~\cite{tucker2018mirage, Ilyas2020A, Engstrom2020Implementation, andrychowicz2021what}. We find several of these suggestions to be useful and include them in our implementation. In our analysis, we focus on factors that are either largely understudied in the existing literature or are completely unique to the multi-agent setting. 


 \section{PPO in Multi-Agent Settings} 
 \vspace{-2mm}
\subsection{Preliminaries}
We study decentralized partially observable Markov decision processes (DEC-POMDP)~\cite{oliehoek2016concise} with shared rewards. A DEC-POMDP is defined by $\langle \mathcal{S}, \mathcal{A}, O, R, P, n, \gamma \rangle$. $\mathcal{S}$ is the state space. $\mathcal{A}$ is the shared action space for  each agent $i$. $o_i=O(s;i)$ is the local observation for agent $i$ at global state $s$. $P(s'|s,A)$ denotes the transition probability from $s$ to $s'$ given the joint action $A=(a_1,\ldots,a_n)$ for all $n$ agents. $R(s,A)$ denotes the shared reward function. $\gamma$ is the discount factor.
Agents use a policy $\pi_{\theta}(a_i|o_i)$ parameterized by $\theta$ to produce an action $a_i$ from the local observation $o_i$, and 
jointly optimize the discounted accumulated reward $J(\theta)=\mathbb{E}_{A^t,s^t}\left[\sum_t \gamma^t R(s^t,A^t)\right]$ where $A^t=(a^t_1,\dots,a_n^t)$ is the joint action at time step $t$.

\subsection{MAPPO and IPPO} 

Our implementation of PPO in multi-agent settings closely resembles the structure of PPO in single-agent settings by learning a policy $\pi_{\theta}$ and a value function $V_{\phi}(s)$; these functions are represented as two separate neural networks. $V_{\phi}(s)$ is used for variance reduction and is only utilized during training; hence, it can take as input extra global information not present in the agent's local observation, allowing PPO in multi-agent domains to follow the CTDE structure. For clarity, we refer to PPO with centralized value function inputs as MAPPO (Multi-Agent PPO), and PPO with local inputs for both the policy and value function as IPPO (Independent PPO). We note that both MAPPO and IPPO operate in settings where agents share a common reward, as we focus only on cooperative settings. 

\subsection{Implementation Details}
\begin{enumerate}[nolistsep,leftmargin=*]
    \item[$\bullet$] \textbf{Parameter-Sharing}: In benchmark environments with homogeneous agents (i.e. agents have identical observation and action spaces), we utilize parameter sharing; past works have shown that this improves the efficiency of learning~\citep{christianos2021scaling,terry2021revisiting}, which is also consistent with our empirical findings. In these settings, agents share both the policy and value function parameters. A comparison of using parameter-sharing setting and learning separate parameters per agent can be found in Appendix \ref{app:sharing}. We remark that agents are homogeneous in all benchmarks except for the \emph{Comm} setting in the MPEs.
    \item[$\bullet$] \textbf{Common Implementation Practices:} We also adopt common practices in implementing PPO, including \emph{Generalized Advantage Estimation (GAE)} \cite{Schulmanetal_ICLR2016} with advantage normalization and value-clipping. A full description of hyperparameter search settings, training details, and implementation details are in Appendix \ref{app:training}. The source code for our implementation can be found in \url{https://github.com/marlbenchmark/on-policy}.

\end{enumerate}

\section{Main Results} \label{sec:results} 

\vspace{-1mm}
\subsection{Testbeds, Baselines, and Common Experimental Setup} \vspace{-1mm}
\textbf{Testbed Environments:} We evaluate the performance of MAPPO and IPPO on four cooperative benchmark -- the multi-agent particle-world environment (MPE), the StarCraft  multi-agent challenge (SMAC), the Hanabi challenge, and Google Research Football (GRF) -- and compare these methods' performance to popular off-policy algorithms which achieve state of the art results in each benchmark. Detailed descriptions of each testbed can be found in Appendix \ref{app:env}.

\textbf{Baselines:}
In each testbed, compare MAPPO and IPPO to a set of off-policy baselines, specifically:
\begin{enumerate}[nolistsep,leftmargin=*]
    \item[$\bullet$] \textbf{MPEs}: QMix~\cite{pmlr-v80-rashid18a} and MADDPG~\cite{lowe2017multi}.
    \item[$\bullet$] \textbf{SMAC}: QMix~\cite{pmlr-v80-rashid18a} and SOTA methods including QPlex~\cite{wang2021qplex}, CWQMix~\cite{DBLP:conf/nips/RashidFPW20}, AIQMix~\cite{DBLP:journals/corr/abs-2006-04222} and RODE~\cite{wang2021rode}.
    \item[$\bullet$] \textbf{GRF}: QMix~\cite{pmlr-v80-rashid18a} and SOTA methods including CDS~\citep{li2021celebrating} and TiKick~\citep{huang2021tikick}.
    \item[$\bullet$] \textbf{Hanabi}: SAD~\cite{hu2019simplified} and VDN~\cite{sunehag2018value}.
\end{enumerate}

\textbf{Common Experimental setup: } Here we give a brief description of the experimental setup common to all testbeds. Specific settings for each testbed are described later in Sec. 4.2-4.5.
\begin{enumerate}[nolistsep,leftmargin=*]
\item[$\bullet$] \textbf{Hyper-parameters Search}: For a fair comparison, we re-implement MADDPG and QMix
and tune each method using a grid-search over a set of hyper-parameters such as learning rate, target network update rate, and network architecture. We ensure that the size of this grid-search is equivalent to the size used to tune MAPPO and IPPO. We also test various relevant implementation tricks including value/reward normalization, hard and soft target network updates for Q-learning, and the input representation to the critic/mixer network.
\item[$\bullet$] \textbf{Training Compute:} Experiments are performed on a desktop machine with 256 GB RAM, one 64-core CPU, and one GeForce RTX 3090 GPU used for forward action computation and training updates.
\end{enumerate}

\textbf{Empirical Findings:} In the majority of environments, PPO achieves results better or comparable to the off-policy methods with \textbf{\emph{comparable sample efficiency}}. 


\subsection{MPE Testbed}
\textbf{Experimental Setting:} We consider the three cooperative tasks proposed in \cite{lowe2017multi}: the physical deception task (\emph{Spread}), the simple reference task (\emph{Reference}), and the cooperative communication task (\emph{Comm}). As the MPE environment does not provide a global input, we follow \cite{lowe2017multi} and concatenate all agents' local observations to form a global state which is utilized by MAPPO and the off-policy methods. Furthermore, \emph{Comm} is the only task without homogenous agents; hence, we do not utilize parameter sharing for this task. All results are averaged over ten seeds.

\begin{figure}[t!]
\vspace{-2mm}
	{\centering
        {
        \includegraphics[width=0.25\textwidth]{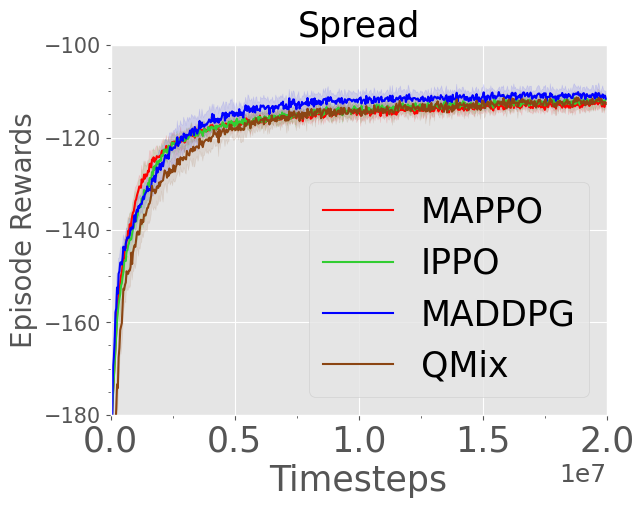}
        \includegraphics[width=0.25\textwidth]{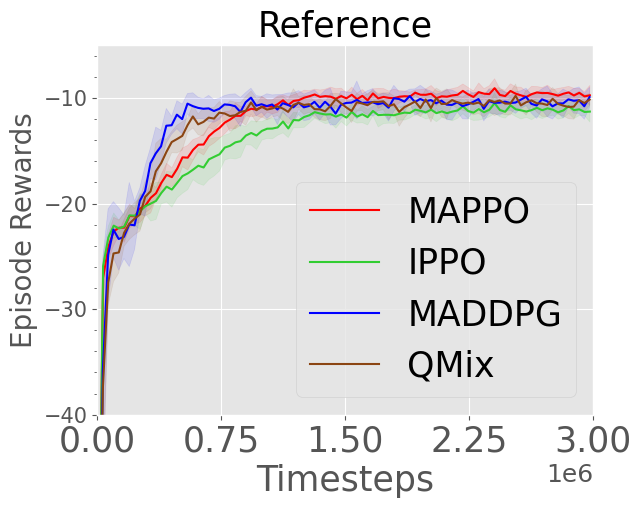}
        \includegraphics[width=0.25\textwidth]{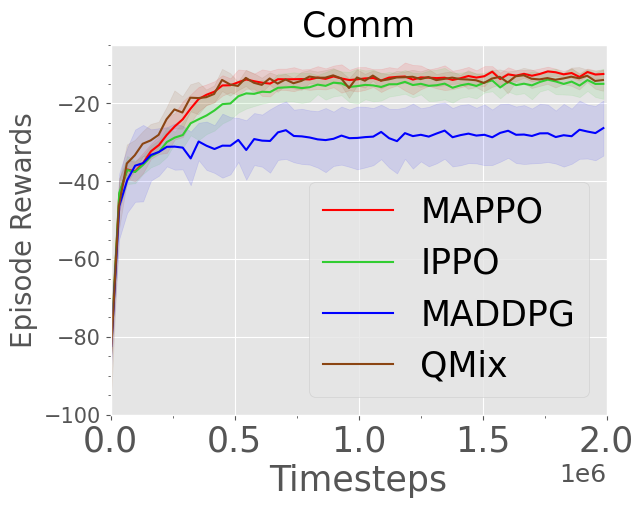}
        }
    }
    \vspace{-1mm}
	\centering \caption{Performance of different algorithms in the MPEs.}
	\vspace{-4mm}
\label{fig:MPE-all}
\end{figure}

\textbf{Experimental Results:} The performance of each algorithm at convergence is shown in Fig.~\ref{fig:MPE-all}. MAPPO achieves performance comparable and even superior to the off-policy baselines; we particularly see that MAPPO performs very similarly to QMix on all tasks and exceeds the performance of MADDPG in the \emph{Comm} task, all while using a comparable number of environment steps. Despite not utilizing global information, IPPO also achieves similar or superior performance to centralized off-policy methods. Compared to MAPPO, IPPO converges to \emph{slightly} lower final returns in several environments (\emph{Comm} and \emph{Reference}). 


\begin{table*}[t!]
\centering
\addtolength{\tabcolsep}{-4pt}    
\begin{tabular}{ccccc|ccc}
\toprule 
Map              & MAPPO\scriptsize{(FP)}  & MAPPO\scriptsize{(AS)}  & IPPO       & QMix       & RODE*        & MAPPO*\scriptsize{(FP)}  & MAPPO*\scriptsize{(AS)}  \\
\midrule
\scriptsize{2m vs\_1z}        & \textbf{100.0}\scriptsize{(0.0)} & \textbf{100.0\scriptsize{(0.0)}} & \textbf{100.0}\scriptsize{(0.0)} & \textbf{95.3}\scriptsize{(5.2)}  & /           & \emph{\underline{100.0}}\scriptsize{(0.0)} & \emph{\underline{100.0}}\scriptsize{(0.0)} \\
\scriptsize{3m}               & \textbf{100.0}\scriptsize{(0.0)} & \textbf{100.0}\scriptsize{(1.5)} & \textbf{100.0}\scriptsize{(0.0)} & 96.9\scriptsize{(1.3)}  & /           & \emph{\underline{100.0}}\scriptsize{(0.0)} & \emph{\underline{100.0}}\scriptsize{(1.5)} \\
\scriptsize{2svs1sc}          & \textbf{100.0}\scriptsize{(0.0)} & \textbf{100.0}\scriptsize{(0.0)} & \textbf{100.0}\scriptsize{(1.5)}   & 96.9\scriptsize{(2.9)}  & \emph{\underline{100.0}}\scriptsize{(0.0)}  & \emph{\underline{100.0}}\scriptsize{(0.0)} & \emph{\underline{100.0}}\scriptsize{(0.0)} \\
\scriptsize{2s3z}             & \textbf{100.0}\scriptsize{(0.7)} & \textbf{100.0}\scriptsize{(1.5)} & \textbf{100.0}\scriptsize{(0.0)} & 95.3\scriptsize{(2.5)}  & \emph{\underline{100.0}}\scriptsize{(0.0)}  & 96.9\scriptsize{(1.5)}  & 96.9\scriptsize{(1.5)}  \\
\scriptsize{3svs3z}           & \textbf{100.0}\scriptsize{(0.0)} & \textbf{100.0}\scriptsize{(0.0)} & \textbf{100.0}\scriptsize{(0.0)}   & \textbf{96.9}\scriptsize{(12.5)} & /           & \emph{\underline{100.0}}\scriptsize{(0.0)} & \emph{\underline{100.0}}\scriptsize{(0.0)} \\
\scriptsize{3svs4z}          & \textbf{100.0}\scriptsize{(1.3)} & \textbf{98.4}\scriptsize{(1.6)}  & \textbf{99.2}\scriptsize{(1.5)}  & \textbf{97.7}\scriptsize{(1.7)}  & /           & \emph{\underline{100.0}}\scriptsize{(2.1)} & \emph{\underline{100.0}}\scriptsize{(1.5)} \\
\scriptsize{so many baneling} & \textbf{100.0}\scriptsize{(0.0)} & \textbf{100.0}\scriptsize{(0.7)} & \textbf{100.0}\scriptsize{(1.5)} & 96.9\scriptsize{(2.3)}  & /           & \emph{\underline{100.0}}\scriptsize{(1.5)} & 96.9\scriptsize{(1.5)}  \\
\scriptsize{8m}               & \textbf{100.0}\scriptsize{(0.0)} & \textbf{100.0}\scriptsize{(0.0)} & \textbf{100.0}\scriptsize{(0.7)} & 97.7\scriptsize{(1.9)}  & /           & \emph{\underline{100.0}}\scriptsize{(0.0)} & \emph{\underline{100.0}}\scriptsize{(0.0)} \\
\scriptsize{MMM}              & \textbf{96.9}\scriptsize{(0.6)}  & 93.8\scriptsize{(1.5)}  & \textbf{96.9}\scriptsize{(0.0)}  & \textbf{95.3}\scriptsize{(2.5)}  & /           & \emph{\underline{93.8}}\scriptsize{(2.6)}  & \emph{\underline{96.9}}\scriptsize{(1.5)}  \\
\scriptsize{1c3s5z}           & \textbf{100.0}\scriptsize{(0.0)} & 96.9\scriptsize{(2.6)}  & \textbf{100.0}\scriptsize{(0.0)} & 96.1\scriptsize{(1.7)}  & 
\emph{\underline{100.0}}\scriptsize{(0.0)}  & \emph{\underline{100.0}}\scriptsize{(0.0)} & 96.9\scriptsize{(2.6)}  \\
\scriptsize{bane vs bane}     & \textbf{100.0}\scriptsize{(0.0)} & \textbf{100.0}\scriptsize{(0.0)} & \textbf{100.0}\scriptsize{(0.0)} & \textbf{100.0}\scriptsize{(0.0)} & \emph{\underline{100.0}}\scriptsize{(46.4)} & \emph{\underline{100.0}}\scriptsize{(0.0)} & \emph{\underline{100.0}}\scriptsize{(0.0)} \\
\scriptsize{3svs5z}           & \textbf{100.0}\scriptsize{(0.6)} & \textbf{99.2}\scriptsize{(1.4)}  & \textbf{100.0}\scriptsize{(0.0)} & \textbf{98.4}\scriptsize{(2.4)}  & 78.9\scriptsize{(4.2)}   & \emph{\underline{98.4}}\scriptsize{(5.5)}  & \emph{\underline{100.0}}\scriptsize{(1.2)} \\
\scriptsize{2cvs64zg}         & \textbf{100.0}\scriptsize{(0.0)} & \textbf{100.0}\scriptsize{(0.0)} & 98.4\scriptsize{(1.3)}  & 92.2\scriptsize{(4.0)}  & \emph{\underline{100.0}}\scriptsize{(0.0)}  & \emph{\underline{96.9}}\scriptsize{(3.1)}  & 95.3\scriptsize{(3.5)}  \\
\scriptsize{8mvs9m}           & \textbf{96.9}\scriptsize{(0.6)} & \textbf{96.9}\scriptsize{(0.6)}  & \textbf{96.9}\scriptsize{(0.7)}  & 92.2\scriptsize{(2.0)}  & /           & \emph{\underline{84.4}}\scriptsize{(5.1)}  & \emph{\underline{87.5}}\scriptsize{(2.1)}  \\
\scriptsize{25m}              & \textbf{100.0}\scriptsize{(1.5)} & \textbf{100.0}\scriptsize{(4.0)} & \textbf{100.0}\scriptsize{(0.0)} & 85.9\scriptsize{(7.1)}  & /           & \emph{\underline{96.9}}\scriptsize{(3.1)}  & \emph{\underline{93.8}}\scriptsize{(2.9)}  \\
\scriptsize{5mvs6m}           & \textbf{89.1}\scriptsize{(2.5)}  & \textbf{88.3}\scriptsize{(1.2)}  & \textbf{87.5}\scriptsize{(2.3)}  & 75.8\scriptsize{(3.7)}  & \emph{\underline{71.1}}\scriptsize{(9.2)}   & \emph{\underline{65.6}}\scriptsize{(14.1)} & \emph{\underline{68.8}}\scriptsize{(8.2)}  \\
\scriptsize{3s5z}             & \textbf{96.9}\scriptsize{(0.7)}  & \textbf{96.9}\scriptsize{(1.9)}  & \textbf{96.9}\scriptsize{(1.5)}  & 88.3\scriptsize{(2.9)}  & \emph{\underline{93.8}}\scriptsize{(2.0)}   & 71.9\scriptsize{(11.8)} & 53.1\scriptsize{(15.4)} \\
\scriptsize{10mvs11m}         & \textbf{96.9}\scriptsize{(4.8)}  & \textbf{96.9}\scriptsize{(1.2)}  & \textbf{93.0}\scriptsize{(7.4)}  & \textbf{95.3}\scriptsize{(1.0)}  & \emph{\underline{95.3}}\scriptsize{(2.2)}   & 81.2\scriptsize{(8.3)}  & \emph{\underline{89.1}}\scriptsize{(5.5)}  \\
\scriptsize{MMM2}             & \textbf{90.6}\scriptsize{(2.8)}  & \textbf{87.5}\scriptsize{(5.1)}  & \textbf{86.7}\scriptsize{(7.3)}  & \textbf{87.5}\scriptsize{(2.6)}  & \emph{\underline{89.8}}\scriptsize{(6.7)}   & 51.6\scriptsize{(21.9)} & 28.1\scriptsize{(29.6)} \\
\scriptsize{3s5zvs3s6z}       & \textbf{84.4}\scriptsize{(34.0)} & 63.3\scriptsize{(19.2)} & \textbf{82.8}\scriptsize{(19.1)} & \textbf{82.8}\scriptsize{(5.3)}  & \emph{\underline{96.8}}\scriptsize{(25.11)} & \emph{\underline{75.0}}\scriptsize{(36.3)} & 18.8\scriptsize{(37.4)} \\
\scriptsize{27mvs30m}         & \textbf{93.8}\scriptsize{(2.4)}  & 85.9\scriptsize{(3.8)}  & 69.5\scriptsize{(11.8)} & 39.1\scriptsize{(9.8)}  & \emph{\underline{96.8}}\scriptsize{(1.5)}   & \emph{\underline{93.8}}\scriptsize{(3.8)}  & \emph{\underline{89.1}}\scriptsize{(6.5)}  \\
\scriptsize{6hvs8z}           & \textbf{88.3}\scriptsize{(3.7)}  & \textbf{85.9}\textbf{}\scriptsize{(30.9)} & \textbf{84.4}\scriptsize{(33.3)} & 9.4\scriptsize{(2.0)}   & \emph{\underline{78.1}}\scriptsize{(37.0)}  & \emph{\underline{78.1}}\scriptsize{(5.6)}  & \emph{\underline{81.2}}\scriptsize{(31.8)} \\
\scriptsize{corridor}         & \textbf{100.0}\scriptsize{(1.2)} & \textbf{98.4}\scriptsize{(0.8)}  & \textbf{98.4}\scriptsize{(3.1)}  & 84.4\scriptsize{(2.5)}  & \emph{\underline{65.6}}\scriptsize{(32.1)}  & \emph{\underline{93.8}}\scriptsize{(3.5)}  & \emph{\underline{93.8}}\scriptsize{(2.8)}  \\
\bottomrule
\end{tabular}
\centering
\caption{Median evaluation win rate and standard deviation on all the SMAC maps for different methods, Columns with ``*'' display results using the same number of timesteps as RODE. We bold all values within 1 standard deviation of the maximum and among the ``*'' columns, we denote all values within 1 standard deviation of the maximum with underlined italics. AS next to MAPPO indicates an agent-specific centralized input to the value function; FP indicates a similar agent-specific centralized input, but with redundant information removed.  
} 
\label{tab:SMAC-results}

\end{table*}


\subsection{SMAC Testbed} 

\textbf{Experimental Setting:} We evaluate MAPPO with two different centralized value function inputs -- labeled \emph{AS} and \emph{FP} -- that combines agent-agnostic global information with agent-specific local information. These inputs are described fully in Section \ref{Sec: Factors}. All off-policy baselines utilize both the agent-agnostic global state and agent-specific local observations as input. Specifically, for agent $i$, the local Q-network (which computes actions at execution) takes in only the local agent-specific observation $o_i$ as input while the global mixer network takes in the agent-agnostic global state $s$ as input. For each random seed, we follow the evaluation metric proposed in \cite{wang2021rode}: we compute the win rate over 32 evaluation games after each training iteration and take the median of the final ten evaluation win-rates as the performance for each seed. 

\textbf{Experimental Results:} We report the median win rates over six seeds in Table \ref{tab:SMAC-results}, which compares the PPO-based methods to QMix and RODE. Full results are deferred to Table 2 and Table 3 in Appendix. MAPPO, IPPO, and QMix are trained until convergence or reaching 10M environment steps. Results for RODE are obtained using the statistics from \cite{wang2021rode}. We observe that IPPO and MAPPO with both the \emph{AS} and \emph{FP} inputs achieve strong performance in the vast majority of SMAC maps. In particular, MAPPO and IPPO perform at least as well as QMix in most maps despite using the same number of samples. Comparing different value functions inputs, we observe that the performance of IPPO and MAPPO is highly similar, with the methods performing strongly in all but one map each. We also observe that MAPPO achieves performance comparable or superior to RODE's in 10 of 14 maps while using the same number of training samples. With more samples, the performance of MAPPO and IPPO continue to improve and ultimately match or exceed RODE's performance in nearly every map. As shown in Appendix \ref{app:SMAC-allresults}, 
MAPPO and IPPO perform comparably or superior to other other off-policy methods such as QPlex, CWQMix, and AIQMix in terms of both final performance and sample-efficiency.

Overall, MAPPO's effectiveness in nearly every SMAC map suggests that simple PPO-based algorithms can be strong baselines in challenging MARL problems. 

\subsection{Google Football Testbed}
\textbf{Experimental Setting:} We evaluate MAPPO in several GRF academy scenarios, namely 3v.1, counterattack (CA) easy and hard, corner, pass-shoot (PS), and run-pass-shoot (RPS). In these scenarios, a team of agents attempts to score a goal against scripted opponent player(s). As the agents' local observations contain a full description of the environment state, there is no distinction between MAPPO and IPPO; for consistency, we label the results with PPO in Table \ref{tab:football} as ``MAPPO''.  We utilize GRF's dense-reward setting in which all agents share a single reward which is the sum of individual agents' dense rewards. We compute the success rate over 100 rollouts of the game and report the average success rate over the last 10 evaluations, averaged over 6 seeds.

\textbf{Experimental Results:} We compare MAPPO with QMix and several SOTA methods, including CDS, a method that augments the environment reward with an intrinsic reward, and TiKick, an algorithm which combines online RL fine-tuning and large-scale offline pre-training. All methods except TiKick are trained for 25M environment steps in all scenarios with the exception of CA (hard) and Corner, in which methods are trained for 50M environment steps. 

We generally observe in Table \ref{tab:football} that MAPPO achieves comparable or superior performance to other off-policy methods in \emph{all} settings, despite not utilizing an intrinsic reward as is done in CDS. Comparing MAPPO to QMix, we observe that MAPPO clearly outperforms QMix in each scenario, again while using the \emph{same number of training samples}. MAPPO additionally outperforms TiKick on 4/5 scenarios, despite the fact that TiKick performs pre-training on a set of human expert data.

\begin{table}[!t]
\centering
\begin{tabular}{cccc|c}
\toprule
Scen. &       MAPPO &        QMix &         CDS &      TiKick \\
\midrule
          3v.1 & \textbf{88.03}\scriptsize{(1.06)} &  8.12\scriptsize{(2.83)} & 76.60\scriptsize{(3.27)} & 76.88\scriptsize{(3.15)} \\
     CA\scriptsize{(easy)} & \textbf{87.76}\scriptsize{(1.34)} & 15.98\scriptsize{(2.85)} & 63.28\scriptsize{(4.89)} &           / \\
     CA\scriptsize{(hard)} & \textbf{77.38}\scriptsize{(4.81)} &  3.22\scriptsize{(1.60)} & 58.35\scriptsize{(5.56)} & 73.09\scriptsize{(2.08)} \\
      Corner & \textbf{65.53}\scriptsize{(2.19)} & 16.10\scriptsize{(3.00)} & 3.80\scriptsize{(0.54)} & 33.00\scriptsize{(3.01)} \\
          PS & \textbf{94.92}\scriptsize{(0.68)} &  8.05\scriptsize{(3.66)} & \textbf{94.15}\scriptsize{(2.54)} &           / \\
          RPS & \textbf{76.83}\scriptsize{(1.81)} &  8.08\scriptsize{(4.71)} & 62.38\scriptsize{(4.56)} & 79.12\scriptsize{(2.06)} \\
\bottomrule
\end{tabular}
\vspace{1mm}
\caption{Average evaluation success rate and standard deviation (over six seeds) on GRF scenarios for different methods. All values within 1 standard deviation of the maximum success rate are marked in bold. We separate TiKick from the other methods as it uses pretrained models and thus does not constitute a direct comparison.}
\vspace{-3mm}
\label{tab:football}
\end{table}


\subsection{Hanabi Testbed} 
\textbf{Experimental Setting:} We evaluate MAPPO and IPPO in the full-scale Hanabi game with varying numbers of players (2-5 players). We compare MAPPO and IPPO to strong off-policy methods, namely Value Decomposition Networks (VDN) and Simplified Action Decoder (SAD), a Q-learning variant that has been successful in Hanabi. All methods do not utilize auxiliary tasks. Because each agent's local observation does not contain information about the agent's own cards\footnote{The local observations in Hanabi contain information about the other agent's cards and game state.}, MAPPO utilizes a global-state that adds the agent's own cards to the local observation as input to its value function. VDN agents take only the local observations as input. SAD agents take as input not only the local observation provided by the environment, but also the greedy actions of other players in the past time steps (which is not used by MAPPO and IPPO). Due to algorithmic restrictions, no additional global information is utilized by SAD and VDN during centralized training. We follow \citep{hu2019simplified} and report the average returns across at-least 3 random seeds as well as the best score achieved by any seed. The returns are averaged over 10k games.

\begin{table}[!t]
\centering
\begin{tabular}{cccccc} 
\toprule
\# Players         &  Metric    & MAPPO   & IPPO    & SAD         & VDN          \\
\midrule
\multirow{2}{*}{2} & Avg. & 23.89\scriptsize{(0.02)} & \textbf{24.00}\scriptsize{(0.02)} & 23.87\scriptsize{(0.03)} & 23.83\scriptsize{(0.03)}  \\
                   & Best & \textbf{24.23}\scriptsize{(0.01)} & 24.19\scriptsize{(0.02)} & 24.01\scriptsize{(0.01)} & 23.96\scriptsize{(0.01)}  \\
                    \hline
\multirow{2}{*}{3} & Avg. &   \textbf{23.77}\scriptsize{(0.20)}         & 23.25\scriptsize{(0.33)} & 23.69\scriptsize{(0.05)} & 23.71\scriptsize{(0.06)}  \\
                   & Best &   \textbf{24.01}\scriptsize{(0.01)}          & 23.87\scriptsize{(0.03)} & 23.93\scriptsize{(0.01)} & 23.99\scriptsize{(0.01)}  \\
                   \hline
\multirow{2}{*}{4} & Avg. &  \textbf{23.57}\scriptsize{(0.13)}           & 22.52\scriptsize{(0.37)} & 23.27\scriptsize{(0.26)} & 23.03\scriptsize{(0.15)}  \\
                   & Best &   23.71\scriptsize{(0.01)}          & 23.06\scriptsize{(0.03)} & \textbf{23.81}\scriptsize{(0.01)}  & 23.79\scriptsize{(0.00)}  \\
                   \hline
\multirow{2}{*}{5} & Avg. &   \textbf{23.04}\scriptsize{(0.10)}          & 20.75\scriptsize{(0.56)} & 22.06\scriptsize{(0.23)} & 21.28\scriptsize{(0.12)}  \\
                   & Best & \textbf{23.16}\scriptsize{(0.01)}            & 22.54\scriptsize{(0.02)} & 23.01\scriptsize{(0.01)} & 21.80\scriptsize{(0.01)}  \\
\bottomrule
\end{tabular}
\vspace{1mm}
\caption{Best and Average evaluation scores of MAPPO, IPPO, SAD, and VDN on Hanabi-Full. Results are reported over at-least 3 seeds.}
\label{tab:Hanabi-results}
\vspace{-6mm}
\end{table}

\textbf{Experimental Results:} The reported results for SAD and VDN are obtained from ~\cite{hu2019simplified}. All methods are trained for at-most 10B environment steps. As demonstrated in Table~\ref{tab:Hanabi-results}, MAPPO is able to produce results comparable or superior to the best and average returns achieved by SAD and VDN in nearly every setting, while utilizing the same number of environment steps. This demonstrates that even in environments such as Hanabi which require reasoning over other players' intents based on their actions, MAPPO can achieve strong performance, despite not explicitly modeling this intent.

IPPO's performance is comparable with MAPPO's in the 2-agent setting. However, as the agent number grows, MAPPO shows a clear margin of improvement over both IPPO and off-policy methods, which suggests that a centralized critic input can be crucial.


\section{Factors Influential to PPO's Performance} \label{Sec: Factors} 
\vspace{-2mm}
In this section, we analyze five factors that we find are especially influential to MAPPO's performance: value normalization,  value function inputs, training data usage, policy/value clipping, and batch size. We find that these factors exhibit clear trends in terms of performance; using these trends, we give best-practice suggestions for each factor. We study each factor in a set of appropriate representative environments.
All experiments are performed using MAPPO (i.e., PPO with centralized value functions) for consistency. Additional results can be found in Appendix \ref{app:ablation-studies}.

\subsection{Value Normalization}
\begin{figure*}[ht]
	\centering
    \includegraphics[width=1.0\linewidth]{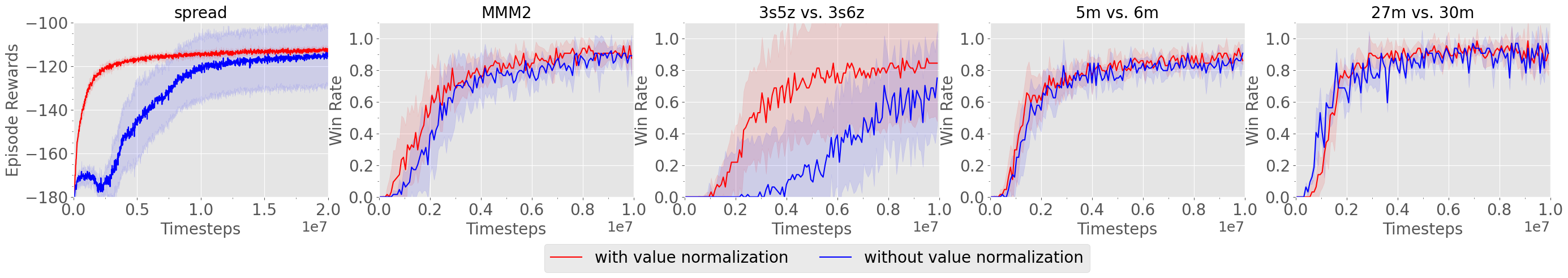}
    \vspace{-4mm}
	\centering \caption{Impact of value normalization on MAPPO's performance in SMAC and MPE.}
	\vspace{-3mm}
\label{fig:Abaltion-popart}
\end{figure*}

Through the training process of MAPPO, value targets can drastically change due to differences in the realized returns, leading to instability in value learning. To mitigate this issue, we standardize the targets of the value function by using running estimates of the average and standard deviation of the value targets. Concretely, during value learning, the value network regresses to normalized target values. When computing the GAE, we use the running average to denormalize the output of the value network so that the value outputs are properly scaled. We find that using value normalization never hurts training and often  improves the final performance of MAPPO significantly.

\textbf{Empirical Analysis: } We study the impact of value-normalization in the MPE \emph{spread} environment and several SMAC environments - results are shown in Fig. \ref{fig:Abaltion-popart}. In \emph{Spread}, where the episode returns range from below -200 to 0, value normalization is critical to strong performance. Value normalization also has positive impacts on several SMAC maps, either by improving final performance or by reducing the training variance.

\textbf{Suggestion 1:} Utilize value normalization to stabilize value learning. 

\subsection{Input Representation to Value Function}\label{sec:value_inputs} 
\begin{figure*}[ht]
	\centering
    \includegraphics[width=1.0\linewidth]{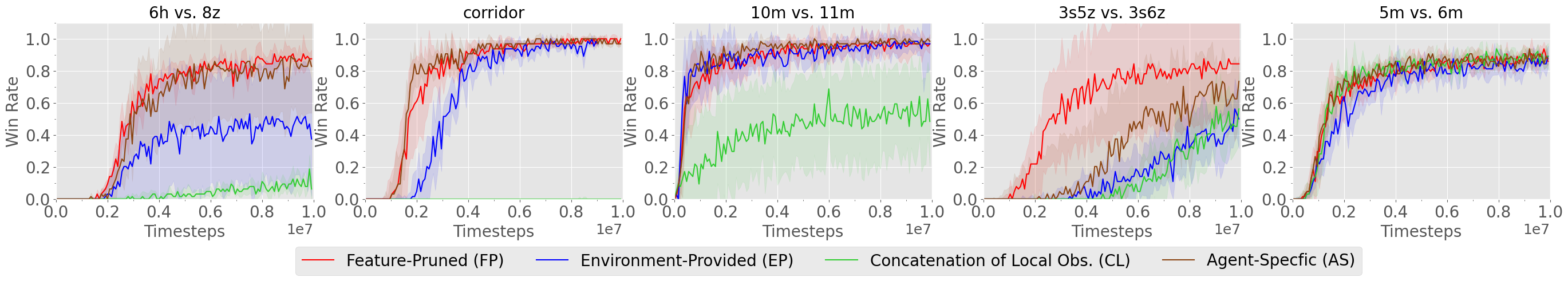}
    \vspace{-4mm}
	\centering \caption{Effect of different value function input representations (described in Fig. \ref{fig:Global-States}).} 
	\vspace{-2mm}
\label{fig:Abaltion-state}
\end{figure*}

The fundamental difference between many multi-agent CTDE PG algorithms and fully decentralized PG methods is the input to the value network. Therefore, the representation of the value input becomes an important aspect of the overall algorithm.  The assumption behind using centralized value functions is that observing the full global state can make value learning easier. An accurate value function further improves policy learning through variance reduction.

\begin{figure}[!t]
	\centering
    \vspace{-7mm}
    \includegraphics[width=0.70\linewidth]{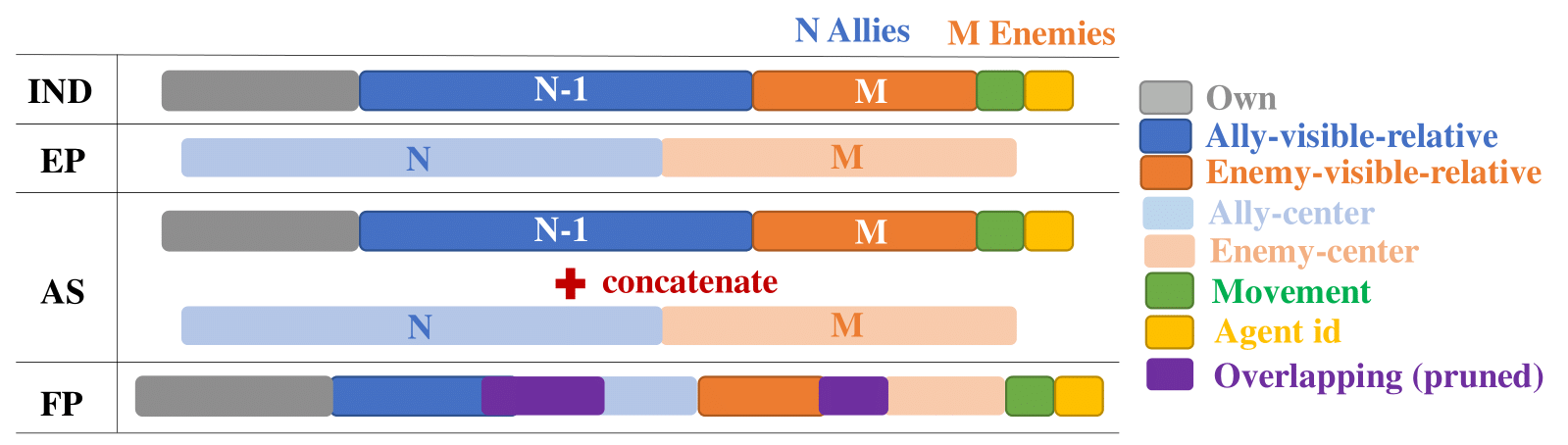}
	\centering \caption{Different value function inputs with example features contained in each state (SMAC-specific). $\emph{IND}$ refers to using decentralized inputs (agents' local observations), \emph{EP} refers to the environment provided global state, \emph{AS} is an agent-specific global state which concatenates \emph{EP} and \emph{IND}, and \emph{FP} is an agent-specific global state which prunes overlapping features from \emph{AS}. \emph{EP} omits important local data such as agent ID and available actions.}
	\vspace{-3mm}
\label{fig:Global-States}
\end{figure}

Past works have typically used two forms of global states. \cite{lowe2017multi} use a \textbf{concatenation of local observations (CL)} global state which is formed by concatenating all local agent observations. While it can be used in most environments, the \emph{CL} state dimensionality grows with the number of agents and can omit important global information which is unobserved by all agents; these factors can make value learning difficult. Other works, particularly those studying SMAC, utilize an \textbf{Environment-Provided global state (EP)} which contains general global information about the environment state \cite{foerster2017counterfactual}. However, the \emph{EP} state typically contains information common to all agents and can omit important local agent-specific information. This is true in SMAC, as shown in Fig. \ref{fig:Global-States}.

To address the weaknesses of the \emph{CL} and \emph{EP} states, we allow the value function to leverage both global and local information by forming an \textbf{Agent-Specific Global State (AS)} which creates a global state for agent $i$ by concatenating the \emph{EP} state and $o_i$, the local observation for agent $i$. This provides the value function with a more comprehensive description of the environment state. However, if there is overlap in information between $o_i$ and the $\emph{EP}$ global state, then the \emph{AS} state will have redundant information which unnecessarily increases the input dimensionality to the value function. As shown in Fig. \ref{fig:Global-States}, this is the case in SMAC. To examine the impact of this increased dimensionality, we  create a \textbf{Featured-Pruned Agent-Specific Global State (FP)} by removing repeated features in the \emph{AS} state.

\textbf{Emperical Analysis:} We study the impact of these different value function inputs in SMAC, which is the only considered benchmark that provides different options for centralized value function inputs. The results in  Fig.~\ref{fig:Abaltion-state} demonstrate that using the \emph{CL} state, which is much higher dimensional than the other global states, is ineffective, particularly in maps with many agents. In comparison, using the \emph{EP} global state achieves stronger performance but notably achieves subpar performance in more difficult maps, likely due to the lack of important local information. The \emph{AS} and \emph{FP} global states both achieve strong performance, with the \emph{FP} state outperforming \emph{AS} states on only several maps. This demonstrates that state dimensionality, agent-specific features, and global information are all important in forming an effective global state. We note that using the \emph{FP} state requires knowledge of which features overlap between
the \emph{EP} state and the agents' local observations, and evaluate MAPPO with this state to demonstrate that limiting the value function input dimensionality can further improve performance. 

\textbf{Suggestion 2:} When available, include both local, agent-specific features and global features in the value function input. Also check that these features do not unnecessarily increase the input dimension. 

\subsection{Training Data Usage} \label{subsec: data}

\begin{figure*}[h!]
	\centering
    \vspace{-3mm}
	\begin{minipage}{1.0\linewidth}
    	\centering
    	\subfigure[effect of different training epochs.\label{fig:ablation-ppo-data-epoch}]{
    	    \label{fig:Ablation-ppo-SMAC}
            \includegraphics[width=1.0\linewidth]{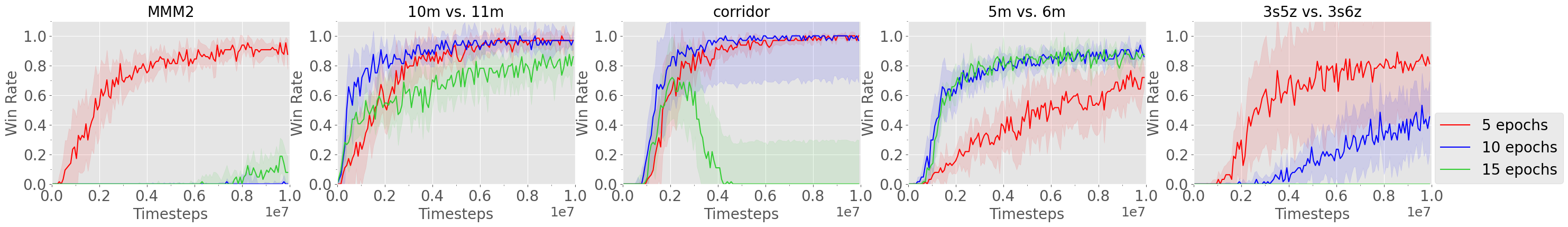}
         }
    \end{minipage}
    \begin{minipage}{1.0\linewidth}
        \centering
        \subfigure[effect of different mini-batch numbers.\label{fig:ablation-ppo-data-minibatch}]{
        \centering
         \label{fig:Ablation-mini-SMAC}
            {
            \includegraphics[width=1.0\linewidth]{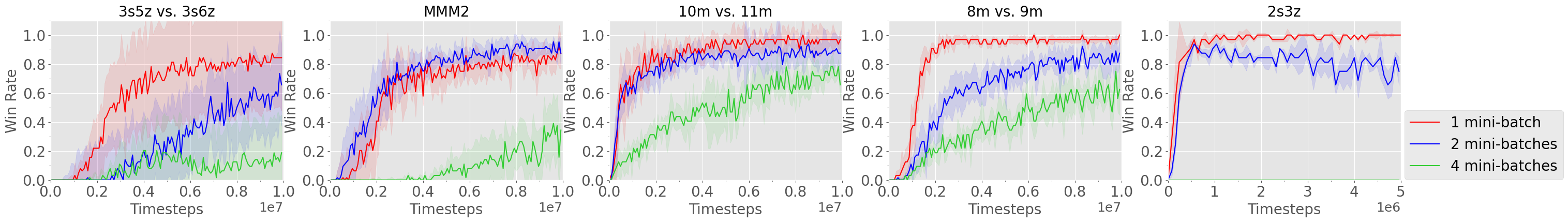}
        	}
        }
    \end{minipage}
    \vspace{-9pt}
    \centering 
    \caption{Effect of epoch and mini-batch number on MAPPO's performance in SMAC.} 
    \vspace{-2mm}
\label{}
\end{figure*}

\begin{figure*}[!h]
	\centering
	\vspace{-12mm}
	\subfigure[effect of different training epochs.]
	{\centering
        {\includegraphics[width=0.2\textwidth]{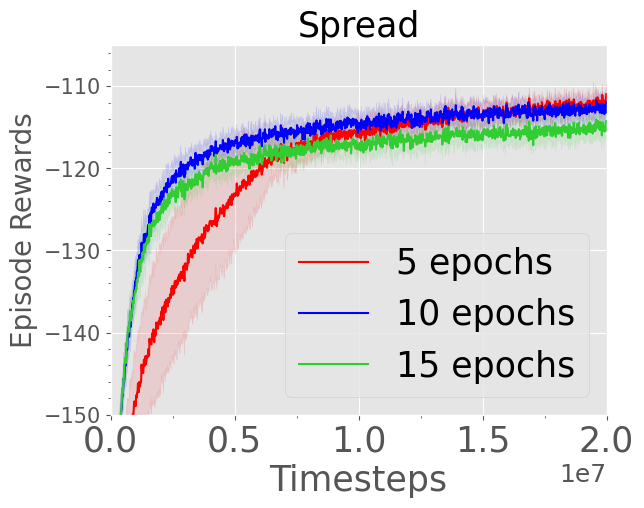}
        \includegraphics[width=0.2\textwidth]{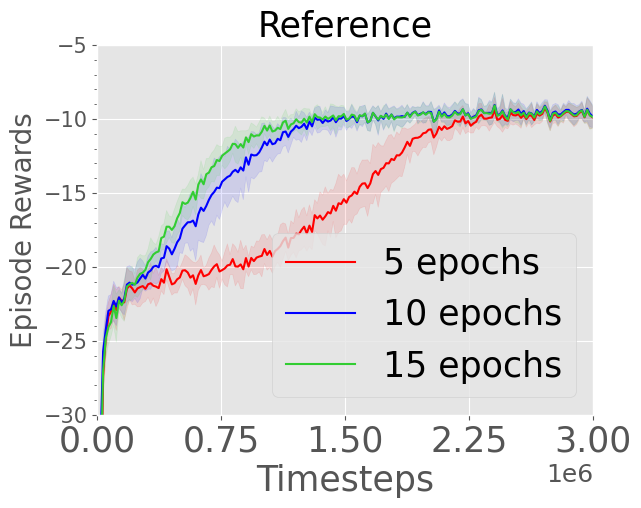}
        \includegraphics[width=0.2\textwidth]{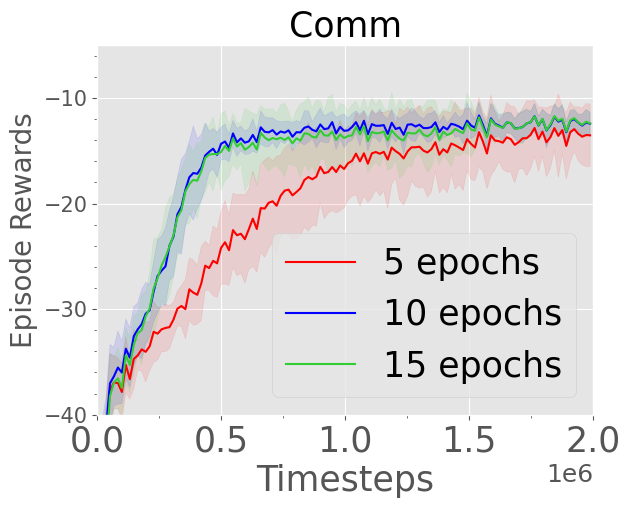}
    	}
    	\label{fig:Ablation-ppo-epoch-MPE}
    }
    \subfigure[effect of different mini-batch numbers.]
    {\centering
        {\includegraphics[width=0.2\textwidth]{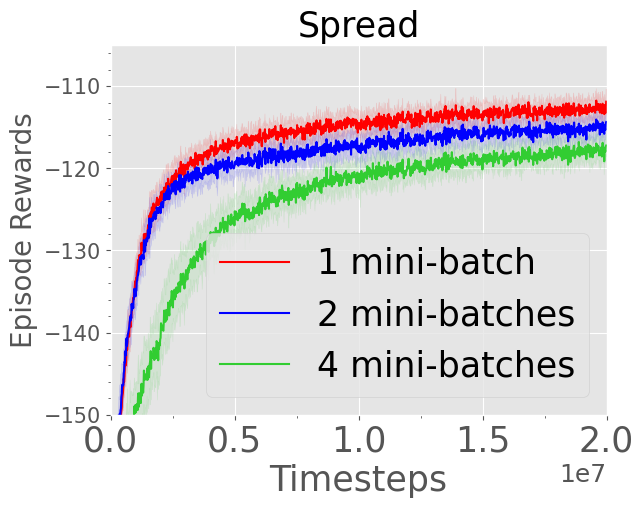}
        \includegraphics[width=0.2\textwidth]{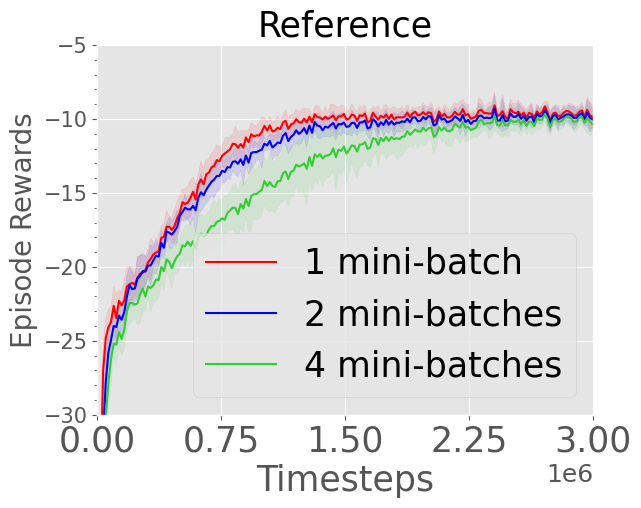}
        \includegraphics[width=0.2\textwidth]{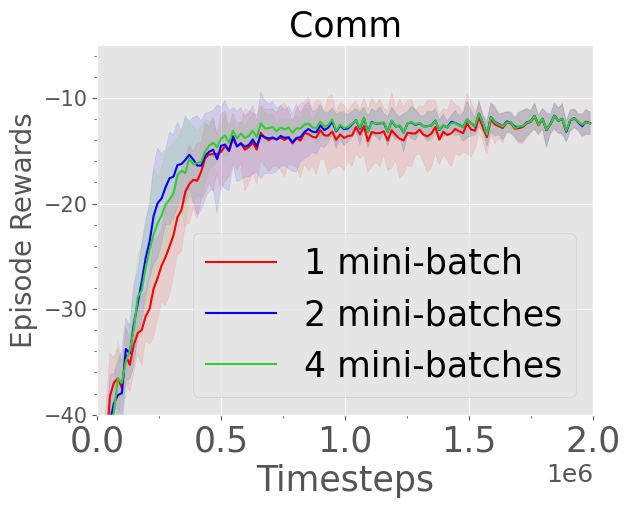}
    	}
    	\label{fig:Ablation-mini-MPE}
    }
    \vspace{-2mm}
	\centering 
	\caption{Effect of epoch and mini-batch number on MAPPO's performance in MPE.}
\label{fig:Ablation-ppo-mini-MPE}
\vspace{-4mm}
\end{figure*}

An important feature of PPO is the use of importance sampling for off-policy corrections, allowing sample reuse.
\cite{stable-baselines} suggest splitting a large batch of collected samples into mini-batches and training for multiple epochs. In single-agent continuous control domains, the common practice is to split a large batch into about 32 or 64 mini-batches and train for tens of epochs. 
However, we find that in multi-agent domains, MAPPO's performance degrades when samples are re-used too often. Thus, we use 15 epochs for easy tasks, and 10 or 5 epochs for difficult tasks. We hypothesize that this pattern could be a consequence of non-stationarity in MARL: using fewer epochs per update limits the change in the agents' policies, which could improve the stability of policy and value learning.
Furthermore, similar to the suggestions by \cite{Ilyas2020A}, we find that using more data to estimate gradients typically leads to improved practical performance. Thus, we split the training data into at-most two mini-batches and avoid mini-batching in the majority of situations.

\textbf{Experimental Analysis:} We study the effect of training epochs in SMAC maps in Fig.~\ref{fig:ablation-ppo-data-epoch}. We observe detrimental effects when training with large epoch numbers: when training with 15 epochs, MAPPO consistently learns a suboptimal policy, with particularly poor performance in the very difficult MMM2 and Corridor maps. In comparison, MAPPO performs well using 5 or 10 epochs.
The performance of MAPPO is also highly sensitive to the number of mini-batches per training epoch. We consider three mini-batch values: 1, 2, and 4. A mini-batch of 4 indicates that we split the training data into 4 mini-batches to run gradient descent.
Fig.~\ref{fig:Ablation-mini-SMAC} demonstrates that using more mini-batches negatively affects MAPPO's performance: when using 4 mini-batches, MAPPO fails to solve any of the selected maps while using 1 mini-batch produces the best performance on 22/23 maps. 
As shown in Fig.~\ref{fig:Ablation-ppo-mini-MPE}, similar conclusions can be drawn in the MPE tasks. In \emph{Reference} and \emph{Comm}, the simplest MPE tasks, all chosen epoch and minibatch values result in the same final performance, and using 15 training epochs even leads to faster convergence.
However, in the harder \emph{Spread} task, we observe a similar trend to SMAC: fewer epochs and no mini-batch splitting produces the best results. 

\textbf{Suggestion 3: } Use at most 10 training epochs on difficult environments and 15 training epochs on easy environments. Additionally, avoid splitting data into mini-batches. 

\subsection{PPO Clipping}
\begin{figure*}[ht]
	\centering
    \vspace{-3mm}
    \includegraphics[width=1.0\linewidth]{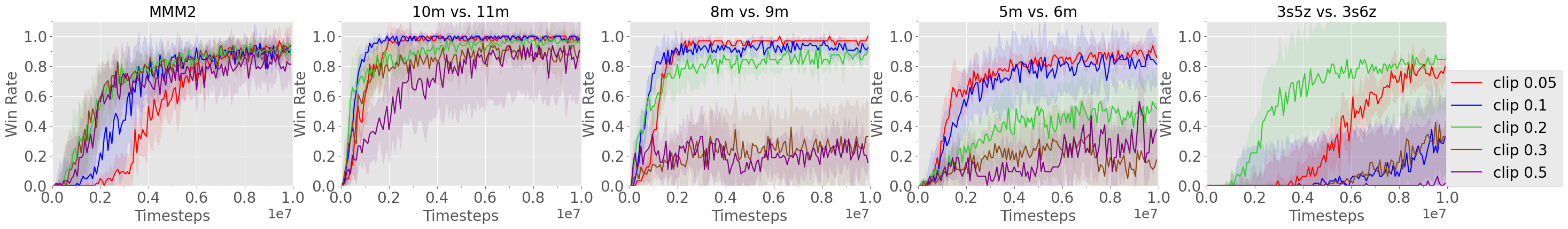}
	\centering \caption{Effect of different clipping strengths on MAPPO's performance in SMAC.}
\label{fig:Abaltion-clip}
\end{figure*}
Another core feature of PPO is the use of clipped importance ratio and value loss to prevent the policy and value functions from drastically changing between iterations. Clipping strength is controlled by the $\epsilon$ hyperparameter: large $\epsilon$ values allow for larger updates to the policy and value function. 
Similar to the number of training epochs, we hypothesize that policy and value clipping can limit the non-stationarity which is a result of the agents' policies changing during training. For small $\epsilon$, agents' policies are likely to change less per update, which we posit improves overall learning stability at the potential expense of learning speed. In single-agent settings, a common $\epsilon$ value is 0.2 \cite{Engstrom2020Implementation, andrychowicz2021what}.

\textbf{Experimental Analysis:} We study the impact of PPO clipping strengths, controlled by the $\epsilon$ hyperparameter, in SMAC (Fig.~\ref{fig:Abaltion-clip}). Note that  $\epsilon$ is the same for both policy and value clipping.  We generally that with small $\epsilon$ terms such as 0.05, MAPPO's learning speed is slowed in several maps, including hard maps such as MMM2 and 3s5z vs. 3s6z. However, final performance when using $\epsilon = 0.05$ is consistently high and the performance is more stable, as demonstrated by the smaller standard deviation in the training curves. We also observe that large $\epsilon$ terms such as 0.2, 0.3, and 0.5, which allow for larger updates to the policy and value function per gradient step, often result in sub-optimal performance. 

\textbf{Suggestion 4:} For the best PPO performance, maintain a clipping ratio $\epsilon$ under 0.2; within this range, tune $\epsilon$ as a trade-off between training stability and fast convergence. 
\vspace{-2mm}

\subsection{PPO Batch Size} 
\vspace{-2mm}
\begin{figure*}[ht]
	\centering
	\vspace{-4mm}
	\subfigure[SMAC]
	{\centering
        {\includegraphics[width=0.23\textwidth]{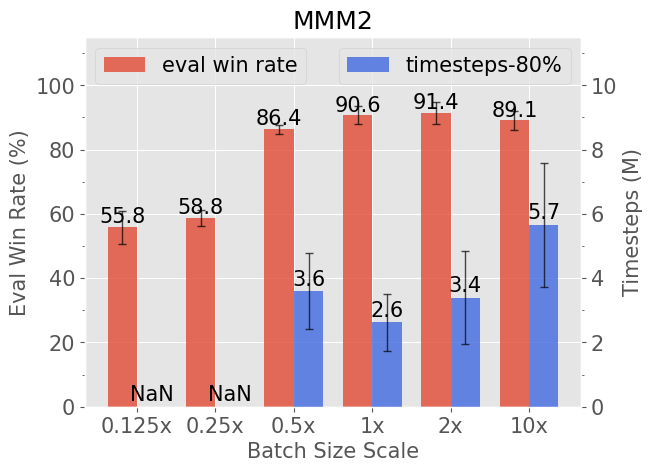}
        \includegraphics[width=0.23\textwidth]{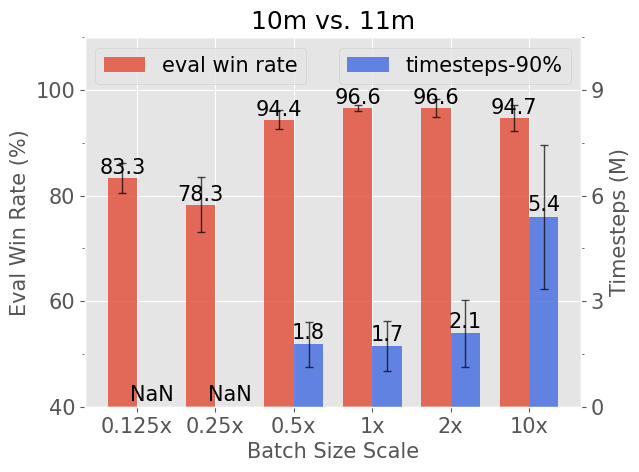}
    	}
    	\label{fig:Ablation-ppo-batch-SMAC}
    }
    \subfigure[GRF]
	{\centering
        {\includegraphics[width=0.23\textwidth]{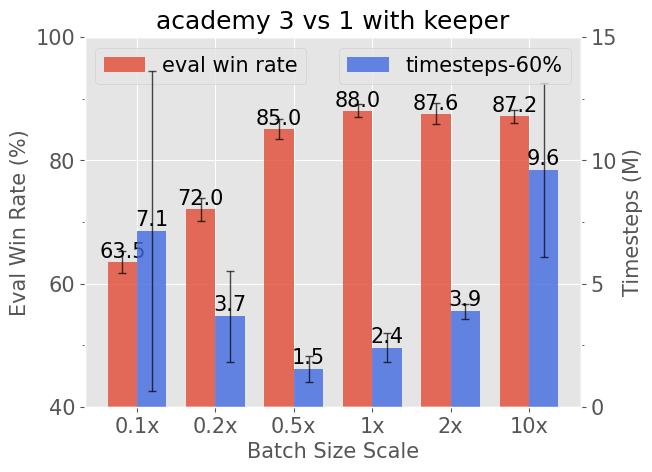}
        \includegraphics[width=0.23\textwidth]{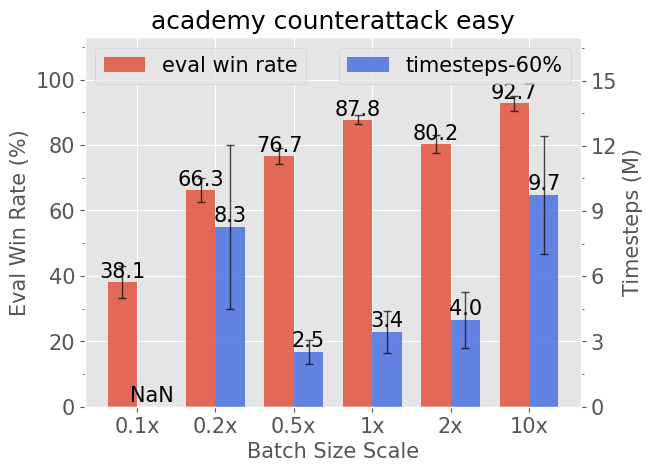}
    	}
    	\label{fig:Ablation-ppo-batch-GRF}
    }
    \vspace{-3mm}
	\centering 
	\caption{Effect of batch size on MAPPO's performance in SMAC and GRF. Red bars show the final win-rates. The blue bars show the number of environment steps required to achieve a strong win-rate (80\% or 90\% in SMAC and 60\% in GRF) as a measure of sample efficiency. ``NaN'' means such a win-rate was never reached. 
	The x-axis specifies the batch-size as a multiple of the batch-size used in our main results. 
	A sufficiently large batch-size is required to achieve the best final performance/sample efficiency; further increasing the batch size may hurt sample efficiency.
	}
\label{fig:Ablation-ppo-batch}
\vspace{-2mm}
\end{figure*}

During training updates, PPO samples a batch of on-policy trajectories which are used to estimate the gradients for the policy and value function objectives. Since the number of mini-batches is fixed in our training (see Sec. \ref{subsec: data}), a larger batch generally will result in more accurate gradients, yielding better updates to the value functions and policies. However, the accumulation of the batch is constrained by the amount of available compute and memory: collecting a large set of trajectories requires extensive parallelism for efficiency and the batches need to be stored in GPU memory. Using an unnecessarily large batch-size can hence be wasteful in terms of required compute and sample-efficiency.


\textbf{Experimental Analysis:} The impact of various batch sizes on both final task performance and sample-efficiency is demonstrated in Fig. \ref{fig:Ablation-ppo-batch}. We observe that in nearly all cases, there is a critical batch-size setting - when the batch-size is below this critical point, the final performance of MAPPO is poor, and further tuning the batch size produces the optimal final performance and sample-efficiency.
However, continuing to increase the batch size may not result in improved final performance and in-fact can worsen sample-efficiency.






\textbf{Suggestion 5:} Utilize a large batch size to achieve best task performance with MAPPO. Then, tune the batch size to optimize for sample-efficiency. 
\vspace{-2mm}
\section{Conclusion} \label{sec:conc} 
\vspace{-3mm}
This work demonstrates that PPO, an on-policy policy gradient RL algorithm, achieves strong results in both final returns and sample efficiency that are comparable to the state-of-the-art methods on a variety of cooperative multi-agent challenges, which suggests that properly configured PPO can be a competitive baseline for cooperative MARL tasks.
We also identify and analyze five key implementation and hyperparameter factors that are influential in PPO's performance in these 
settings. Based on our empirical studies, we give concrete suggestions for the best practices with respect to these factors.
There are a few limitations in this work that point to directions for future study. Firstly, our benchmark environments all use discrete action spaces, are all cooperative, and in the vast majority of cases, contain homogeneous agents. In future work, we aim to test PPO on a wider range of domains such as competitive games and MARL problems with continuous action spaces and heterogeneous agents. Furthermore, our work is primarily empirical in nature, and does not directly analyze the theoretical underpinnings of PPO. We believe that the empirical analysis of our suggestions can serve as starting points for further analysis into PPO's properties in MARL.

\vspace{-2mm}
\subsubsection*{Acknowledgments}
\vspace{-2mm}
This research is supported by NSFC (U20A20334, U19B2019 and M-0248), Tsinghua-Meituan Joint Institute for Digital Life, Tsinghua EE Independent Research Project, Beijing National Research Center for Information Science and Technology (BNRist), Beijing Innovation Center for Future Chips and 2030 Innovation Megaprojects of China (Programme on New Generation Artificial Intelligence) Grant No. 2021AAA0150000.

\clearpage
\newpage

\appendix

\input{appendix.tex}

\newpage

\bibliographystyle{plain}

\bibliography{neurips_data_2022}

\end{document}

%% file: appendix.tex
\section{MAPPO Details}
\label{app:mappo-details}

\begin{algorithm}
    \caption{Recurrent-MAPPO}
    \label{algo:mappo}
\begin{algorithmic}
    \STATE  Initialize $\theta$, the parameters for policy $\pi$ and $\phi$, the parameters for critic $V$, using Orthogonal initialization (Hu et al., 2020) \; 
    \STATE Set learning rate $\alpha$
    \WHILE{$step \leq step_{\text{max}}$}
    \STATE set data buffer $D = \{\}$
    \FOR{$i = 1$ {\bfseries to} $batch\_size$}
    \STATE $\tau = []$ empty list
    \STATE initialize $h_{0, \pi}^{(1)}, \dots h_{0, \pi}^{(n)}$ actor RNN states
    \STATE initialize $h_{0, V}^{(1)}, \dots h_{0, V}^{(n)}$ critic RNN states
    \FOR{$t = 1$ {\bfseries to} $T$}
    \FORALL{agents $a$}
    \STATE $p_t^{(a)}, h_{t, \pi}^{(a)} = \pi(o_t^{(a)}, h_{t-1, \pi}^{(a)}; \theta)$
    \STATE $u_t^{(a)} \sim p_t^{(a)}$
    \STATE $v_t^{(a)}, h_{t, V}^{(a)} = V(s_t^{(a)},h_{t-1, V}^{(a)}; \phi) $
    \ENDFOR 
    \STATE Execute actions $\boldsymbol{u_t}$, observe $r_t, s_{t+1}, \boldsymbol{o_{t+1}}$
    \STATE $\tau += [s_t, \boldsymbol{o_t}, \boldsymbol{h_{t, \pi}}, \boldsymbol{h_{t, V}}, \boldsymbol{u_t}, r_t, s_{t+1}, \boldsymbol{o_{t+1}}]$
    \ENDFOR
    \STATE Compute advantage estimate $\hat{A}$ via GAE on $\tau$, using PopArt
    \STATE Compute reward-to-go $\hat{R}$ on $\tau$ and normalize with PopArt
    \STATE Split trajectory $\tau$ into chunks of length L
    \FOR{l = 0, 1, .., T//L}
    \STATE $D = D \cup (\tau[l : l + T, \hat{A}[l : l + L], \hat{R}[l : l + L])$
    \ENDFOR 
    \ENDFOR 
    \FOR{mini-batch $k=1,\dots,K$}
    \STATE $b \leftarrow$ random mini-batch from D with all agent data
    \FOR{each data chunk $c$ in the mini-batch $b$}
    \STATE update RNN hidden states for $\pi$ and $V$ from first hidden state in data chunk
    \ENDFOR
    \ENDFOR
    \STATE Adam update $\theta$ on $L(\theta)$ with data $b$
    \STATE Adam update $\phi$ on $L(\phi)$ with data $b$
    \ENDWHILE
\end{algorithmic}
\end{algorithm}

MAPPO trains two separate neural networks: an actor network with parameters $\theta$, and a value function network (referred to as a critic) with parameters $\phi$. These networks can be shared amongst all agents if the agents are homogeneous, but each agent can also have its own pair of actor and critic networks. We assume here that all agents share critic and actor networks, for notational convenience.  Specifically, the critic network, denoted as $V_{\phi}$, performs the following mapping: $S \rightarrow \mathbb{R}$. The global state can be agent-specific or agent-agnostic.

The actor network, denoted as $\pi_{\theta}$, maps agent observations $o^{(a)}_t$ to a categorical distribution over actions in discrete action spaces, or to the mean and standard deviation vectors of a Multivariate Gaussian Distribution, from which an action is sampled, in continuous action spaces. 

The actor network is trained to maximize the objective
\newline $L(\theta)\hspace{-3pt} = [\frac{1}{Bn} \sum\limits_{i = 1}^{B} \sum\limits_{k = 1}^n \text{min}( r_{\theta, i}^{(k)}A_i^{(k)}, \text{clip}( r_{\theta, i}^{(k)}, 1 - \epsilon, 1 + \epsilon) A_i^{(k)})] + \sigma \frac{1}{Bn} \sum\limits_{i = 1}^{B} \sum\limits_{k = 1}^n \hspace{-2pt} S[\pi_{\theta}(o_i^{(k)}))]$,
where $r_{\theta, i}^{(k)} = \frac{\pi_{\theta}(a_i^{(k)}|o_i^{(k)})}{\pi_{\theta_{old}}(a_i^{(k)}|o_i^{(k)})}$. $A_i^{(k)}$ is computed using the GAE method, $S$ is the policy entropy, and $\sigma$ is the entropy coefficient hyperparameter. 

The critic network is trained to minimize the loss function
\newline $L(\phi) = \frac{1}{Bn}\sum\limits_{i = 1}^{B} \sum\limits_{k = 1}^n(\text{max}\lbrack(V_\phi(s_i^{(k)})-\hat{R_i})^2,(\text{clip}(V_\phi(s_i^{(k)}),V_{\phi_{old}}(s_i^{(k)})-\varepsilon,V_{\phi_{old}}(s_i^{(k)})+\varepsilon)-\hat{R_i})^2\rbrack$, where $\hat{R_i}$ is the discounted reward-to-go.

In the loss functions above, $B$ refers to the batch size and $n$ refers to the number of agents. 

If the critic and actor networks are RNNs, then the loss functions additionally sum over time, and the networks are trained via Backpropagation Through Time (BPTT). Pseudocode for recurrent-MAPPO is shown in Alg. \ref{algo:mappo}.

\begin{figure*}[bt!]
\vspace{-2mm}
	\centering
    \subfigure[MPE scenarios]{ \label{fig:MPE} 
		\includegraphics[width=0.48\textwidth]{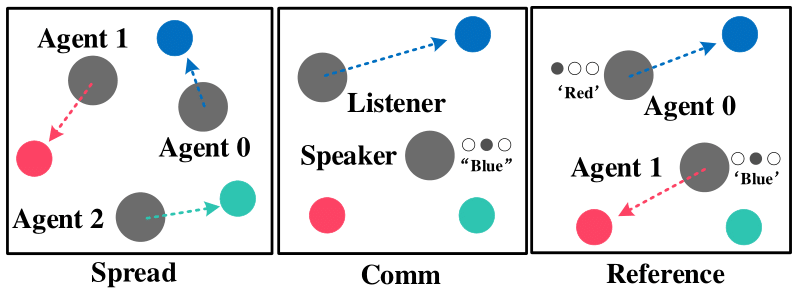}
	}
	\subfigure[4-player Hanabi-Full]{ \label{fig:HanabiPic} 
		\includegraphics[width=0.46\textwidth]{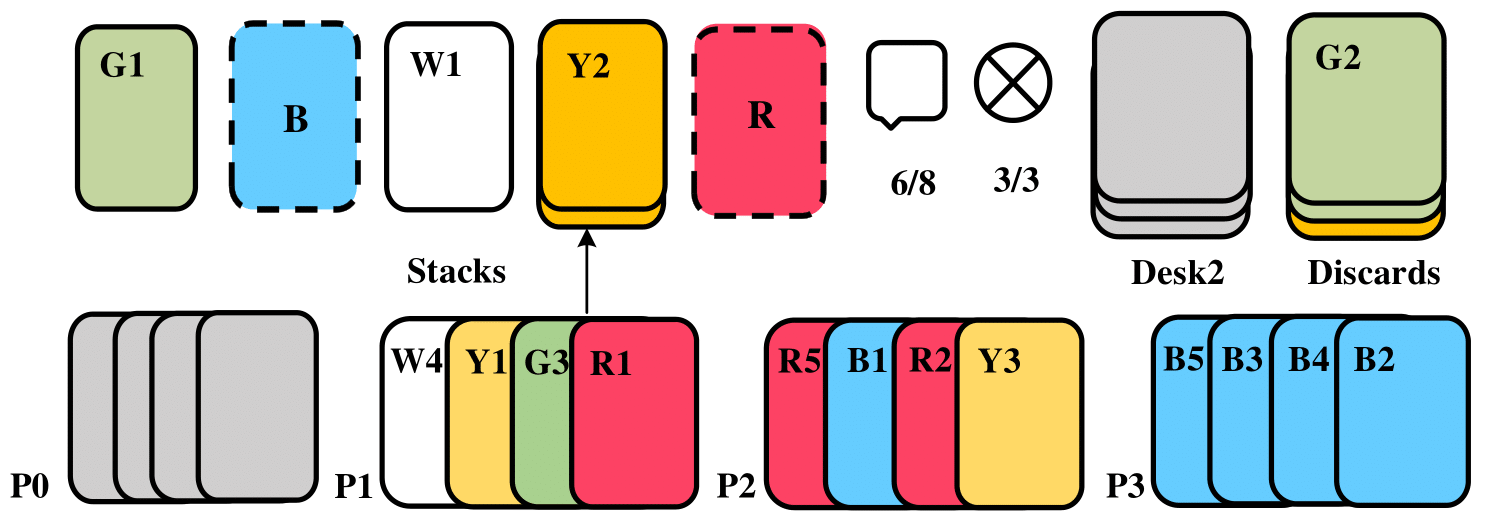}
	}
	\subfigure[SMAC corridor]
	{
	\label{fig:app-SMAC-1}
    \includegraphics[width=0.28\textwidth]{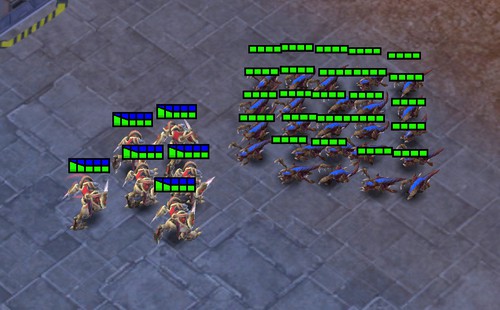}
    }
    \subfigure[SMAC 2c\_vs\_64zg]
	{\label{fig:app-SMAC-2}
    \includegraphics[width=0.28\textwidth]{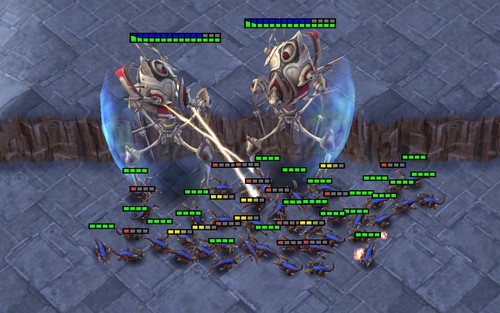}
    }
    \subfigure[GRF academy 3 vs 1 with keeper]
	{\label{fig:app-GRF}
    \includegraphics[width=0.31\textwidth]{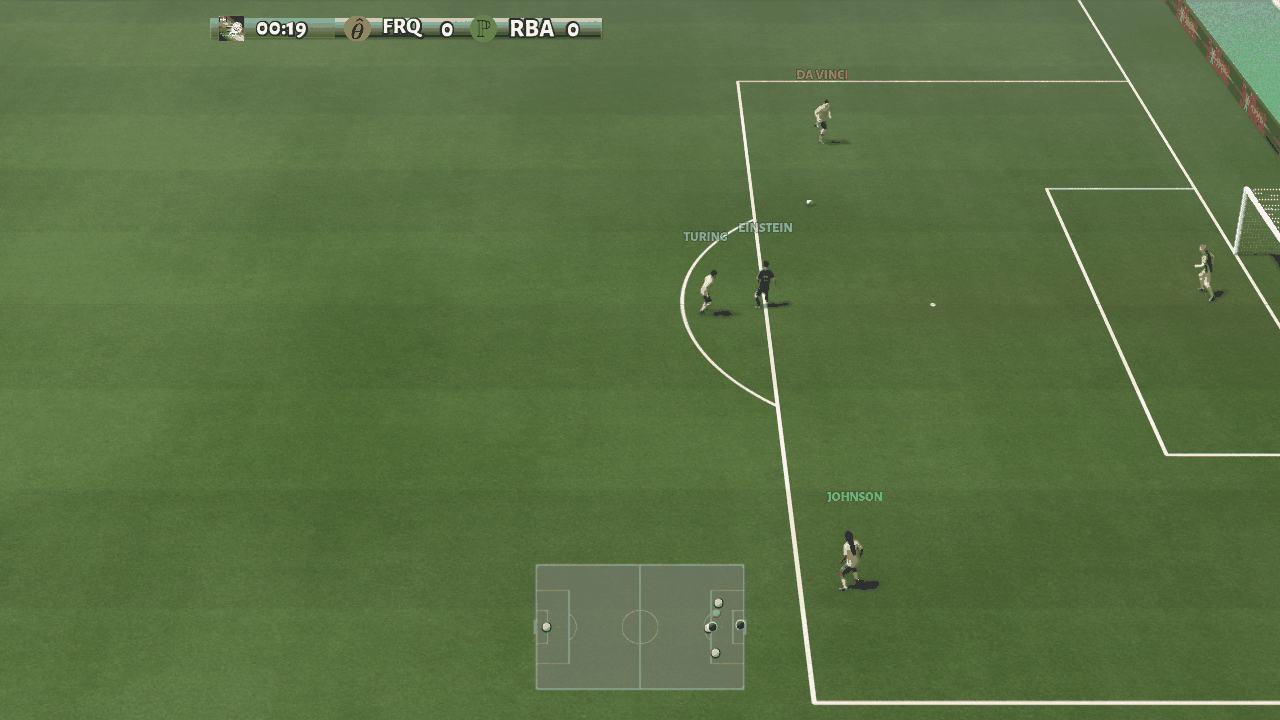}
    }

	\caption{Task visualizations. (a) The MPE domain. \emph{Spread} (left): agents need to cover all the landmarks and do not have a color preference for the landmark they navigate to; \emph{Comm} (middle): the listener needs to navigate to a specific landmarks following the instruction from the speaker; \emph{Reference} (right): both agents only know the other's goal landmark and needs to communicate to ensure both agents move to the desired target.
	(b) The Hanabi domain:  4-player \emph{Hanabi-Full} - figure obtained from (Bard et al., 2020). 
	(c) The \emph{corridor} map in the SMAC domain. 
	(d) The \emph{2c vs. 64zg} map in the SMAC domain.
	(e) The \emph{academy 3 vs 1 with keeper} scenario in the GRF domain.}

\end{figure*}

\section{Testing domains}

\label{app:env}
\textbf{Multi-agent Particle-World Environment (MPE)} was introduced in (Lowe et al., 2017).
MPE consist of various multi-agent games in a 2D world with small particles navigating within a square box. We consider the 3 fully cooperative tasks from the original set shown in Fig.~\ref{fig:MPE}: \emph{Spread}, \emph{Comm}, and \emph{Reference}. Note that since the two agents in \emph{speaker-listener} have different observation and action spaces, this is the only setting in this paper where we do not share parameters but train separate policies for each agent.

\textbf{StarCraftII Micromanagement Challenge (SMAC)} tasks were introduced in (Rashid et al., 2019).
In these tasks, decentralized agents must cooperate to defeat adversarial bots in various scenarios with a wide range of agent numbers (from 2 to 27). We use a global game state to train our centralized critics or Q-functions.
Fig. \ref{fig:app-SMAC-1} and \ref{fig:app-SMAC-2} show two example StarCraftII environments. 

As described in Sec. 5.2, we utilize an agent-specific global state as input to the global state. This agent-specific global state augments the original global state provided by the SMAC environment by adding relevant agent-specific features. 

Specifically, the original global state of SMAC contains information about all agents and enemies - this includes information such as the distance from each agent/enemy to the map center, the health of each agent/enemy, the shield status of each agent/enemy, and the weapon cooldown state of each agent. However, when compared to the local observation of each agent, the global state does not contain agent-specific information including agent id, agent movement options, agent attack options, relative distance to allies/enemies. Note that the local observation contains information only about allies/enemies within a sight radius of the agent. To address the lack of critical local information in the environment provided global state, we create several other global inputs which are specific to each agent, and combine local and global features. The first, which we call \emph{agent-specific (AS)}, uses the concatenation of the environment provided global state and agent \emph{i}'s observation, $o_i$, as the global input to MAPPO's critic during gradient updates for agent \emph{i}. However, since the global state and local agent observations have overlapping features, we additionally create a feature-pruned global state (\emph{FP}) which removes the overlapping features in the $\emph{AS}$ global state. 

\textbf{Hanabi} is a turn-based card game, introduced as a MARL challenge in (Bard et al., 2020) 
, where each agent observes other players' cards except their own cards. A visualization of the game is shown in Fig.~\ref{fig:HanabiPic}. The goal of the game is to send information tokens to others and cooperatively take actions to stack as many cards as possible in ascending order to collect points.

The turn-based nature of Hanabi presents a challenge when computing the reward for an agent during it's turn. We utilize the forward accumulated reward as one turn reward $R_i$; specifically, if there are 4 players and players 0, 1, 2, and 3 execute their respective actions at timesteps \textit{k, k+1, k+2, k+3} respectively, resulting in rewards of $r_k^{(0)}, r_{k+1}^{(1)}, r_{k+2}^{(2)}, r_{k+3}^{(3)}$, then the reward assigned to player 0 will be $R_0 = r_k^{(0)} + r_{k+1}^{(1)} + r_{k+2}^{(2)} + r_{k+3}^{(3)}$ and similarly, the reward assigned to player 1 will be $R_1 = r_{k+1}^{(1)} + r_{k+2}^{(2)} + r_{k+3}^{(3)} + r_{k+4}^{(0)}$. Here, $r_{t}^i$ denotes the reward received at timestep $t$ when agent $i$ is executes a move.

\textbf{Google Research Football (GRF)}, introduced in \cite{kurach2020google}, contains a set of cooperative multi-agent challenges in which a team of agents play a team of bots in various football scenarios. In the scenarios we consider, the goal of the agents is to score a goal against the opposing team. Fig. \ref{fig:app-GRF} shows the example academy scenario. 

The agents' local observations contain a complete description of the environment state at any given time; hence, both the policy and value-function take as input the same observation. At each step, agents share the same reward $R_t$, which is computed as the sum of per-agent rewards $r^{(i)}_t$ which represents the progress made by agent $i$.

\section{Training details} \label{app:training}
\label{app:training-details}
\subsection{Implementation}
\label{app:imple}

\label{app:mappo-imple}
All algorithms utilize parameter sharing - i.e., all agents share the same networks - in all environments except for the \textit{Comm} scenario in the MPE. Furthermore, we tune the architecture and hyperparameters of MADDPG and QMix, and thus use different hyperparameters than the original implementations. However, we ensure that the performance of the algorithms in the baselines matches or exceeds the results reported in their original papers. 

For each algorithm, certain hyperparameters are kept constant across all environments; these are listed in Tables~\ref{tab:MAPPO-common-parameters} and~\ref{tab:QMixMADDPG-common-parameters} for MAPPO, QMix, and MADDPG, respectively. These values are obtained either from the PPO baselines implementation in the case of MAPPO, or from the original implementations for QMix and MADDPG. Note that since we use parameter sharing and combine all agents' data, the actual batch-sizes will be larger with more agents. 

In these tables, ``recurrent data chunk length'' refers to the length of chunks that a trajectory is split into before being used for training via BPTT (only applicable for RNN policies). ``Max clipped value loss'' refers to the value-clipping term in the value loss. ``Gamma'' refers to the discount factor, and ``huber delta'' specifies the delta parameter in the Huber loss function. ``Epsilon'' describes the starting and ending value of $\epsilon$ for $\epsilon$-greedy exploration, and ``epsilon anneal time'' refers to the number of environment steps over which $\epsilon$ will be annealed from the starting to the ending value, in a linear manner. ``Use feature normalization'' refers to whether the feature normalization is applied to the network input.

\subsection{Parameter Sharing}\label{app:sharing}
In the main results which are presented, we utilize parameter sharing - a technique which has been shown to be beneficial in a variety of state-of-the-art methods ~\cite{christianos2021scaling, terry2021revisiting} in all algorithms for a fair comparison. Specifically, both the policy and value network parameters are shared across all agents. In this appendix section, we include results which demonstrate the benefit of parameter sharing. Table~\ref{tab:share-smac} shows median evaluation win rate (with standard deviation in parantheses) on selected SMAC maps over 6 random seeds. MAPPO-Ind is MAPPO denotes MAPPO without parameter sharing - e.g., each agent has a separate policy and value function network. We observe that MAPPO with parameter sharing outperforms MAPPO without parameter sharing by a clear margin, supporting our decision to adopt parameter sharing in all PPO experiments and all baselines used in our results. A more theoretical analysis of the effect of parameter sharing can be found in ~\cite{https://doi.org/10.48550/arxiv.2206.07505}.

\begin{table}
\centering
\begin{tabular}{ccccc}
\toprule
Map & MAPPO & MAPPO-Ind\\
\midrule
      1c3s5z & \textbf{100.0(0.0)} & 99.1(0.7)\\
         2s3z & \textbf{100.0(0.7)} &     99.1(0.9) \\
     3s\_vs\_5z & \textbf{100.0(0.6)} &    93.8(1.8)\\
         3s5z & \textbf{96.9(0.7)} &     80.4(3.3) \\
 3s5z\_vs\_3s6z & 84.4(34.0) &   37.8(5.6)\\
     5m\_vs\_6m & \textbf{89.1(2.5)} &    44.4(2.9)\\
     6h\_vs\_8z & \textbf{88.3(3.7)} &   11.4(2.5)\\
  10m\_vs\_11m & \textbf{96.9(4.8)} & 78.4(2.7) \\
     corridor & \textbf{100.0(1.2)} &   82.2(1.8)\\
         MMM2 & 90.6(2.8) &     13.0(3.7)\\
\bottomrule
\end{tabular}
\caption{Median evaluation win rate (standard deviation) on selected SMAC maps over 6 random seeds.}
\label{tab:share-smac}
\end{table}



\subsection{Death Masking}
\begin{figure*}[h]
	\centering
    \includegraphics[width=1.0\textwidth]{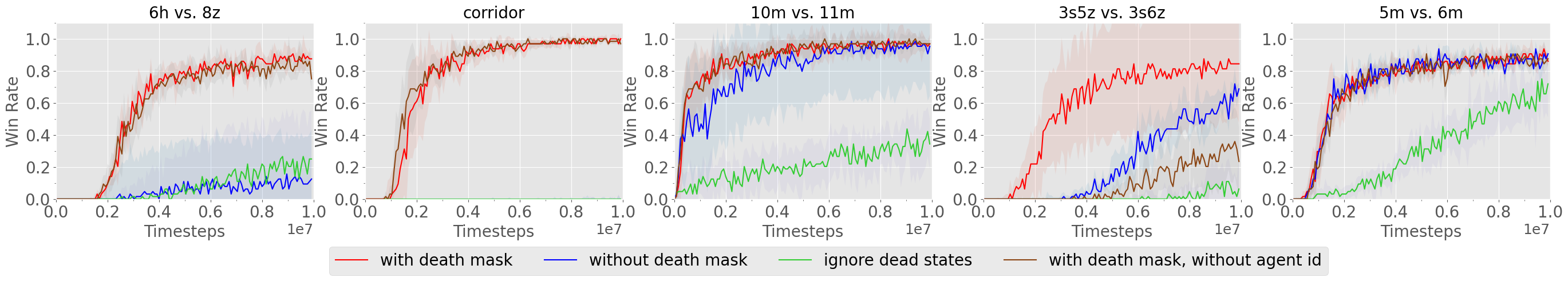}
	\centering \caption{The effect of death mask on MAPPO's performance in SMAC.}
\label{fig:Abaltion-deathmask}
\vspace{-2mm}
\end{figure*}
In SMAC, it is possible through the course of an episode for certain agents to become inactive, or ``die'' while other agents remain active in the environment. In this setting, while the local observation for a dead agent becomes all zeros except for the agent's ID, the value-state still contains other nonzero features about the environment. When computing the GAE for an agent during training, it is unclear how to handle timesteps in which the agent is dead. We consider four options: (1) in which we replace the value state for a dead agent with a zero state containing the agent ID (similar to it's local observation). We refer to this as ``death masking''; (2) MAPPO without death masking, i.e., still using the nonzero global state as value input; (3) completely drop the transition samples after an agent dies (note that we still need to accumulate rewards after the agent dies to correctly estimate episode returns); and (4) replacing the global state with a pure zero-state which does not include the agent ID.
 Fig.~\ref{fig:Abaltion-deathmask} demonstrates that variant (1) significantly outperforms variants (2) and (3), and consistently achieves overall strong performance. Including the agent id in the death mask, as is done in variant (1), is particularly important in maps which agents may take on different roles, as demonstrated by the superior performance of variant (1) compared to variant (4), which does not contain the agent ID in the death-mask zero-state, in the 3s5z vs. 3s6z map.

\textbf{Justification of Death Masking}
\label{app:math}
Let $\boldsymbol{0}_a$ be a zero vector with agent \emph{a}'s agent ID appended to the end. 
The use of agent ID leads to an agent-specific value function depending on an agent's type or role. It has been empirically justified that such an agent-specific feature is particularly helpful when the environment contains heterogeneous agents. 

We now provide some intuition as to why using $\boldsymbol{0_a}$ as the critic input when agents are dead appears to be a better alternative to using the usual agent-specific global state as the input to the value function. Note that our global state to the value network has agent-specific information, such as available actions and relative distances to other agents. When an agent dies, these agent-specific features become zero, while the remaining agent-agnostic features remain nonzero - this leads to a drastic distribution shift in the critic input compared to states in which the agent is alive. In most SMAC maps, an agent is dead in only a small fraction of the timesteps in a batch (about 20\%); due to their relative infrequency in the training data the states in which an agent is dead will likely have large value prediction error. 
Moreover, it is also possible that training on these out of distribution inputs harms the feature representation of the value network.

Although replacing the states at which an agent is dead with a fixed vector $\boldsymbol{0_a}$ also results in a distribution shift, the replacement results in there being only 1 vector which captures the state at which an agent is dead - thus, the critic is more likely to be able to fit the average post-death reward for agent $a$ to the input $\boldsymbol{0_a}$. Our ablation on the value function fitting error provide some weight to this hypothesis.

Another possible mechanism of handling agent deaths is to completely skip value learning in states in which an agent is dead, by essentially terminating an agent's episode when it dies. Suppose the game episode is $T$ and the agent dies at timestep $d$. If we are not learning on dead state then, in order to correctly accumulate the episode return, we need to replace the reward $r_d$ at timestep $d$ by the total return $R_d$ at time $d$, i.e., $r_d\gets R_d=\sum_{t=d}^T \gamma^{t-d}r_t$. We would then need to compute the GAE only on those states in which the agent is alive. While this approach is theoretically correct (we are simply treating the state where the agent died as a terminal state and assigning the accumulated discounted reward as a terminal reward), it can have negative ramifications in the policy learning process, as outlined below.

The GAE is an exponentially weighted average of $k$-step returns intended to trade off between bias and variance. Large $k$ values result in a low bias, but high variance return estimate, whereas small $k$ values result in a high bias, low variance return estimate. However, since the entire post death return $R_d$ replaces the single timestep reward $r_d$ at timestep $d$, computing the 1-step return estimate at timestep $d$ essentially becomes a ($T-d$)-step estimate, eliminating potential benefits of value function truncation of the trajectory and potentially leading to higher variance. This potentially dampens the benefit that could come from using the GAE at the timesteps in which an agent is dead. 

We analyze the impact of the death masking by comparing different ways of handling dead agents, including: (1) our death masking, (2) using global states without death masking and (3) ignoring dead states in value learning and in the GAE computation. We first examine the median win rate with these different options in Fig. \ref{fig:app-Ablation-death} and ~\ref{fig:app-Ablation-ignore}. It is evident that our method of death masking, which uses $\boldsymbol{0_a}$ as the input to the critic when an agent is dead, results in superior performance compared to other options. 

Fig.~\ref{fig:app-Ablation-ignore_valueloss} also demonstrates that using the death mask results in a lower values loss in the vast majority of SMAC maps, demonstrating that the accuracy of the value predictions improve when using the death mask. While the arguments here are intuitive the clear experimental benefits suggest that theoretically characterizing the effect of this method would be valuable.

\subsection{Hyperparameters}
\
Tables 4-16 describe the common hyperparameters, hyperparameter grid search values, and chosen hyperparmeters for MAPPO, QMix, and MADDPG in all testing domains. Tables~\ref{tab:typical-MPE}, \ref{tab:typical-SMAC}, \ref{tab:typical-Hanabi}, and \ref{tab:typical-GRF} describe common hyperparameters for different algorithms in each domain. Tables \ref{tab:MAPPO-tune-hyper}, \ref{tab:QMix-tune-hyper}, and \ref{tab:MADDPG-tune-hyper} describe the hyperparameter grid search procedure for the MAPPO, QMix, and MADDPG algorithms, respectively. Lastly, Tables~\ref{tab:diff-MPE},~\ref{tab:diff-SMAC},~\ref{tab:diff-Hanabi} and~\ref{tab:diff-Football} describe the final chosen hyperparameters among fine-tuned parameters for different algorithms in MPE, SMAC, Hanabi, and GRF, respectively. 

For MAPPO, ``Batch Size'' refers to the number of environment steps collected before updating the policy via gradient descent. Since agents do not share a policy only in the MPE speaker-listener, the batch size does not depend on the number of agents in the speaker-listener environment. ``Mini-batch'' refers to the number of mini-batches a batch of data is split into, ``gain'' refers to the weight initialization gain of the last network layer for the actor network. ``Entropy coef'' is the entropy coefficient $\sigma$ in the policy loss. ``Tau'' corresponds to the rate of the polyak average technique used to update the target networks, and if the target networks are not updated in a ``soft'' manner, the ``hard interval'' hyperparameter specifies the number of gradient updates which must elapse before the target network parameters are updated to equal the live network parameters. ``Clip'' refers to the $\epsilon$ hyperparameter in the policy objective and value loss which controls the extent to which large policy and value function changes are penalized. 

MLP network architectures are as follows:
all MLP networks use ``num fc'' linear  layers, whose dimensions are specified by the ``fc layer dim'' hyperparameter. When using MLP networks, ``stacked frames'' refers to the number of previous observations which are concatenated to form the network input: for instance, if ``stacked frames'' equals 1, then only the current observation is used as input, and if ``stacked frames'' is 2, then the current and previous observations are concatenated to form the input. For RNN networks, the network architecture is ``num fc'' fully connected linear layers of dimension ``fc layer dim'', followed by ``num GRU layers'' GRU layers, finally followed by ``num fc after''  linear layers. 

\section{Additional Results}
\label{app:add-details}

\subsection{Additional SMAC Results}
\label{app:SMAC-allresults}

Results of all algorithms in all SMAC maps can be found in Tab. \ref{tab:app-SMAC-results-all} and \ref{tab:app-SMAC-results-cut}.

\begin{figure*}[ht]
\captionsetup{justification=centering}
\centering
\includegraphics[width=0.35\textwidth]{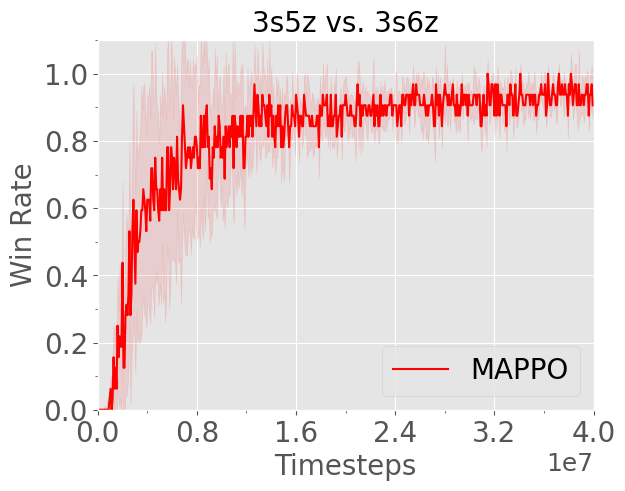}
\centering
\caption{Median win rate of 3s5z vs. 3s6z map after 40M environment steps.}
\label{fig:app-longrun}
\end{figure*}

\begin{figure*}[ht]
	\centering
	\subfigure
	{
        \includegraphics[width=0.25\textwidth]{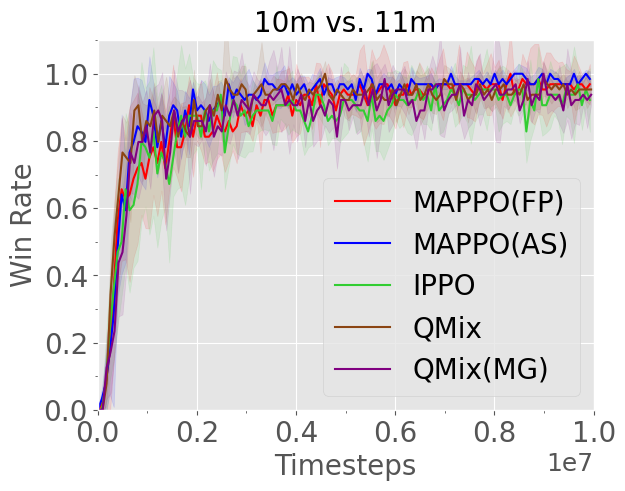}
        \includegraphics[width=0.25\textwidth]{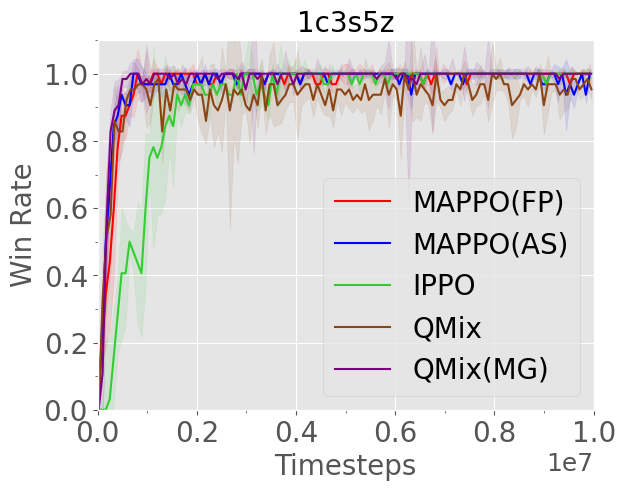}
        \includegraphics[width=0.25\textwidth]{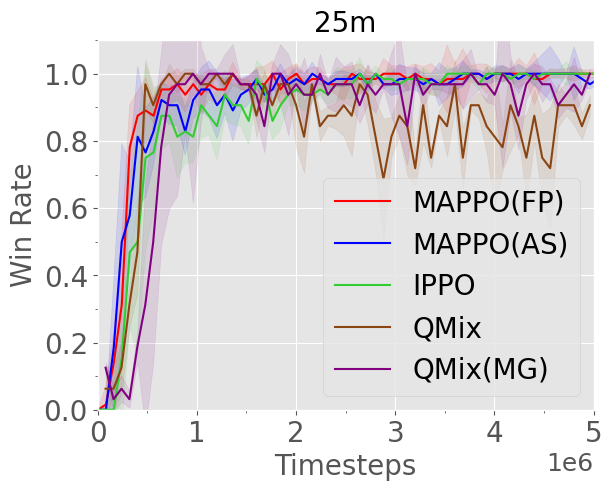}
        \includegraphics[width=0.25\textwidth]{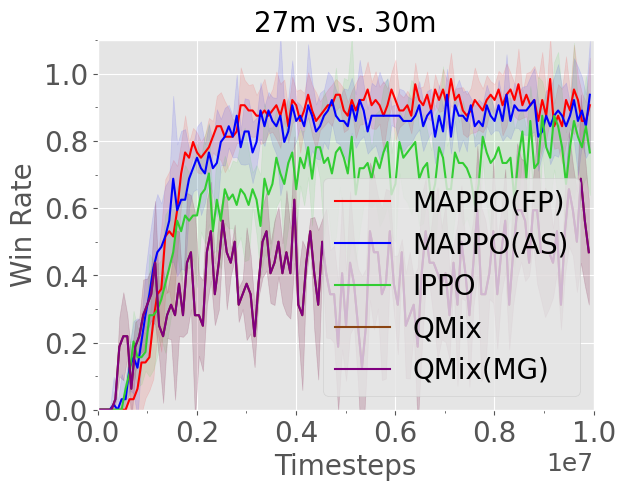}
    }
    \subfigure
    {
        \includegraphics[width=0.25\textwidth]{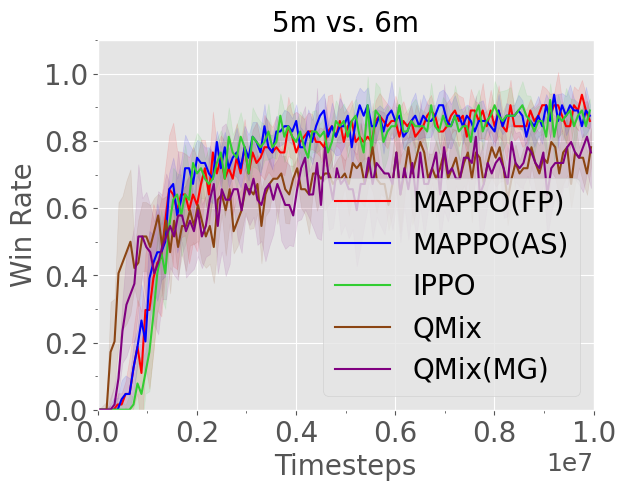}
        \includegraphics[width=0.25\textwidth]{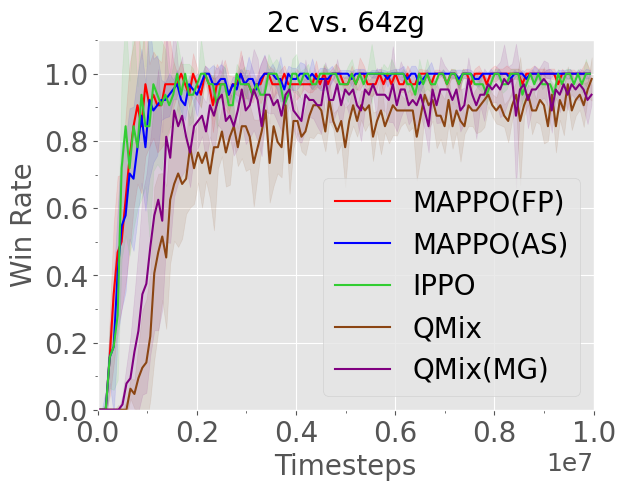}
        \includegraphics[width=0.25\textwidth]{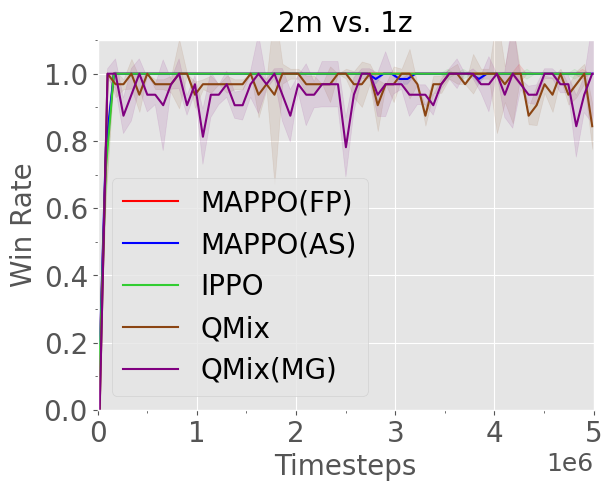}
        \includegraphics[width=0.25\textwidth]{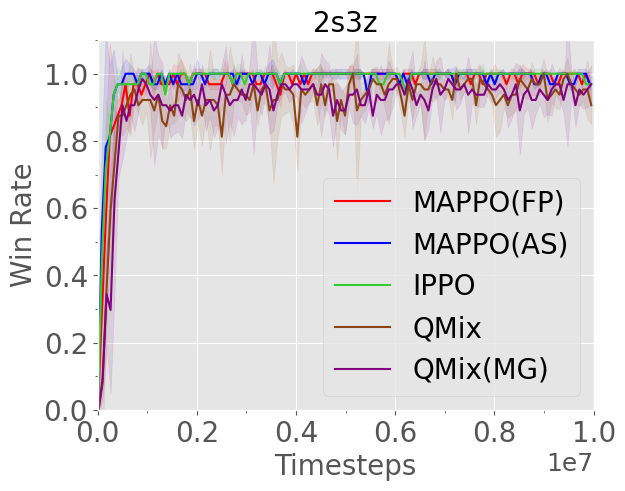}
    }
    \subfigure
    {
        \includegraphics[width=0.25\textwidth]{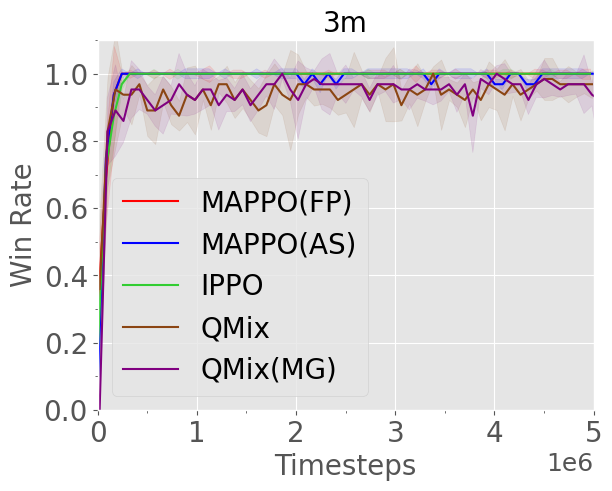}
        \includegraphics[width=0.25\textwidth]{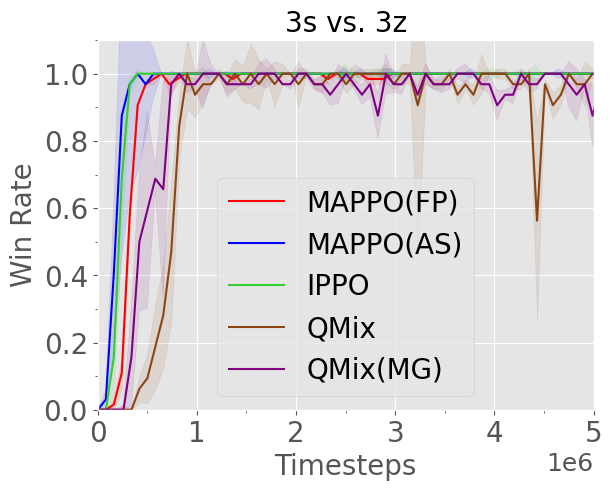}
        \includegraphics[width=0.25\textwidth]{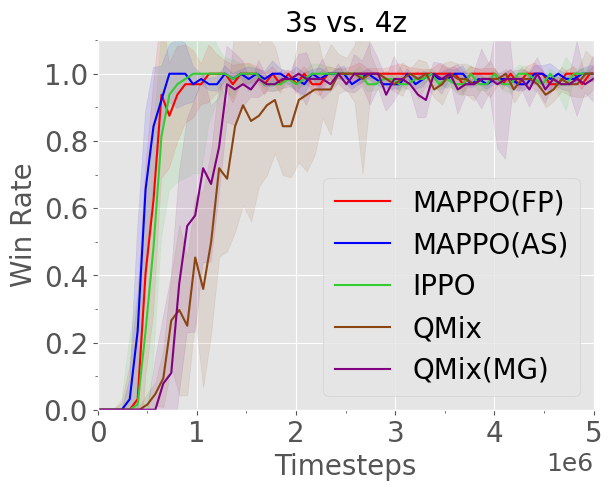}
        \includegraphics[width=0.25\textwidth]{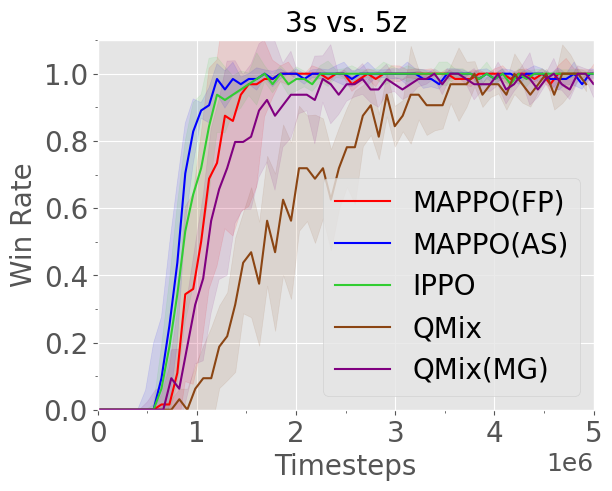}
    	
    }
    \subfigure
    {
        \includegraphics[width=0.25\textwidth]{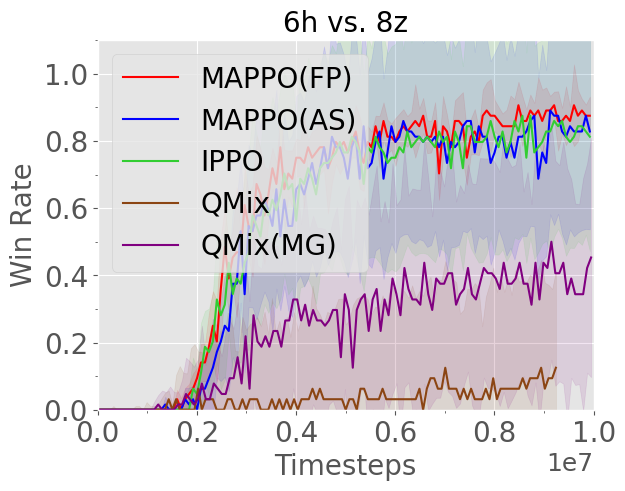}
        \includegraphics[width=0.25\textwidth]{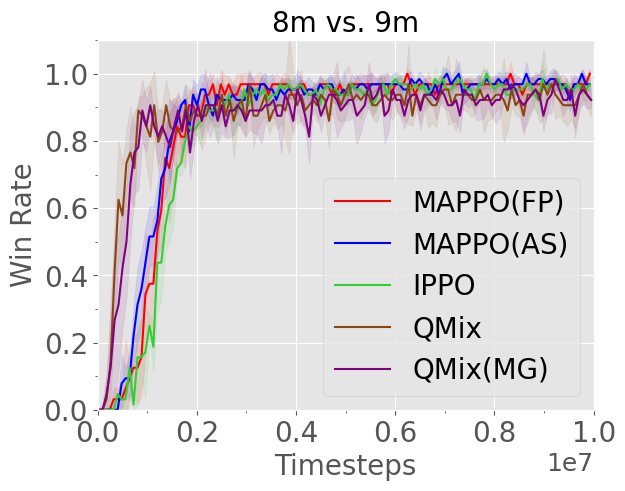}
        \includegraphics[width=0.25\textwidth]{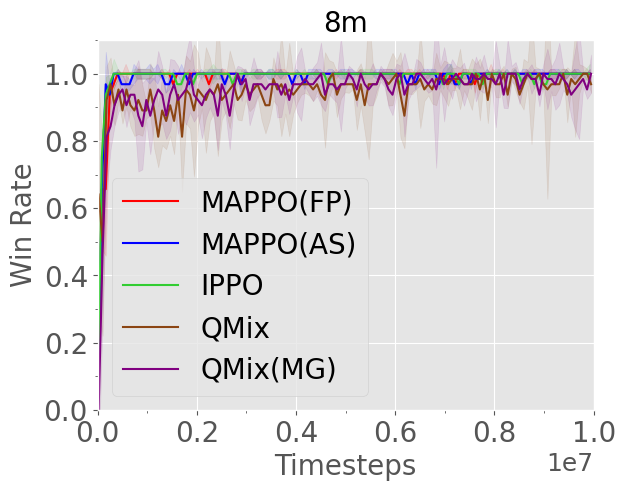}
        \includegraphics[width=0.25\textwidth]{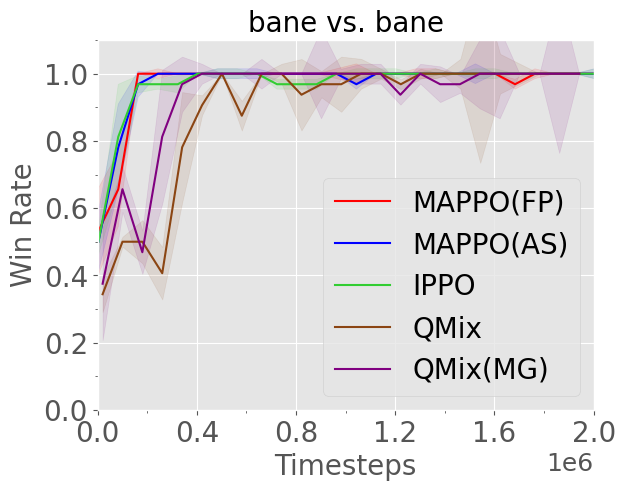}
    }
    \subfigure
    {
        \includegraphics[width=0.25\textwidth]{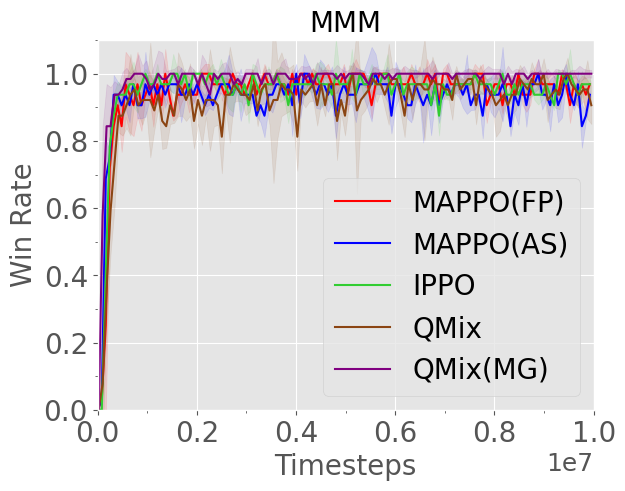}
        \includegraphics[width=0.25\textwidth]{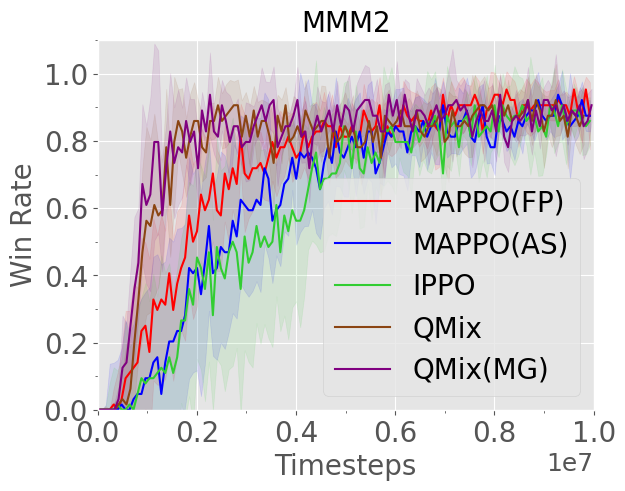}
        \includegraphics[width=0.25\textwidth]{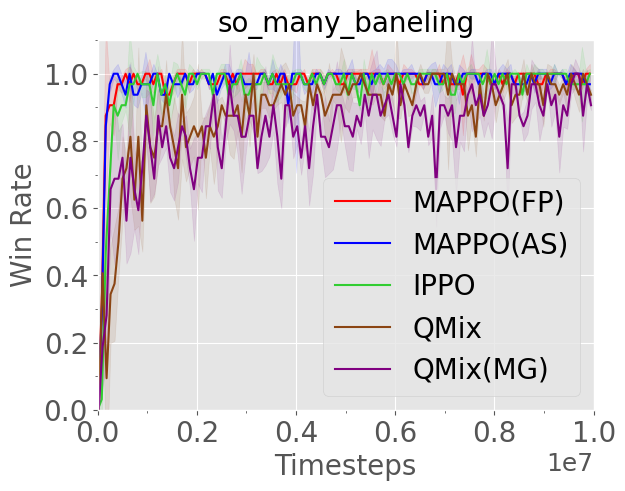}
        \includegraphics[width=0.25\textwidth]{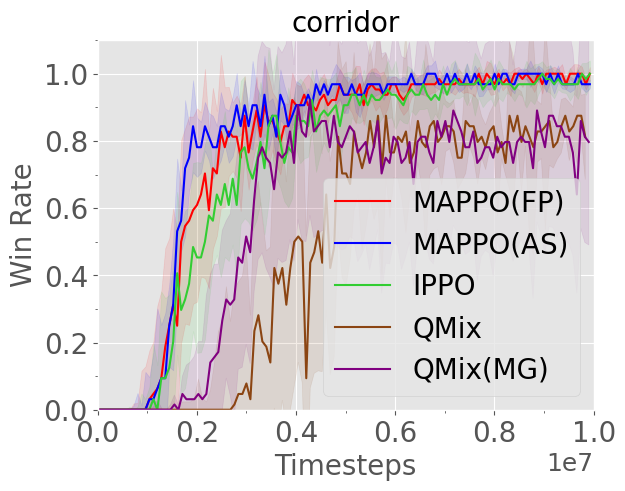}
    
    }
    \subfigure
    {
         
        \includegraphics[width=0.25\textwidth]{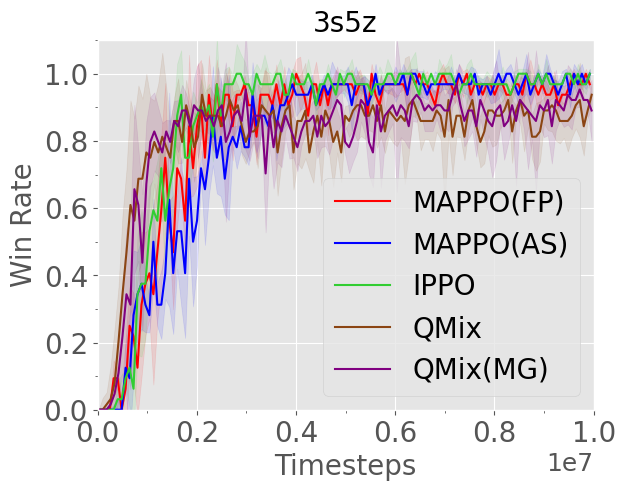}
        \includegraphics[width=0.25\textwidth]{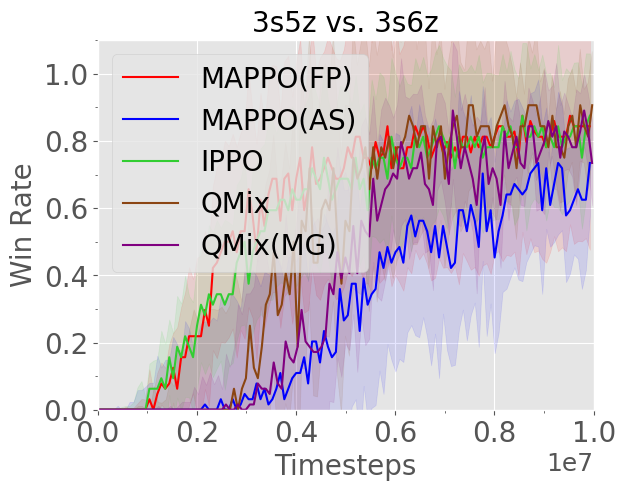}

    }
	\centering \caption{Median evaluation win rate of 23 maps in the SMAC domain.  }
\label{fig:app-win-rate}
\end{figure*}

As MAPPO does not converge within 10M environment steps in the 3s5z vs. 3s6z map, Fig. \ref{fig:app-longrun} shows the performance of MAPPO in 3s5z vs. 3s6z when run until convergence. Fig.~\ref{fig:app-win-rate} presents the evaluation win of MAPPO with different value inputs (\emph{FP} and \emph{AS}), decentralized PPO (IPPO), QMix, and QMix with a modified global state input to the mixer network, which we call QMix (MG). Specifically, QMix(MG) uses a concatenation of the default environment global state, as well as \emph{all} agents' local observations, as the mixer network input.

\begin{figure*}[ht]
	\centering
	\subfigure
	{
        {
        \includegraphics[width=0.25\textwidth]{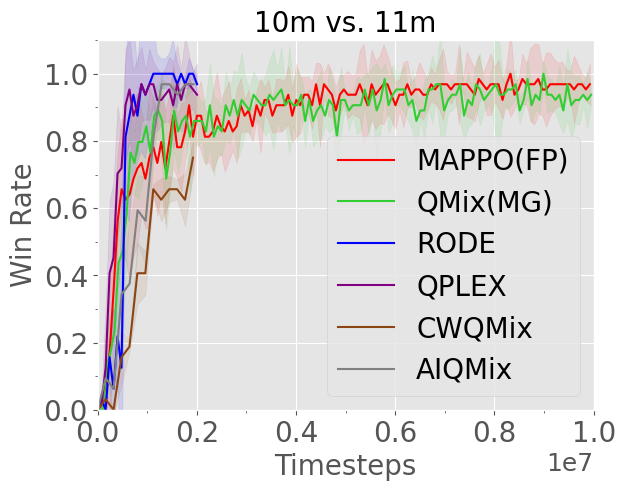}
        \includegraphics[width=0.25\textwidth]{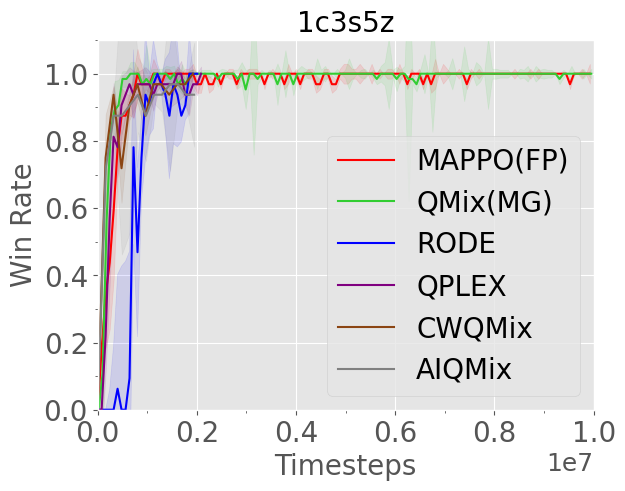}
        \includegraphics[width=0.25\textwidth]{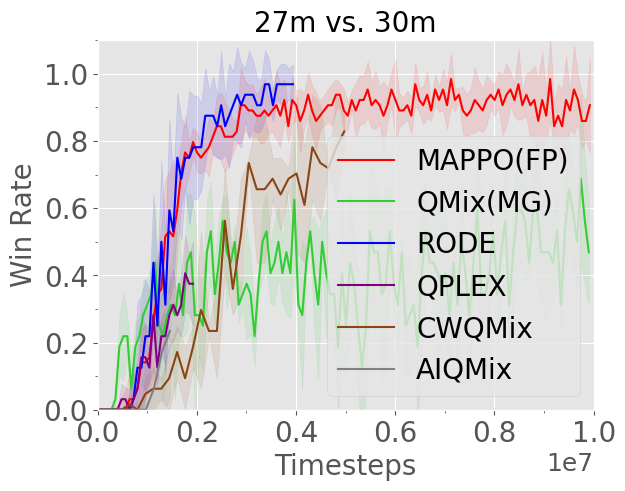}
        \includegraphics[width=0.25\textwidth]{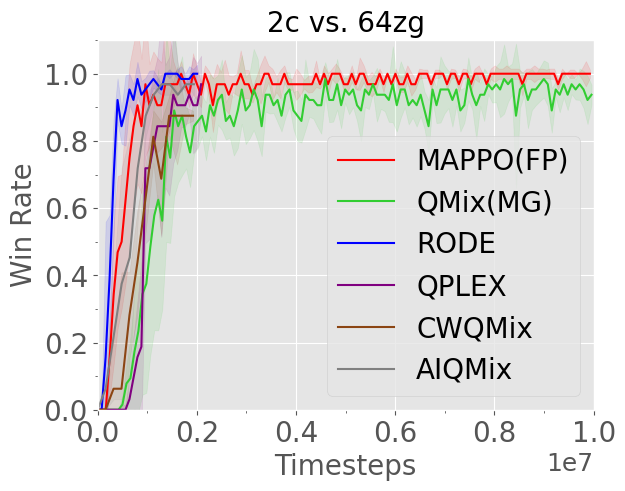}
    	}
    }
    \subfigure
    {
        \includegraphics[width=0.25\textwidth]{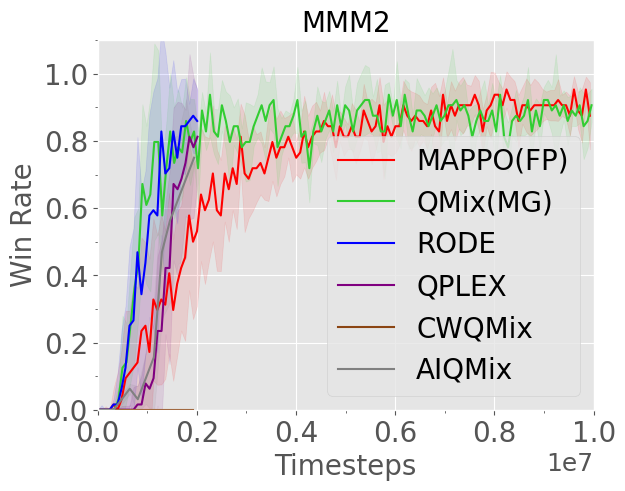}
        \includegraphics[width=0.25\textwidth]{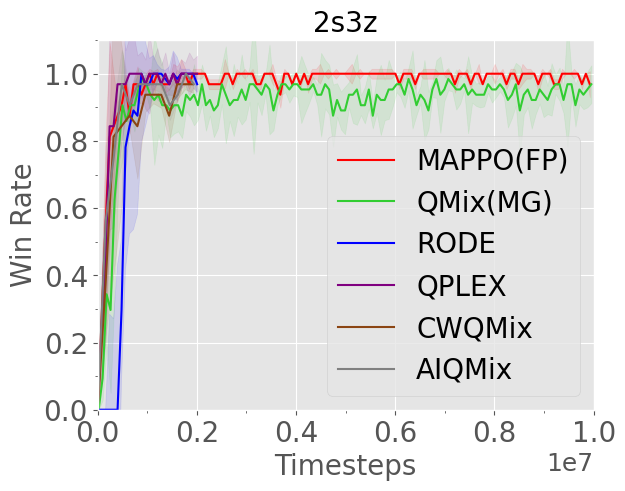}
        \includegraphics[width=0.25\textwidth]{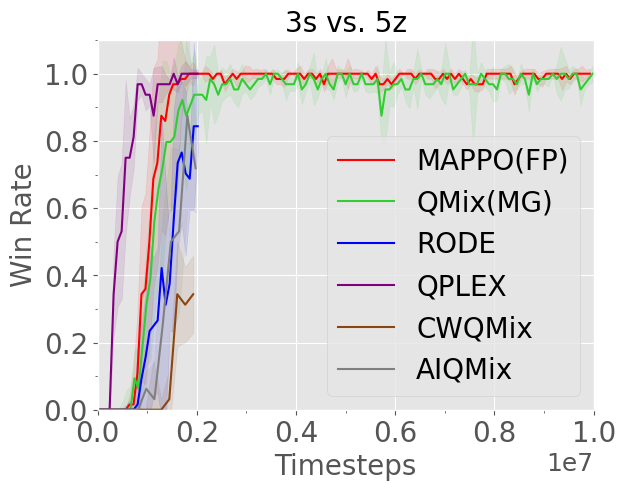}
        \includegraphics[width=0.25\textwidth]{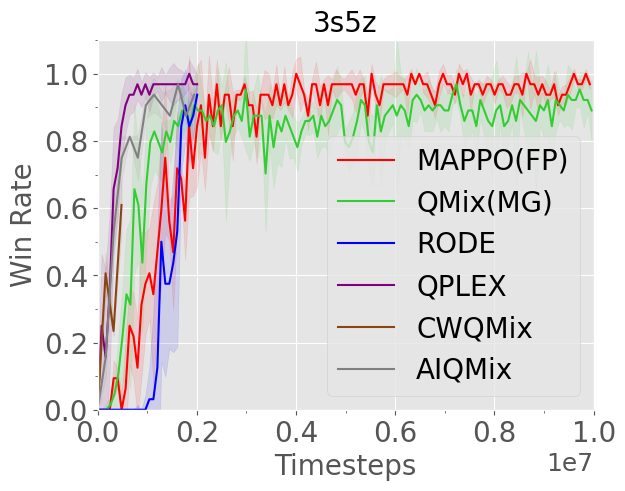}
    	
    }
    \subfigure
    {
        \includegraphics[width=0.25\textwidth]{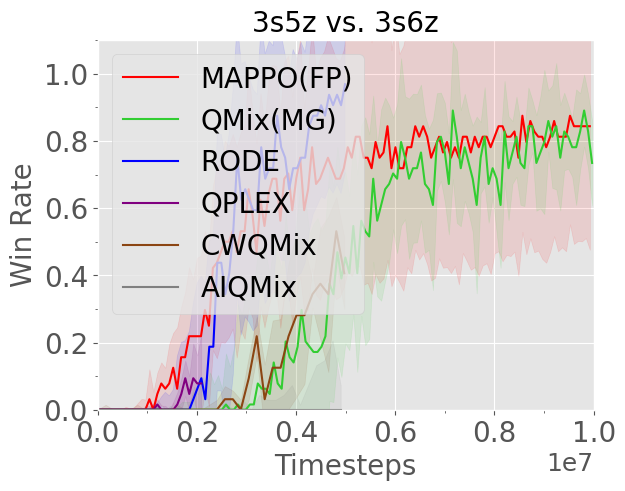}
        \includegraphics[width=0.25\textwidth]{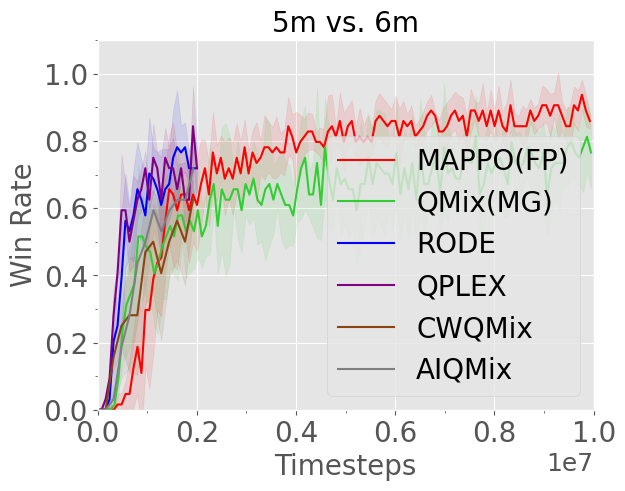}
        \includegraphics[width=0.25\textwidth]{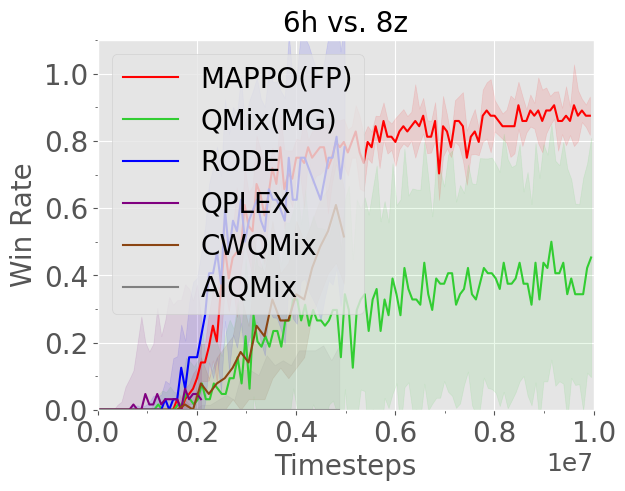}
        \includegraphics[width=0.25\textwidth]{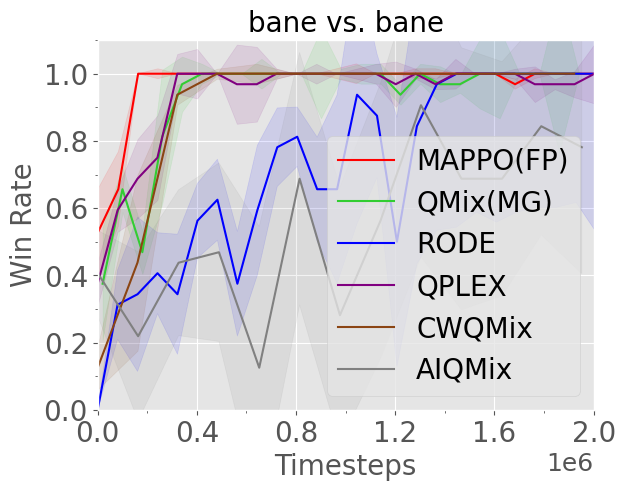}
    	
    }
    \subfigure
    {
        \includegraphics[width=0.25\textwidth]{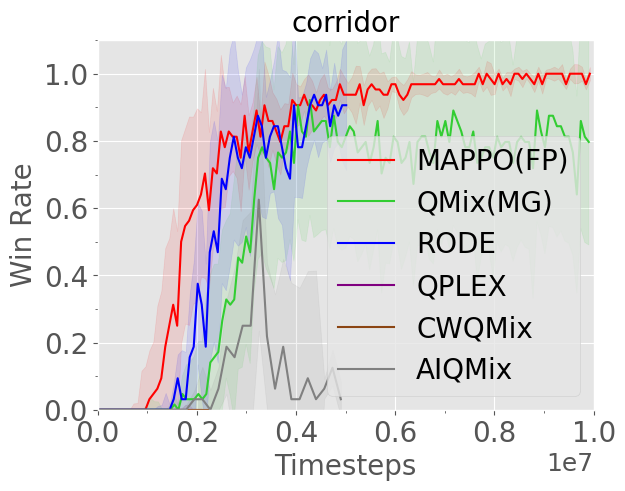}
        \includegraphics[width=0.25\textwidth]{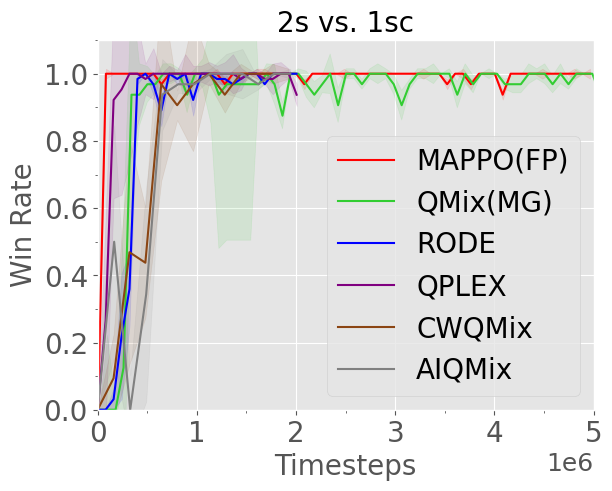}
        
    }
	\centering \caption{Median evaluation win rate of MAPPO(FP), QMix(MG), RODE, QPlEX, CWQMix and AIQMix in the SMAC domain.}
\label{fig:app-rode}
\end{figure*}

Fig. \ref{fig:app-rode} compares the results of MAPPO(FP) to various off-policy baselines, including QMix(MG), RODE, QPLEX, CWQMix, and AIQMix, in many SMAC maps. Both QMIX and RODE utilize both the agent-agnostic global state and agent-specific local observations as input. Specifically, for agent $i$, the local Q-network (which computes actions at execution) takes in only the local agent-specific observation $o_i$ as input while the global mixer network takes in the agent-agnostic global state $s$ as input.  This is also the case for the other value-decomposition methods presented in Appendix Table 1 (QPLEX, CWQMix, and AIQMix).

\addtolength{\tabcolsep}{-4pt}    
\begin{sidewaystable*}[t!]
\centering
\vspace{-1mm}
\captionsetup{justification=centering}
\begin{tabular}{ccccccccccc}
\toprule
             Map & Map Difficulty &  MAPPO(FP) &  MAPPO(AS) &       IPPO &       QMix &   QMix(MG) &        RODE &      QPLEX &     CWQMix &     AIQMix \\
\midrule
        2m\_vs\_1z &           Easy & 100.0(0.0) & 100.0(0.0) & 100.0(0.0) &  95.3(5.2) &  96.9(4.5) &           / &          / &          / &          / \\
              3m &           Easy & 100.0(0.0) & 100.0(1.5) & 100.0(0.0) &  96.9(1.3) &  96.9(1.7) &           / &          / &          / &          / \\
       2s\_vs\_1sc &           Easy & 100.0(0.0) & 100.0(0.0) & 100.0(1.5) &  96.9(2.9) & 100.0(1.4) &    100(0.0) &  98.4(1.6) &   100(0.0) &   100(0.0) \\
            2s3z &           Easy & 100.0(0.7) & 100.0(1.5) & 100.0(0.0) &  95.3(2.5) &  96.1(2.1) &    100(0.0) &   100(4.3) &  93.7(2.2) &  96.9(0.7) \\
        3s\_vs\_3z &           Easy & 100.0(0.0) & 100.0(0.0) & 100.0(0.0) & 96.9(12.5) &  96.9(3.7) &           / &          / &          / &          / \\
        3s\_vs\_4z &           Easy & 100.0(1.3) &  98.4(1.6) &  99.2(1.5) &  97.7(1.9) &  97.7(1.4) &           / &          / &          / &          / \\
so\_many\_baneling &           Easy & 100.0(0.0) & 100.0(0.7) & 100.0(1.5) &  96.9(2.3) &  92.2(5.8) &           / &          / &          / &          / \\
              8m &           Easy & 100.0(0.0) & 100.0(0.0) & 100.0(0.7) &  97.7(1.9) &  96.9(2.0) &           / &          / &          / &          / \\
             MMM &           Easy &  96.9(2.6) &  93.8(1.5) &  96.9(0.0) &  95.3(2.5) & 100.0(0.0) &           / &          / &          / &          / \\
          1c3s5z &           Easy & 100.0(0.0) &  96.9(2.6) & 100.0(0.0) &  96.1(1.7) & 100.0(0.5) &    100(0.0) &  96.8(1.6) &  96.9(1.4) & 92.2(10.4) \\
    bane\_vs\_bane &           Easy & 100.0(0.0) & 100.0(0.0) & 100.0(0.0) & 100.0(0.9) & 100.0(2.1) &   100(46.4) &   100(2.9) &   100(0.0) & 85.9(34.7) \\
        3s\_vs\_5z &           Hard & 100.0(0.6) &  99.2(1.4) & 100.0(0.0) &  98.4(2.4) &  98.4(1.6) &   78.9(4.2) &  98.4(1.4) &  34.4(6.5) & 82.8(10.6) \\
      2c\_vs\_64zg &           Hard & 100.0(0.0) & 100.0(0.0) &  98.4(1.3) &  92.2(4.0) &  95.3(1.5) &    100(0.0) &  90.6(7.3) &  85.9(3.3) &  97.6(2.3) \\
        8m\_vs\_9m &           Hard &  96.9(0.6) &  96.9(0.6) &  96.9(0.7) &  92.2(2.0) &  93.8(2.7) &           / &          / &          / &          / \\
             25m &           Hard & 100.0(1.5) & 100.0(4.0) & 100.0(0.0) &  85.9(7.1) &  96.9(3.8) &           / &          / &          / &          / \\
        5m\_vs\_6m &           Hard &  89.1(2.5) &  88.3(1.2) &  87.5(2.3) &  75.8(3.7) &  76.6(2.6) &   71.1(9.2) &  70.3(3.2) &  57.8(9.1) &  64.1(5.5) \\
            3s5z &           Hard &  96.9(0.7) &  96.9(1.9) &  96.9(1.5) &  88.3(2.9) &  92.2(1.8) & 93.75(1.95) &  96.8(2.2) & 70.3(20.3) &  96.9(2.9) \\
      10m\_vs\_11m &           Hard &  96.9(4.8) &  96.9(1.2) &  93.0(7.4) &  95.3(1.0) &  92.2(2.0) &   95.3(2.2) &  96.1(8.7) &  75.0(3.3) &  96.9(1.4) \\
            MMM2 &     Super Hard &  90.6(2.8) &  87.5(5.1) &  86.7(7.3) &  87.5(2.6) &  88.3(2.2) &   89.8(6.7) & 82.8(20.8) &   0.0(0.0) & 67.2(12.4) \\
    3s5z\_vs\_3s6z &     Super Hard & 84.4(34.0) & 63.3(19.2) & 82.8(19.1) &  82.8(5.3) &  82.0(4.4) & 96.8(25.11) & 10.2(11.0) & 53.1(12.9) &   0.0(0.0) \\
      27m\_vs\_30m &     Super Hard &  93.8(2.4) &  85.9(3.8) & 69.5(11.8) &  39.1(9.8) &  39.1(9.8) &   96.8(1.5) & 43.7(18.7) &  82.8(7.8) & 62.5(34.3) \\
        6h\_vs\_8z &     Super Hard &  88.3(3.7) & 85.9(30.9) & 84.4(33.3) &   9.4(2.0) &  39.8(4.0) &  78.1(37.0) &  1.5(31.0) & 49.2(14.8) &   0.0(0.0) \\
        corridor &     Super Hard & 100.0(1.2) &  98.4(0.8) &  98.4(3.1) &  84.4(2.5) &  81.2(5.9) &  65.6(32.1) &   0.0(0.0) &   0.0(0.0) &  12.5(7.6) \\
\bottomrule
\end{tabular}
\centering
\caption{Median evaluation win rate and standard deviation on all the SMAC maps for different methods, using at most 10M training timesteps.
} 
\label{tab:app-SMAC-results-all}
\vspace{-4mm}
\end{sidewaystable*}

\begin{sidewaystable*}[t!]
\centering
\vspace{-1mm}
\captionsetup{justification=centering}
\begin{tabular}{ccccccccccc}
\toprule
             Map & Map Difficulty & MAPPO(FP)* & MAPPO(AS)* &     IPPO* &     QMix* & QMix(MG)* &        RODE &      QPLEX &     CWQMix &     AIQMix \\
\midrule
        2m\_vs\_1z &           Easy &  100.0(0.0) &  100.0(0.0) & 100.0(0.0) &  96.9(2.8) &  96.9(4.7) &           / &          / &          / &          / \\
              3m &           Easy &  100.0(0.0) &  100.0(1.5) & 100.0(0.0) &  92.2(2.7) &  96.9(2.1) &           / &          / &          / &          / \\
       2s\_vs\_1sc &           Easy &  100.0(0.0) &  100.0(0.0) & 100.0(0.0) &  96.9(1.2) &  96.9(4.6) &    100(0.0) &  98.4(1.6) &   100(0.0) &   100(0.0) \\
            2s3z &           Easy &   96.9(1.5) &   96.9(1.5) & 100.0(0.0) &  95.3(3.9) &  92.2(2.3) &    100(0.0) &   100(4.3) &  93.7(2.2) &  96.9(0.7) \\
        3s\_vs\_3z &           Easy &  100.0(0.0) &  100.0(0.0) & 100.0(0.0) & 100.0(1.5) & 100.0(1.5) &           / &          / &          / &          / \\
        3s\_vs\_4z &           Easy &  100.0(2.1) &  100.0(1.5) & 100.0(1.4) &  87.5(3.2) &  98.4(0.8) &           / &          / &          / &          / \\
so\_many\_baneling &           Easy &  100.0(1.5) &   96.9(1.5) &  96.9(1.5) &  81.2(7.2) &  78.1(6.7) &           / &          / &          / &          / \\
              8m &           Easy &  100.0(0.0) &  100.0(0.0) & 100.0(1.5) &  93.8(5.1) &  93.8(2.7) &           / &          / &          / &          / \\
             MMM &           Easy &   93.8(2.6) &   96.9(1.5) &  96.9(1.5) &  95.3(3.9) & 100.0(1.2) &           / &          / &          / &          / \\
          1c3s5z &           Easy &  100.0(0.0) &   96.9(2.6) &  93.8(5.1) &  95.3(1.2) &  98.4(1.4) &    100(0.0) &  96.8(1.6) &  96.9(1.4) & 92.2(10.4) \\
    bane\_vs\_bane &           Easy &  100.0(0.0) &  100.0(0.0) & 100.0(0.0) & 100.0(0.0) & 100.0(0.0) &   100(46.4) &   100(2.9) &   100(0.0) & 85.9(34.7) \\
        3s\_vs\_5z &           Hard &   98.4(5.5) &  100.0(1.2) & 100.0(2.4) &  56.2(8.8) &  90.6(2.2) &   78.9(4.2) &  98.4(1.4) &  34.4(6.5) & 82.8(10.6) \\
      2c\_vs\_64zg &           Hard &   96.9(3.1) &   95.3(3.5) &  93.8(9.2) &  70.3(3.8) &  84.4(3.7) &    100(0.0) &  90.6(7.3) &  85.9(3.3) &  97.6(2.3) \\
        8m\_vs\_9m &           Hard &   84.4(5.1) &   87.5(2.1) &  76.6(5.6) &  85.9(2.9) &  85.9(4.7) &           / &          / &          / &          / \\
             25m &           Hard &   96.9(3.1) &   93.8(2.9) &  93.8(5.0) &  96.9(4.0) &  93.8(5.7) &           / &          / &          / &          / \\
        5m\_vs\_6m &           Hard &  65.6(14.1) &   68.8(8.2) &  64.1(7.7) &  54.7(3.5) &  56.2(2.1) &   71.1(9.2) &  70.3(3.2) &  57.8(9.1) &  64.1(5.5) \\
            3s5z &           Hard &  71.9(11.8) &  53.1(15.4) & 84.4(12.1) &  85.9(4.6) &  89.1(2.6) & 93.75(1.95) &  96.8(2.2) & 70.3(20.3) &  96.9(2.9) \\
      10m\_vs\_11m &           Hard &   81.2(8.3) &   89.1(5.5) & 87.5(17.5) &  82.8(4.1) &  85.9(2.3) &   95.3(2.2) &  96.1(8.7) &  75.0(3.3) &  96.9(1.4) \\
            MMM2 &     Super Hard &  51.6(21.9) &  28.1(29.6) & 26.6(27.8) &  82.8(4.0) &  79.7(3.4) &   89.8(6.7) & 82.8(20.8) &   0.0(0.0) & 67.2(12.4) \\
    3s5z\_vs\_3s6z &     Super Hard &  75.0(36.3) &  18.8(37.4) & 65.6(25.9) & 56.2(11.3) &  39.1(4.7) & 96.8(25.11) & 10.2(11.0) & 53.1(12.9) &   0.0(0.0) \\
      27m\_vs\_30m &     Super Hard &   93.8(3.8) &   89.1(6.5) & 73.4(11.5) &  34.4(5.4) &  34.4(5.4) &   96.8(1.5) & 43.7(18.7) &  82.8(7.8) & 62.5(34.3) \\
        6h\_vs\_8z &     Super Hard &   78.1(5.6) &  81.2(31.8) & 78.1(33.1) &   3.1(1.5) &  29.7(6.3) &  78.1(37.0) &  1.5(31.0) & 49.2(14.8) &   0.0(0.0) \\
        corridor &     Super Hard &   93.8(3.5) &   93.8(2.8) &  89.1(9.1) & 64.1(14.3) &  81.2(1.5) &  65.6(32.1) &   0.0(0.0) &   0.0(0.0) &  12.5(7.6) \\
\bottomrule
\end{tabular}
\centering
\caption{Median evaluation win rate and standard deviation on all the SMAC maps for different methods, Columns with ``*'' display results using the same number of timesteps as RODE. 
} 
\label{tab:app-SMAC-results-cut}
\vspace{-4mm}
\end{sidewaystable*}

\subsection{Additional GRF Results}

\begin{figure*}[ht]
	\centering
	\subfigure
	{
        {
        \includegraphics[width=0.25\textwidth]{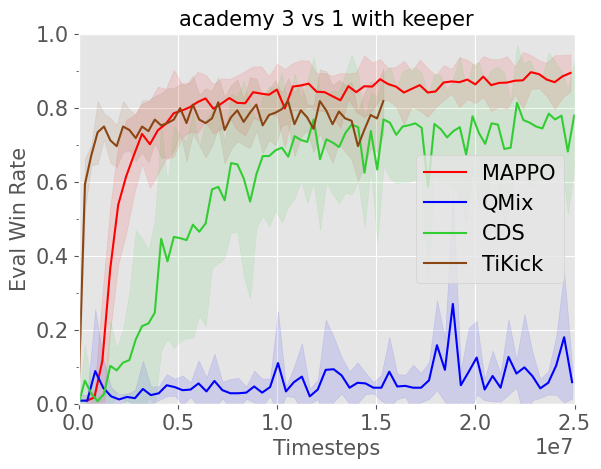}
        \includegraphics[width=0.25\textwidth]{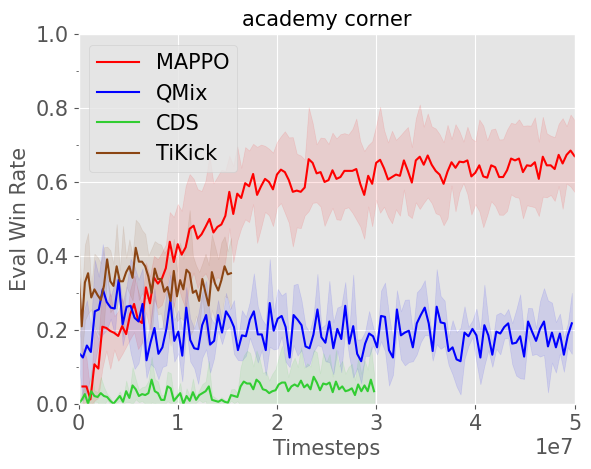}
        \includegraphics[width=0.25\textwidth]{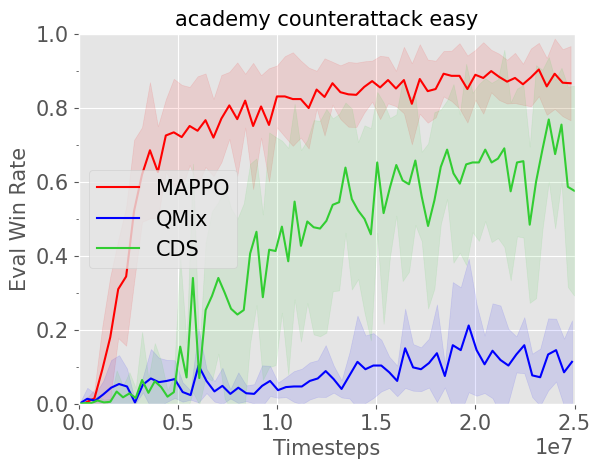}
        \includegraphics[width=0.25\textwidth]{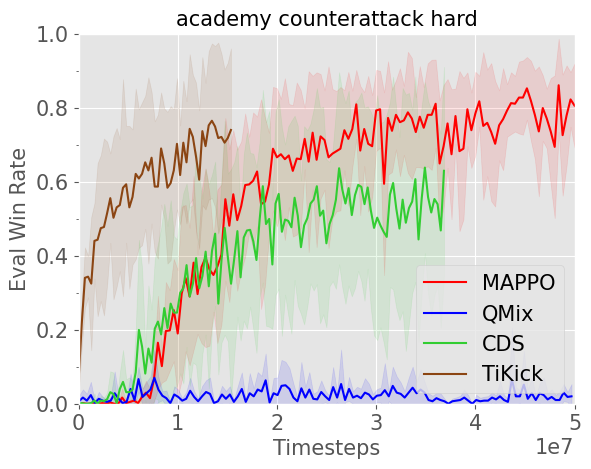}
    	}
    }
    \subfigure
    {
        \includegraphics[width=0.25\textwidth]{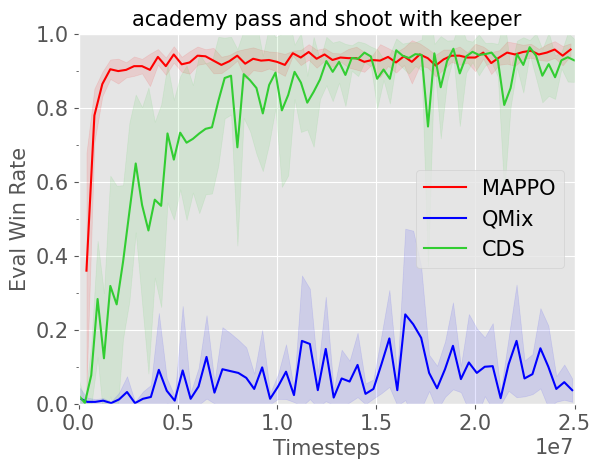}
        \includegraphics[width=0.25\textwidth]{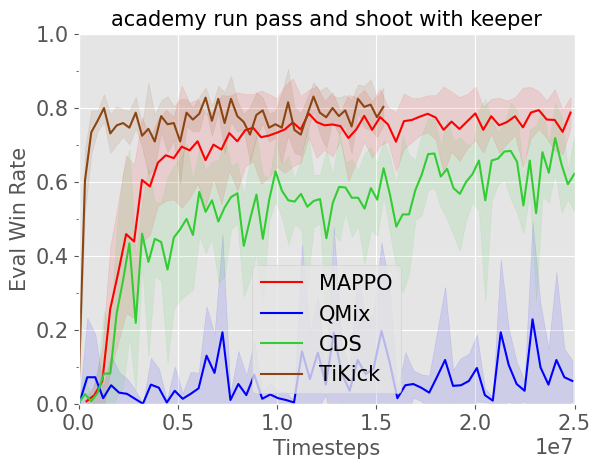}
        
    }
	\centering \caption{Mean evaluation win rate of MAPPO, QMix, CDS, TiKick in the GRF domain.}
\label{fig:app-football}
\end{figure*}

Fig. \ref{fig:app-football} compares the results of MAPPO to various baselines, including QMix, CDS, and TiKick, in 6 academy scenarios.

\section{Ablation Studies}
\label{app:ablation-studies}





\begin{figure*}[ht]
\captionsetup{justification=centering}

	\centering
    \subfigure
	{\centering
        {\includegraphics[width=0.25\textwidth]{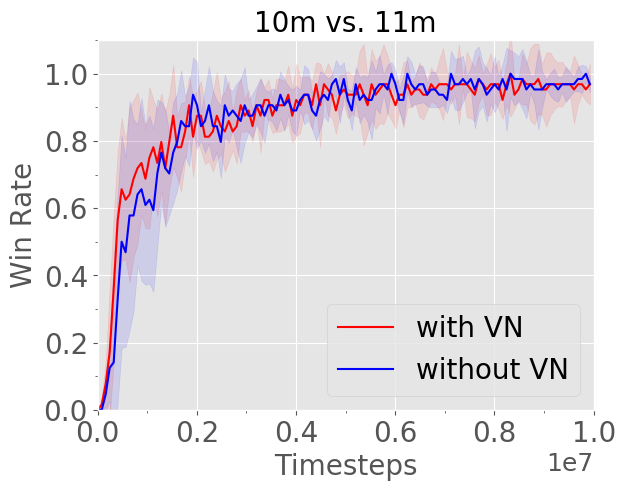}
        \includegraphics[width=0.25\textwidth]{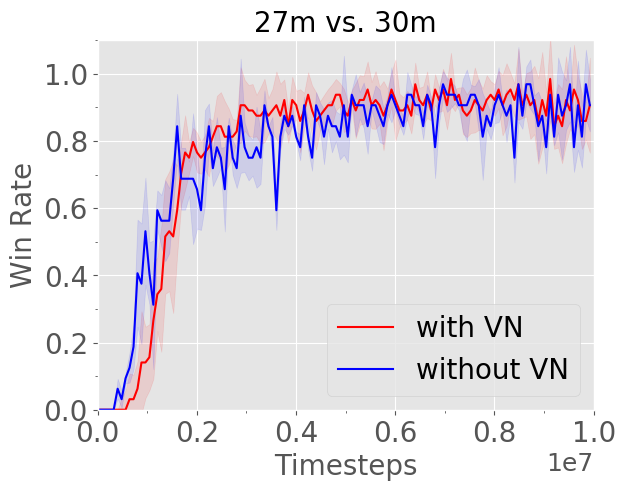}
        \includegraphics[width=0.25\textwidth]{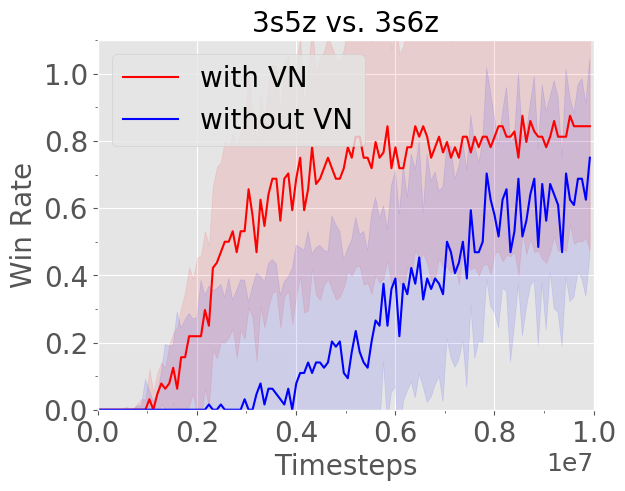}
        \includegraphics[width=0.25\textwidth]{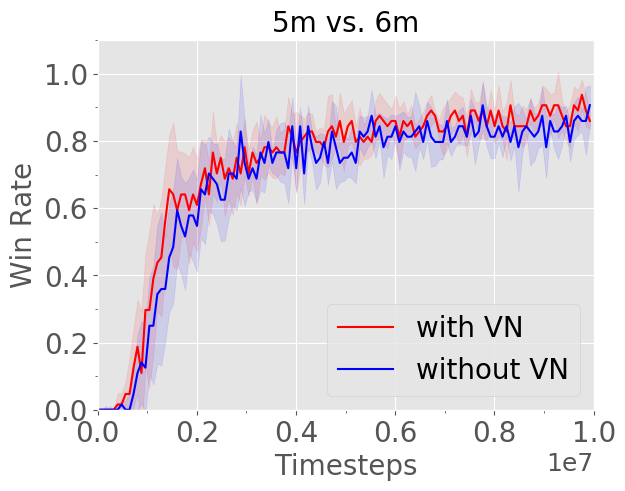}
        
    	}
    }
    \subfigure
	{\centering
	    \includegraphics[width=0.25\textwidth]{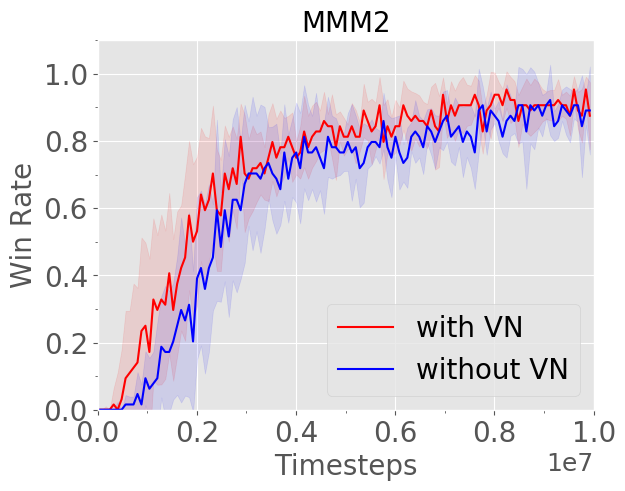}
        \includegraphics[width=0.25\textwidth]{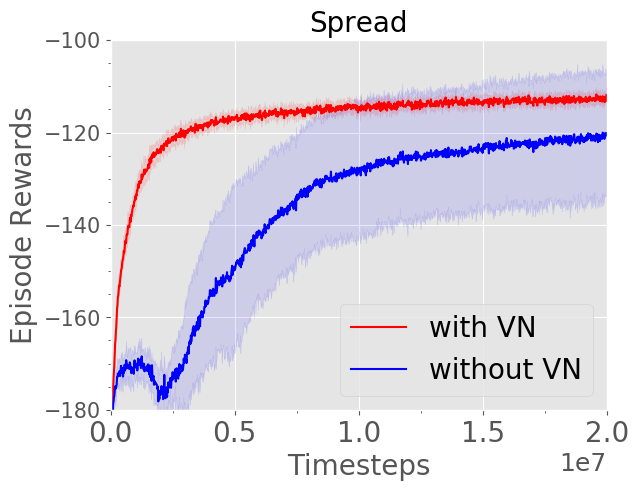}
        \includegraphics[width=0.25\textwidth]{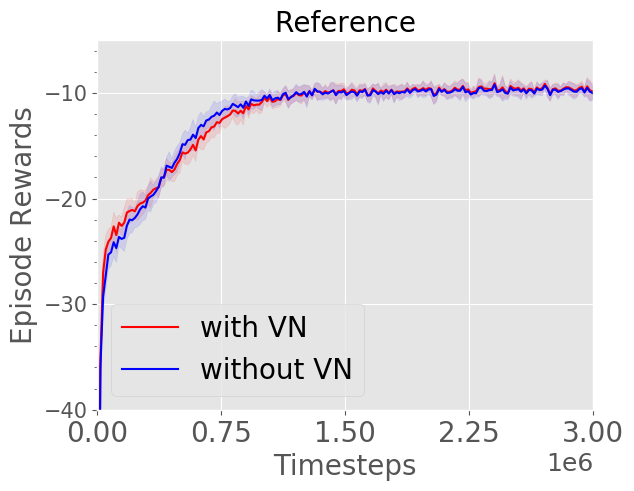}
        \includegraphics[width=0.25\textwidth]{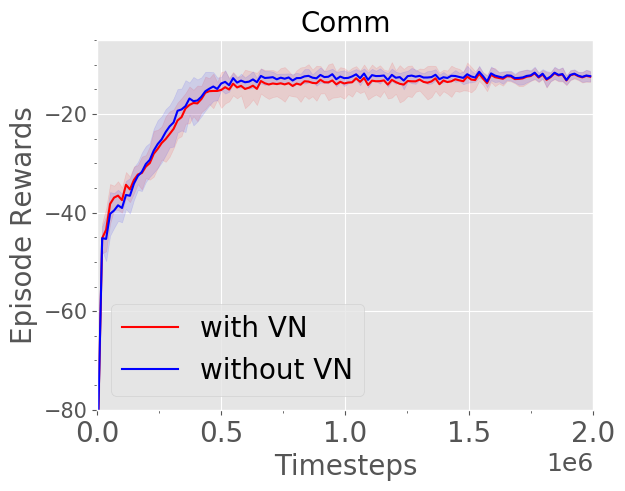}}
    \subfigure
    {\centering
        \includegraphics[width=0.25\textwidth]{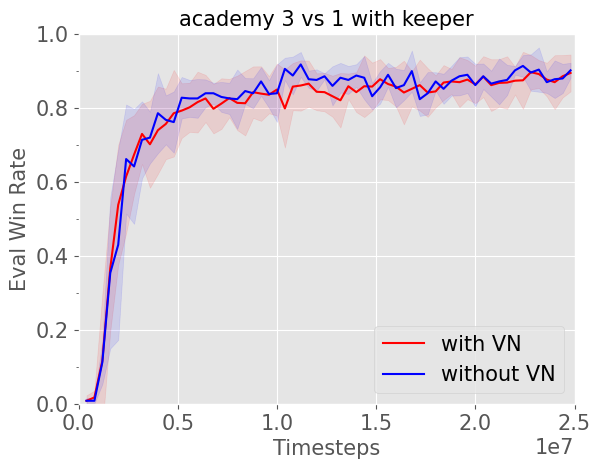}
        \includegraphics[width=0.25\textwidth]{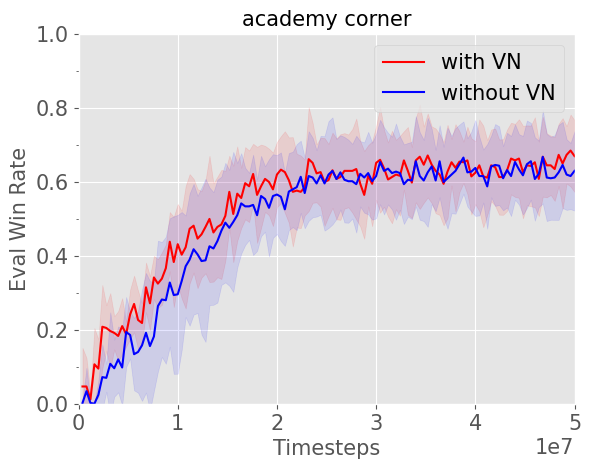}
        \includegraphics[width=0.25\textwidth]{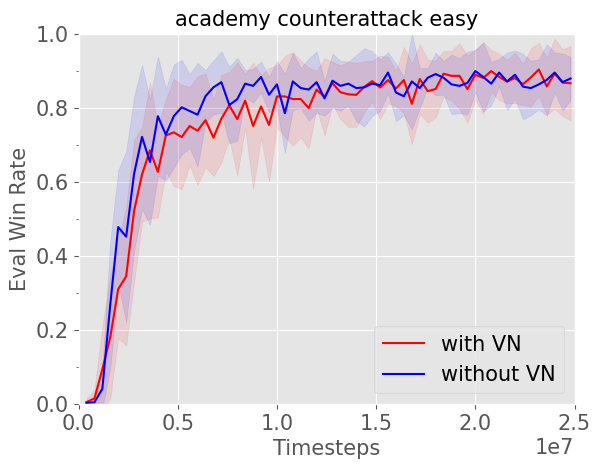}
        \includegraphics[width=0.25\textwidth]{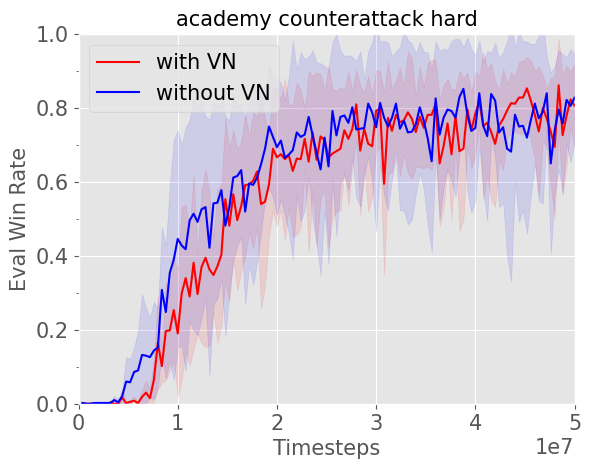}
        }
    \subfigure
    {
        \includegraphics[width=0.25\textwidth]{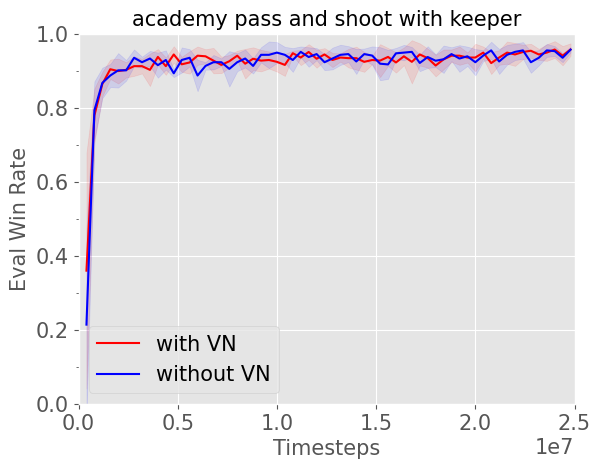}
        \includegraphics[width=0.25\textwidth]{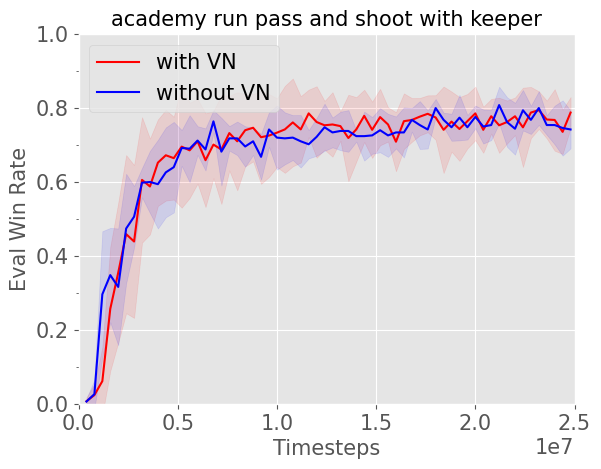}
    }
	\centering \caption{Ablation studies demonstrating the effect of Value Normalization(VN) on MAPPO's performance in the MPE, SMAC, and GRF domains.}
\label{fig:app-Ablation-popart}
\end{figure*}

\begin{figure*}[ht]
\captionsetup{justification=centering}

	\centering
    \subfigure
	{
        \includegraphics[width=0.25\textwidth]{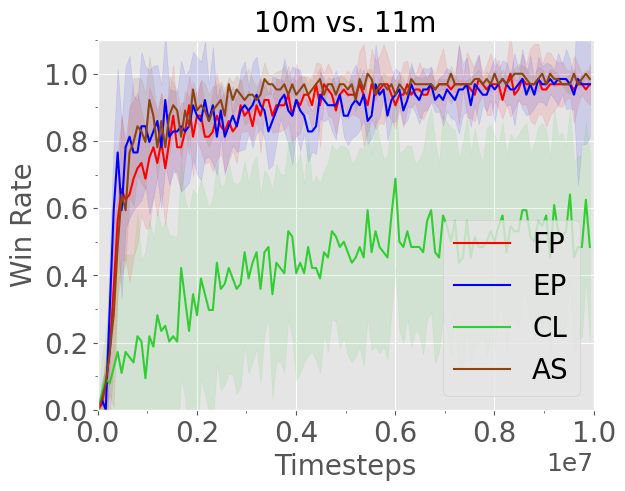}
        \includegraphics[width=0.25\textwidth]{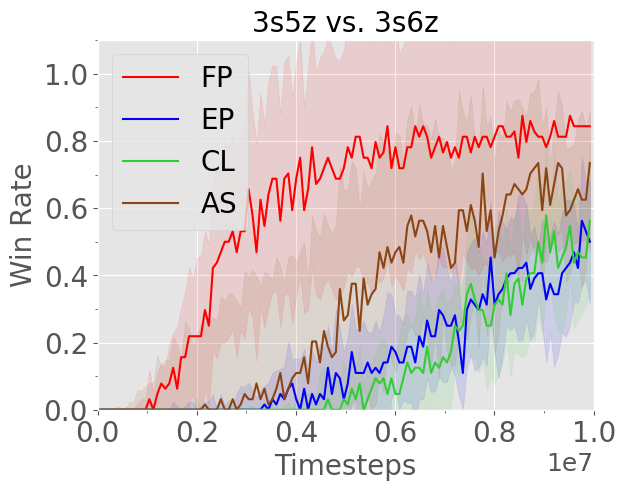}
        \includegraphics[width=0.25\textwidth]{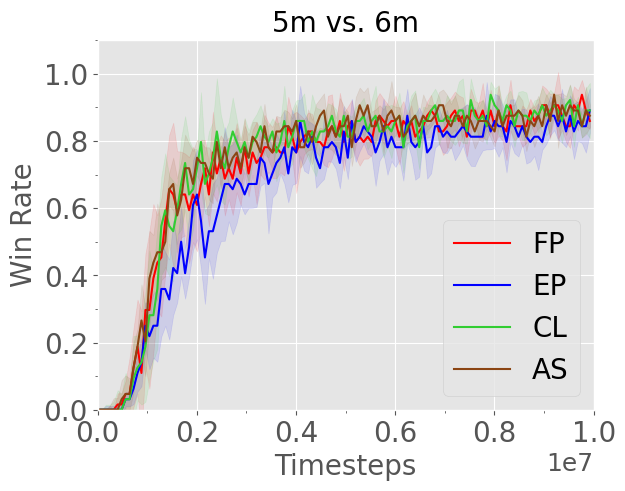}
    	\includegraphics[width=0.25\textwidth]{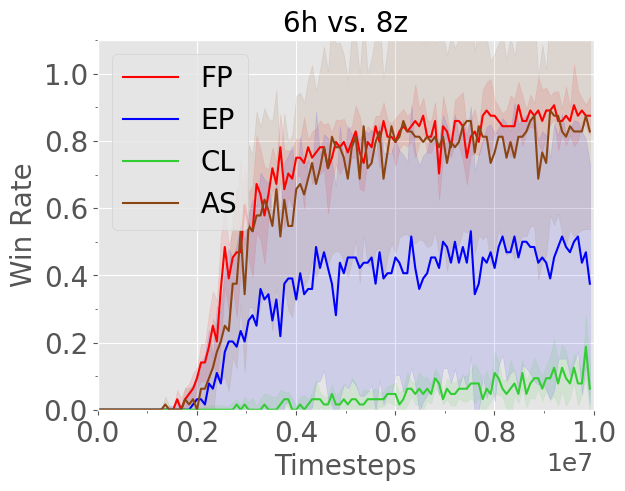}
    }
    \subfigure
	{
        \includegraphics[width=0.25\textwidth]{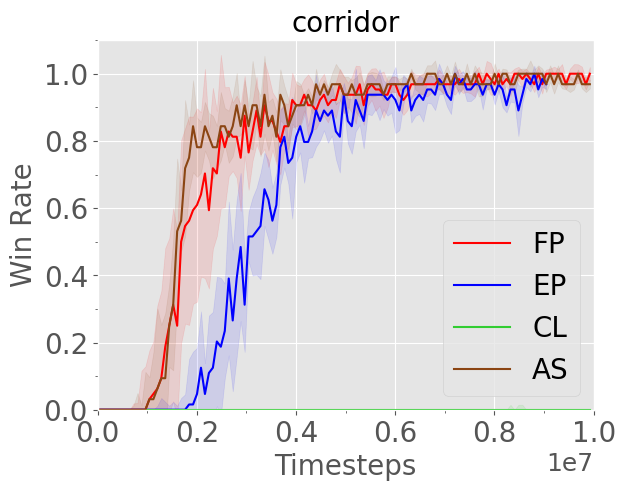}
        \includegraphics[width=0.25\textwidth]{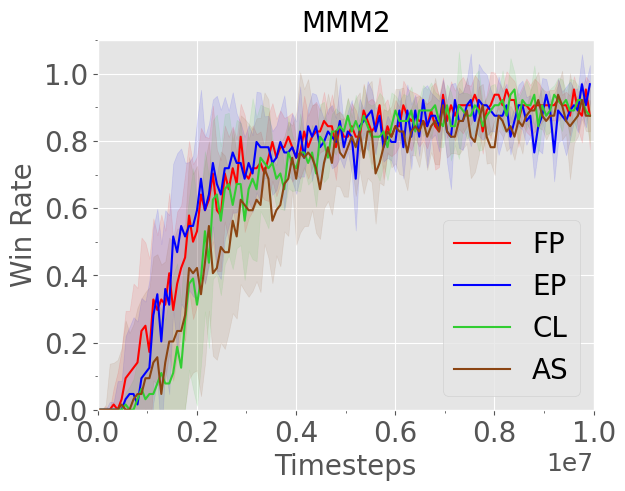}
        \includegraphics[width=0.25\textwidth]{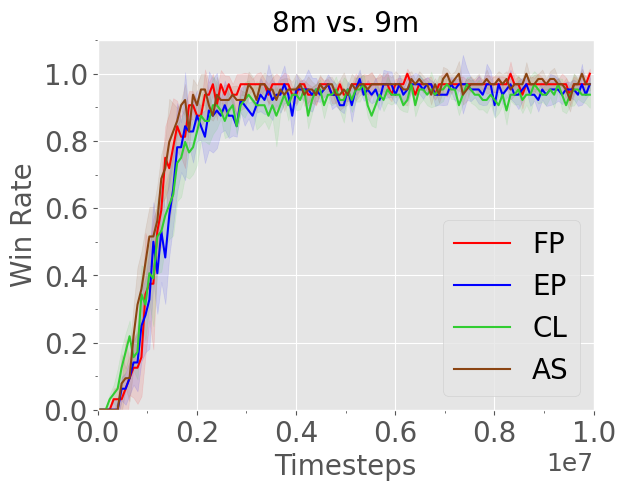}
    }
	\centering 
	\caption{Ablation studies demonstrating the effect of different global state on MAPPO's performance in the SMAC domain.}
\label{fig:app-Ablation-state}
\end{figure*}

\begin{figure*}[ht]
\captionsetup{justification=centering}

	\centering
	\subfigure
	{
        \includegraphics[width=0.25\textwidth]{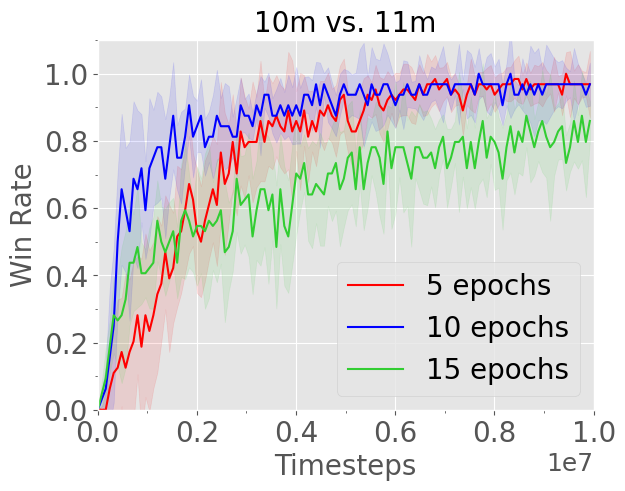}
        \includegraphics[width=0.25\textwidth]{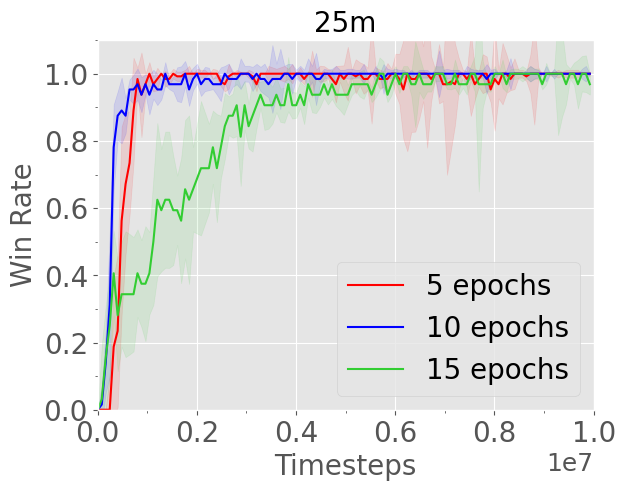}
        \includegraphics[width=0.25\textwidth]{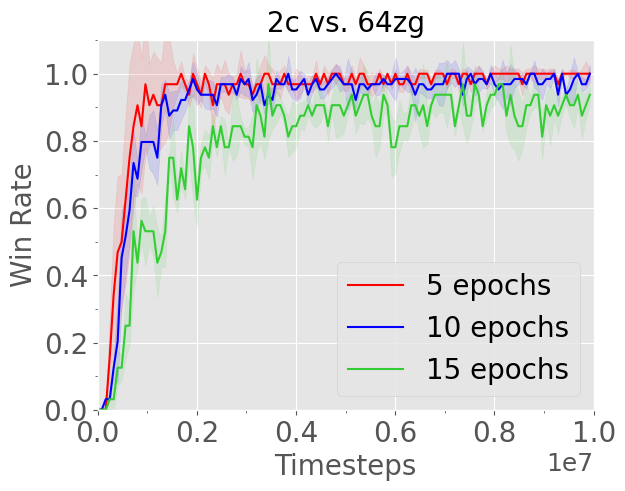}
        \includegraphics[width=0.25\textwidth]{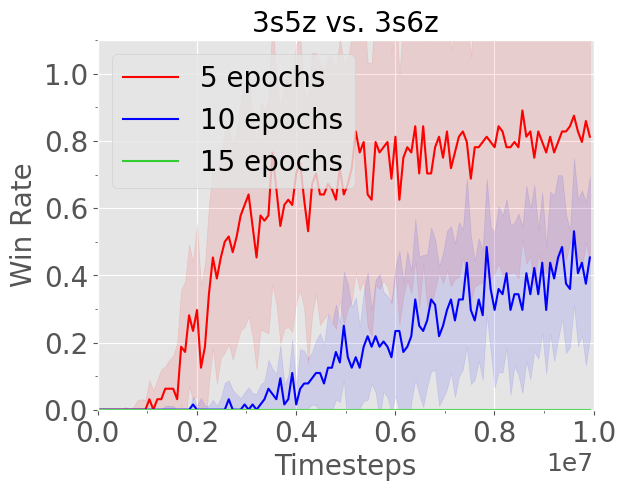}
    }
    \subfigure
    { 
        \includegraphics[width=0.25\textwidth]{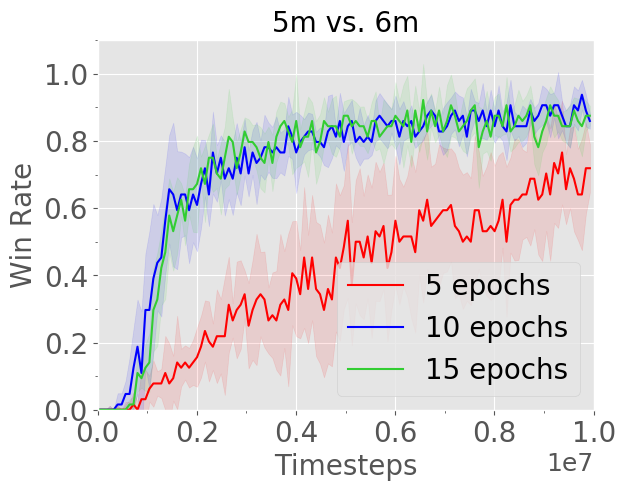}
        \includegraphics[width=0.25\textwidth]{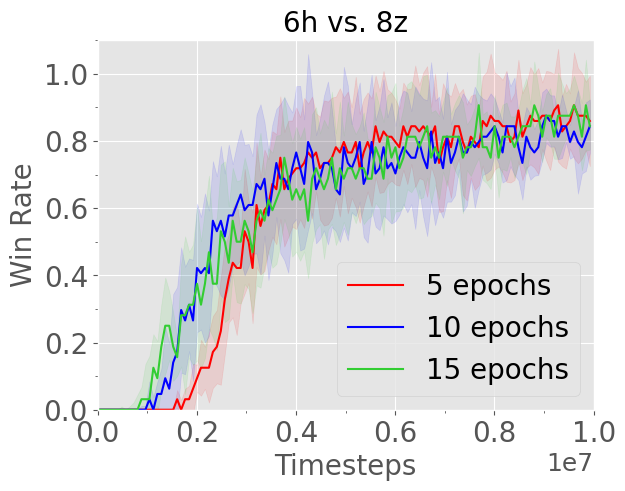}
        \includegraphics[width=0.25\textwidth]{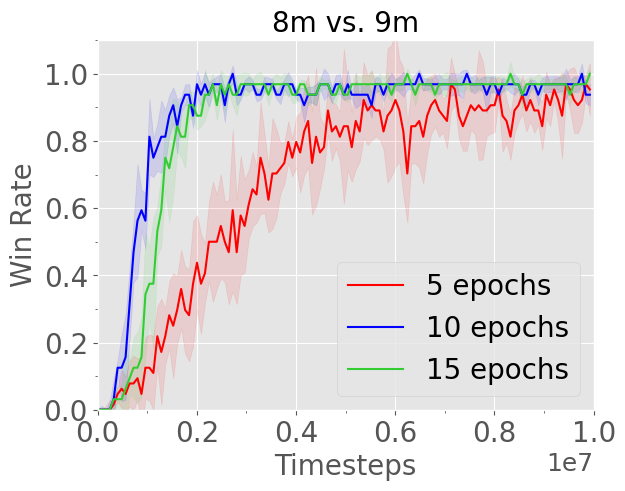}
        \includegraphics[width=0.25\textwidth]{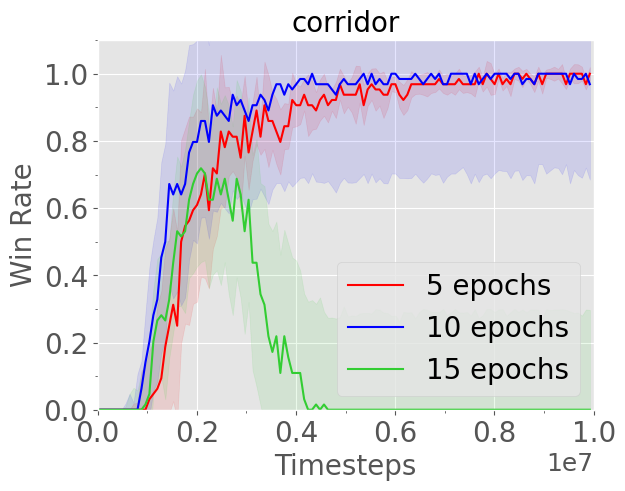}
     }
    \subfigure
	{\centering
        \includegraphics[width=0.25\textwidth]{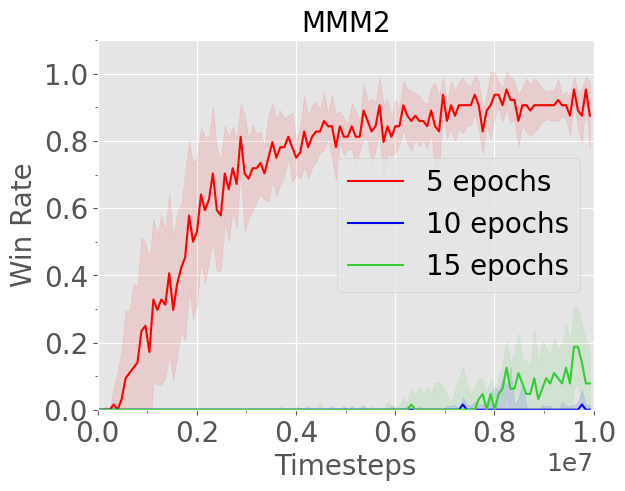}
        \includegraphics[width=0.25\textwidth]{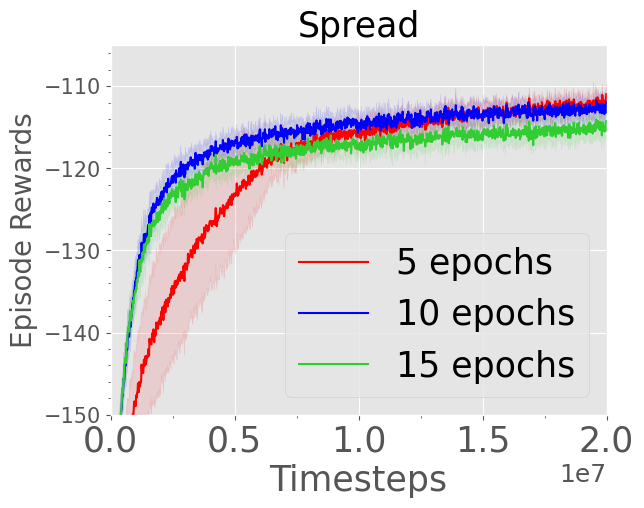}
        \includegraphics[width=0.25\textwidth]{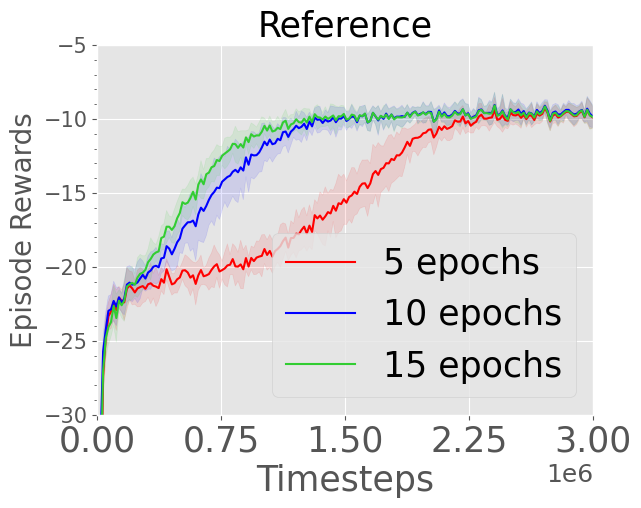}
        \includegraphics[width=0.25\textwidth]{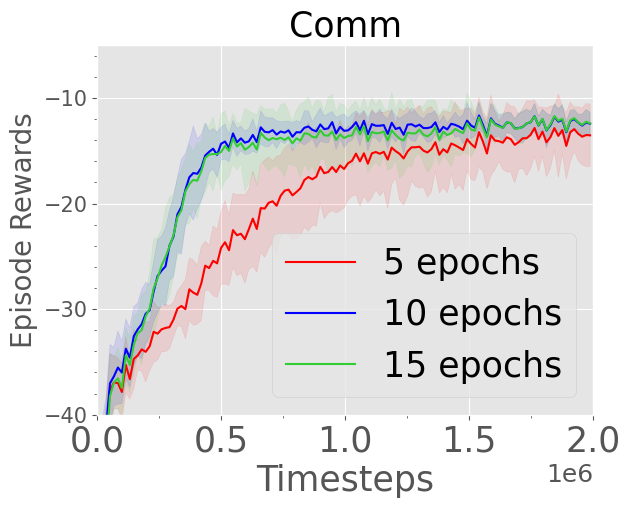}
    }
    \subfigure
	{\centering
        \includegraphics[width=0.25\textwidth]{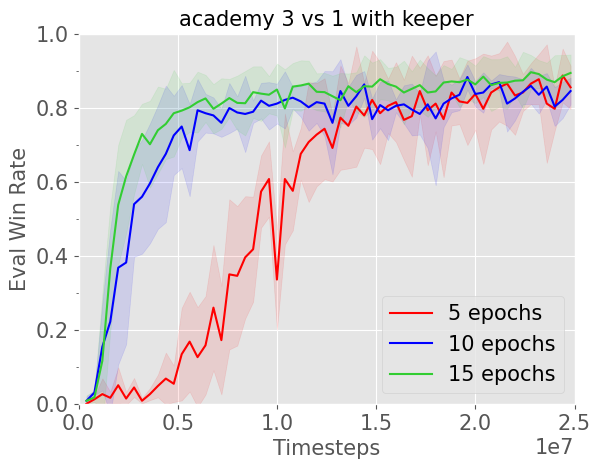}
        \includegraphics[width=0.25\textwidth]{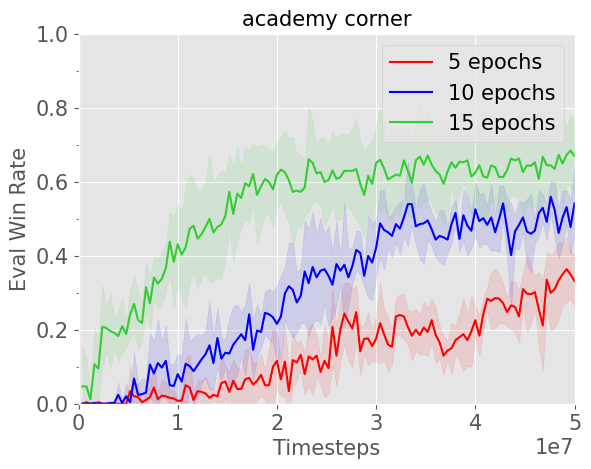}
        \includegraphics[width=0.25\textwidth]{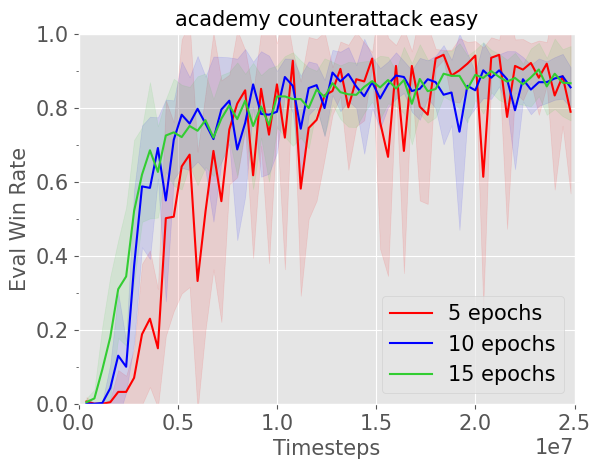}
        \includegraphics[width=0.25\textwidth]{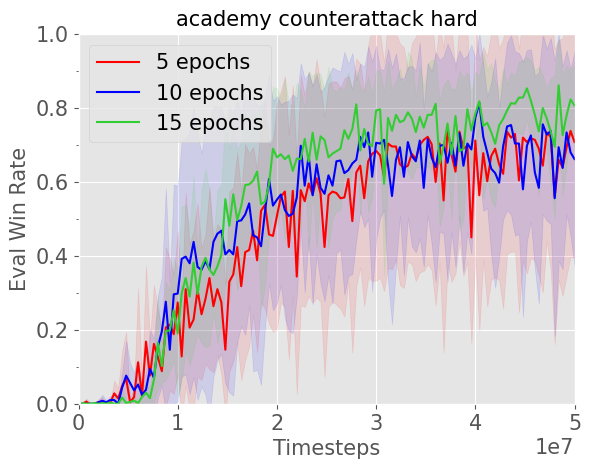}
    }
    \subfigure
	{\centering
        \includegraphics[width=0.25\textwidth]{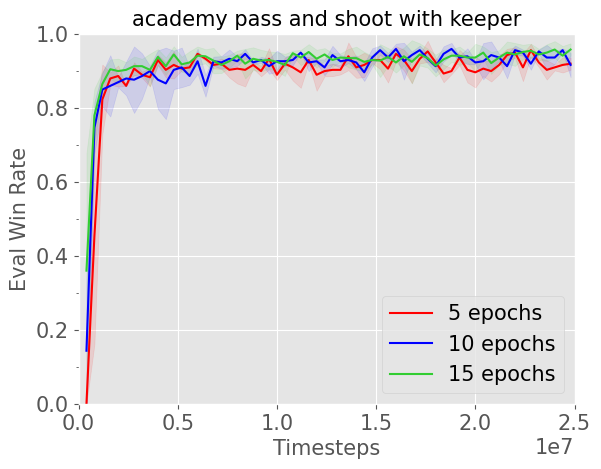}
        \includegraphics[width=0.25\textwidth]{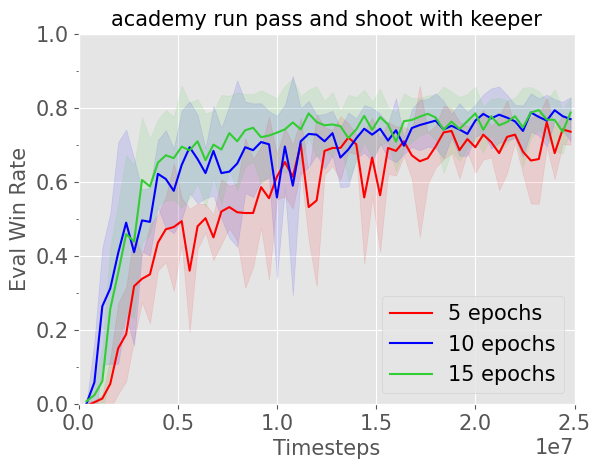}
    }
	\centering \caption{Ablation studies demonstrating the effect of training epochs on MAPPO's performance in the MPE, SMAC, and GRF domains.}
\label{fig:app-Ablation-epoch}
\end{figure*}

\begin{figure*}[ht]
\captionsetup{justification=centering}

	\centering
	\subfigure
	{
        \includegraphics[width=0.25\textwidth]{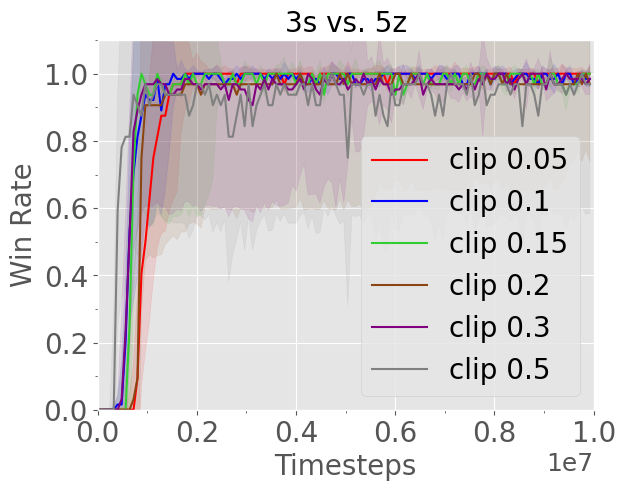}
        \includegraphics[width=0.25\textwidth]{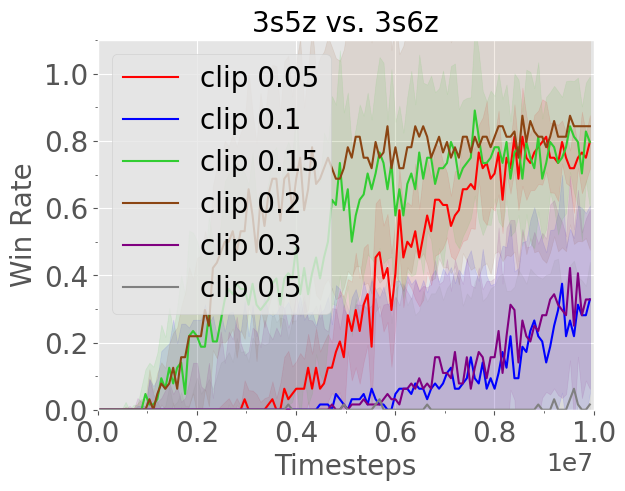}
        \includegraphics[width=0.25\textwidth]{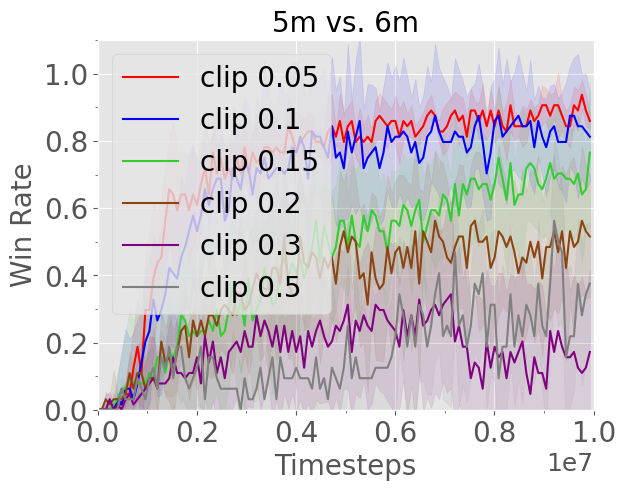}
        \includegraphics[width=0.25\textwidth]{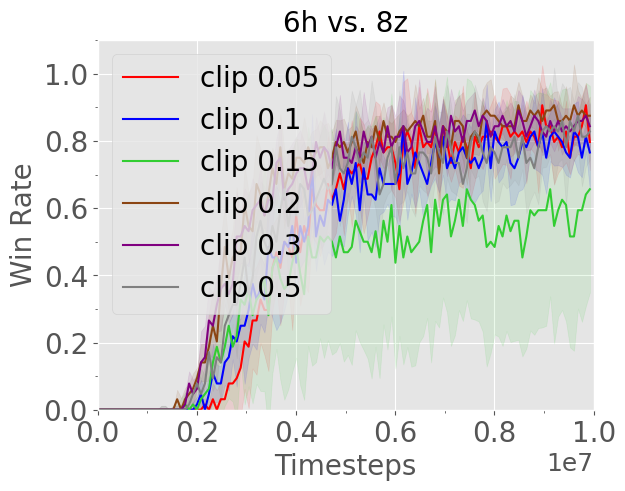}
    }
    \subfigure
    { 
        \includegraphics[width=0.25\textwidth]{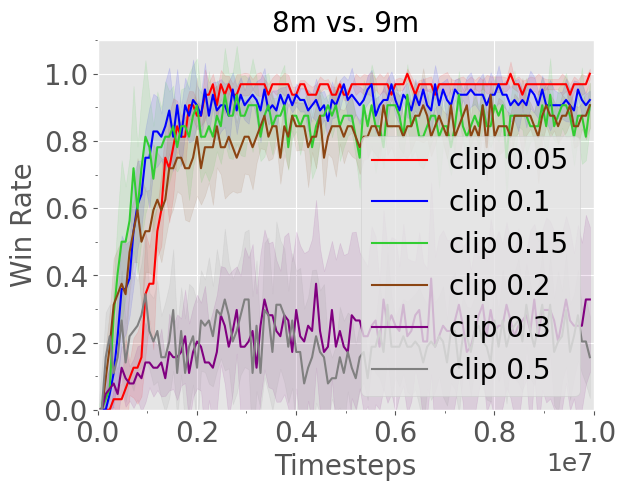}
        \includegraphics[width=0.25\textwidth]{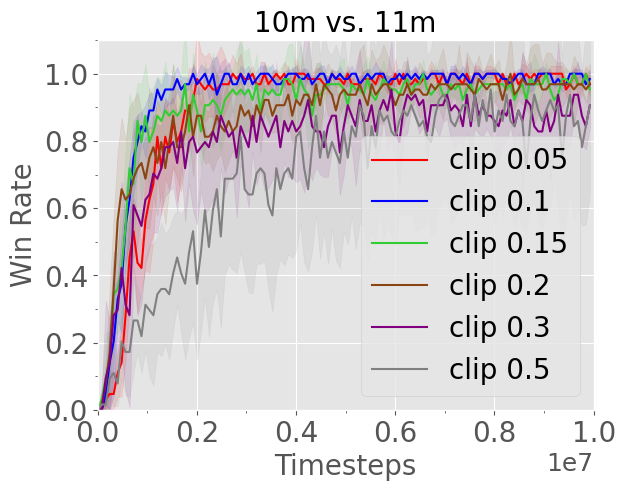}
        \includegraphics[width=0.25\textwidth]{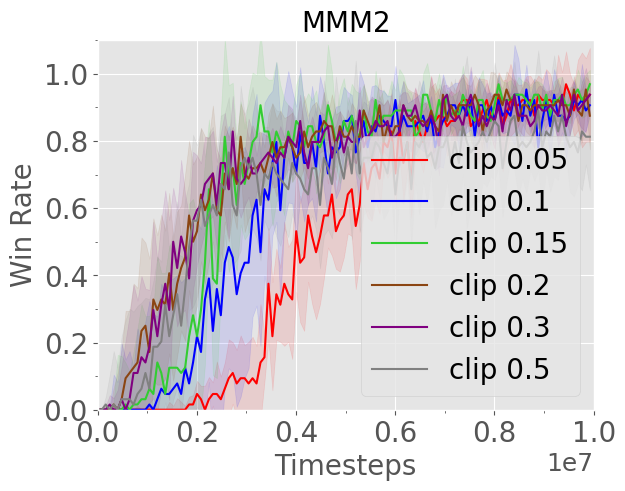}
        \includegraphics[width=0.25\textwidth]{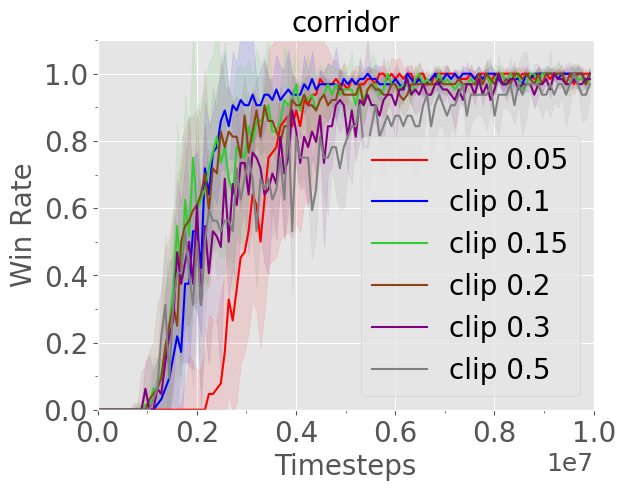}
     }
     \subfigure
     { 
        \includegraphics[width=0.25\textwidth]{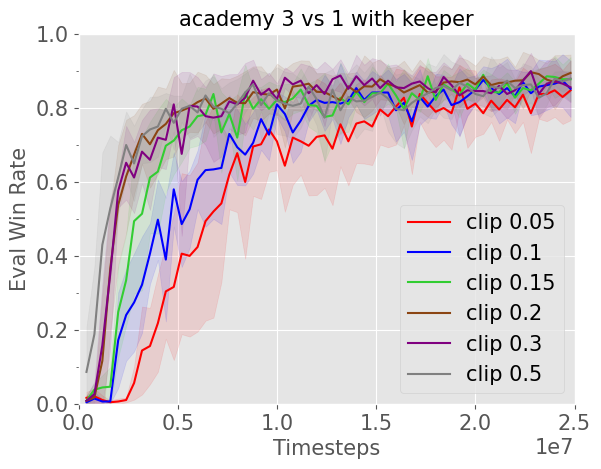}
        \includegraphics[width=0.25\textwidth]{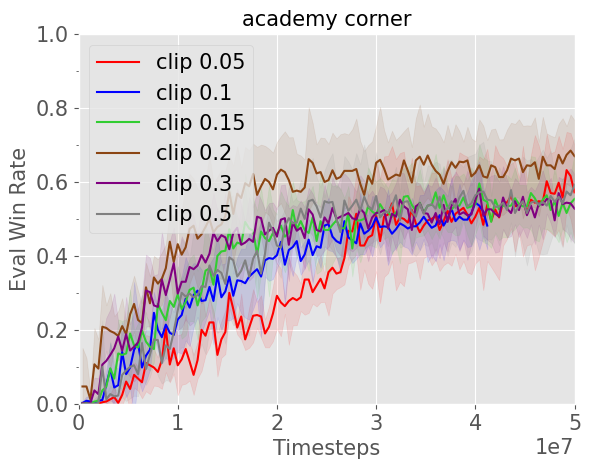}
        \includegraphics[width=0.25\textwidth]{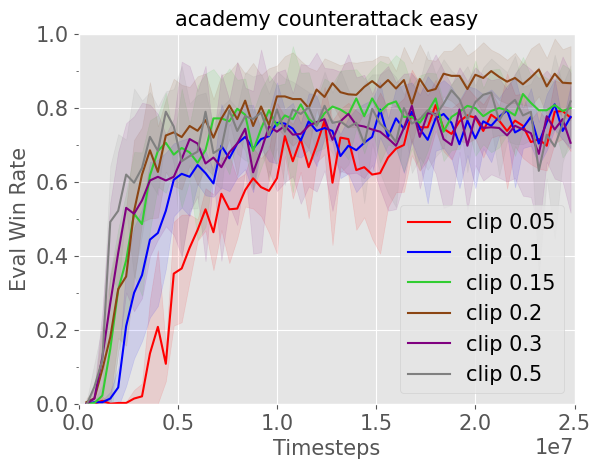}
        \includegraphics[width=0.25\textwidth]{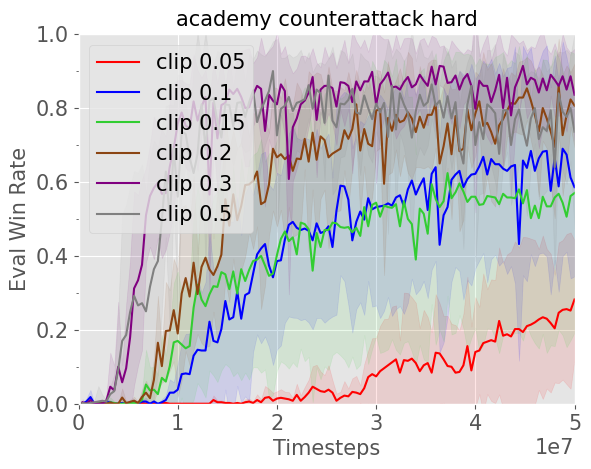}
     }
     \subfigure
    { 
        \includegraphics[width=0.25\textwidth]{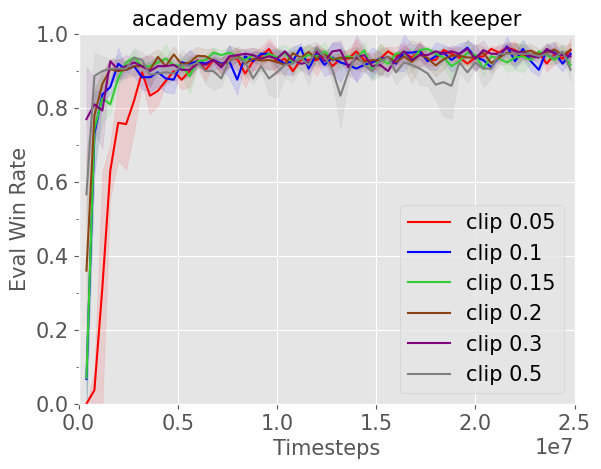}
        \includegraphics[width=0.25\textwidth]{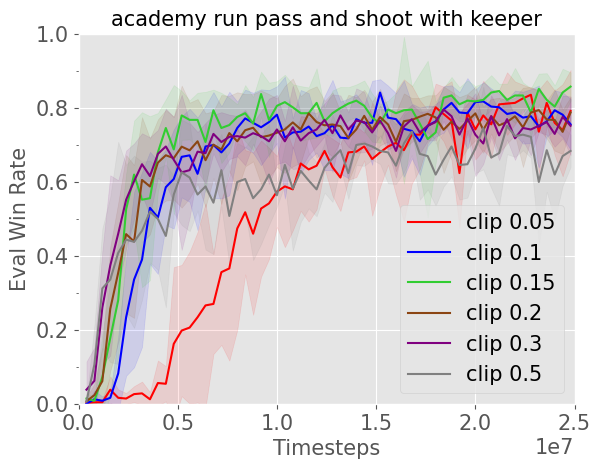}
     }
	\centering \caption{Ablation studies demonstrating the effect of clip term on MAPPO's performance in the SMAC and GRF domain.}
\label{fig:app-Ablation-clip}
\end{figure*}

\begin{figure*}[ht]
    \captionsetup{justification=centering}
	\centering
    \subfigure
	{\centering
        {\includegraphics[width=0.25\textwidth]{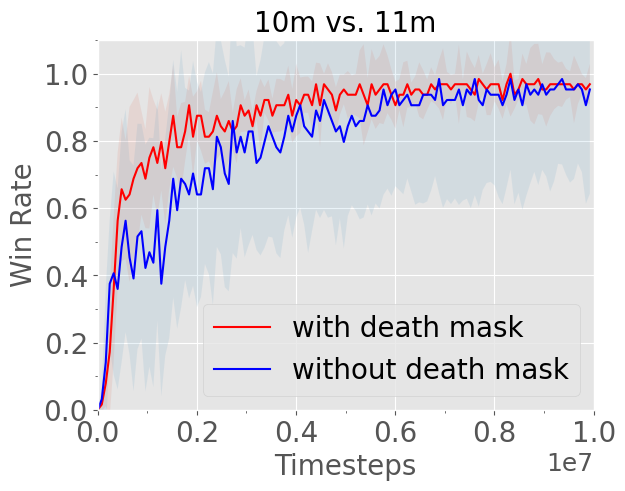}
        \includegraphics[width=0.25\textwidth]{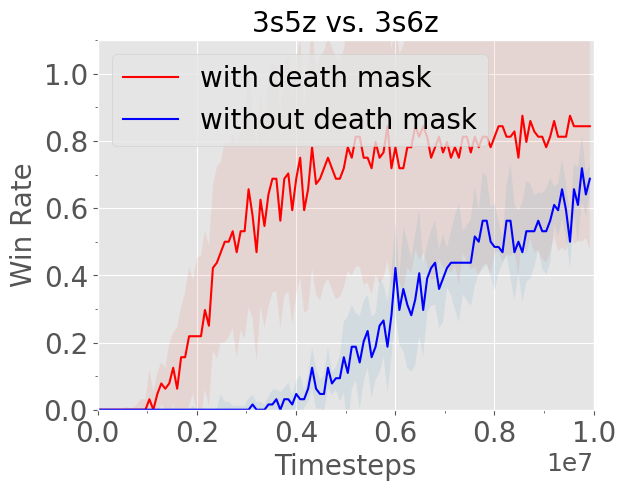}
        \includegraphics[width=0.25\textwidth]{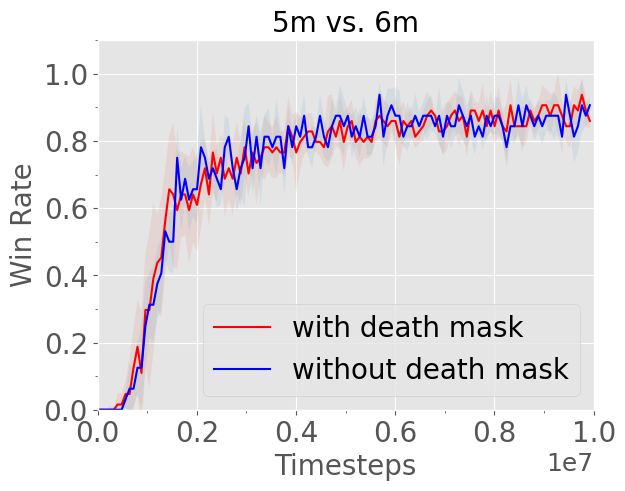}
        \includegraphics[width=0.25\textwidth]{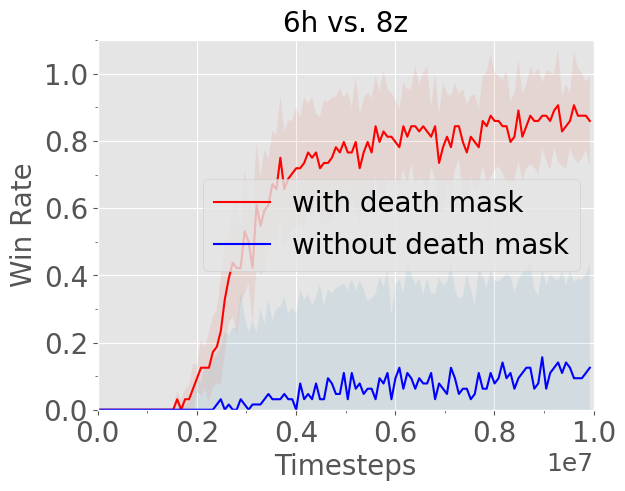}
    	}
    }
    \subfigure
	{\centering
        {
        \includegraphics[width=0.25\textwidth]{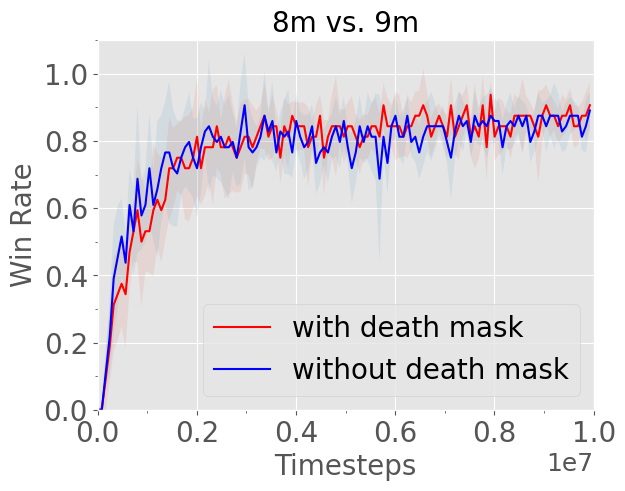}
        \includegraphics[width=0.25\textwidth]{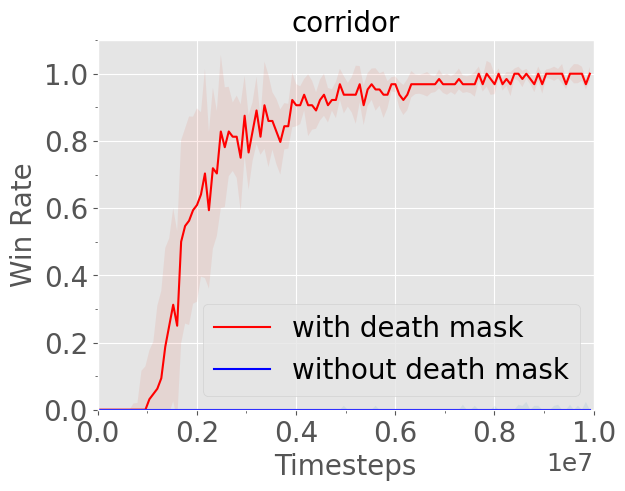}
        \includegraphics[width=0.25\textwidth]{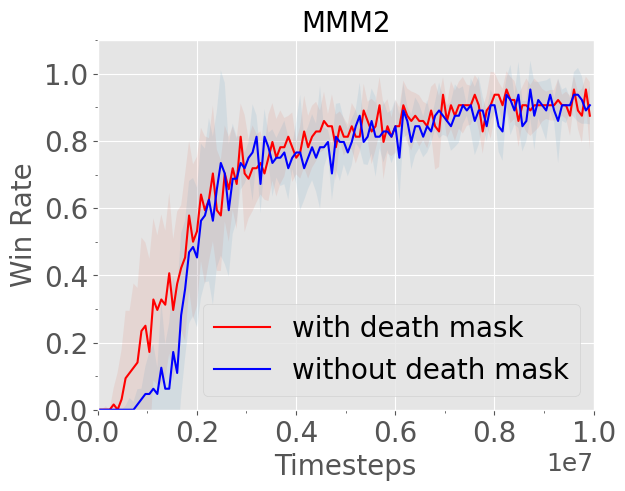}
    	}
    }
	\centering \caption{Ablation studies demonstrating the effect of death mask on MAPPO(FP)'s performance in the SMAC doamin.}
\label{fig:app-Ablation-death}
\end{figure*}

\begin{figure*}[ht]
    \captionsetup{justification=centering}
	\centering
    \subfigure
	{\centering
        {\includegraphics[width=0.25\textwidth]{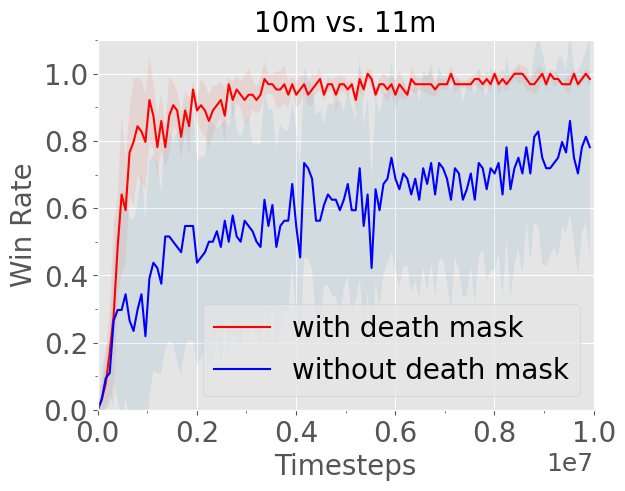}
        \includegraphics[width=0.25\textwidth]{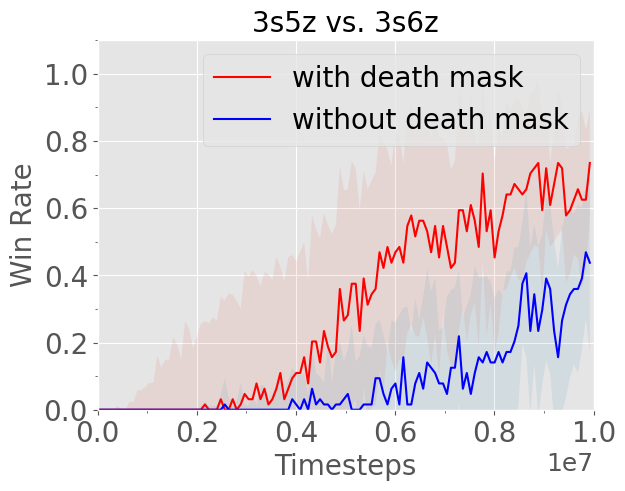}
        \includegraphics[width=0.25\textwidth]{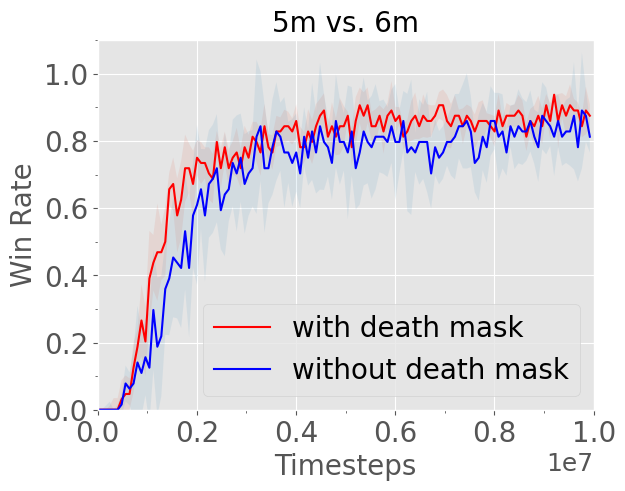}
        \includegraphics[width=0.25\textwidth]{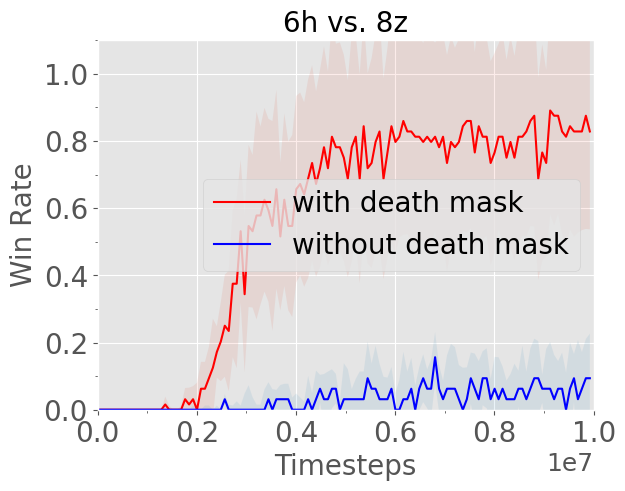}
    	}
    }
    \subfigure
	{\centering
        {
        \includegraphics[width=0.25\textwidth]{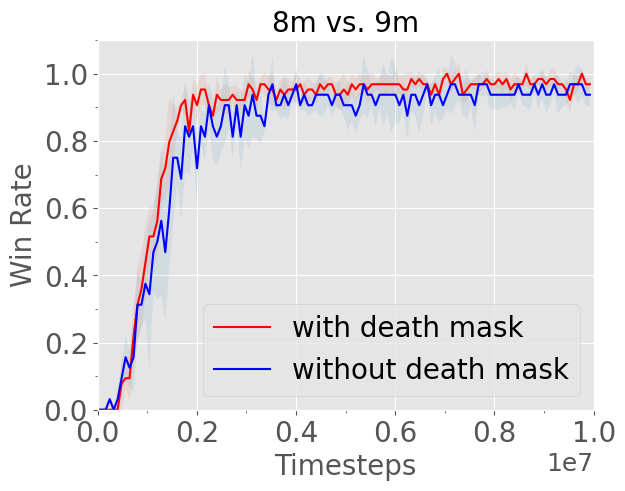}
        \includegraphics[width=0.25\textwidth]{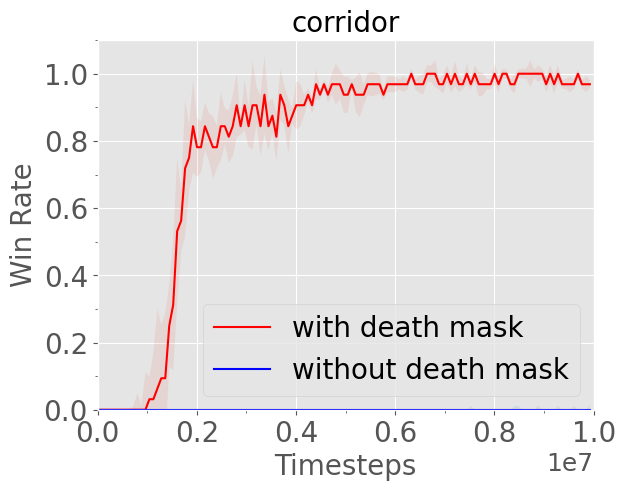}
        \includegraphics[width=0.25\textwidth]{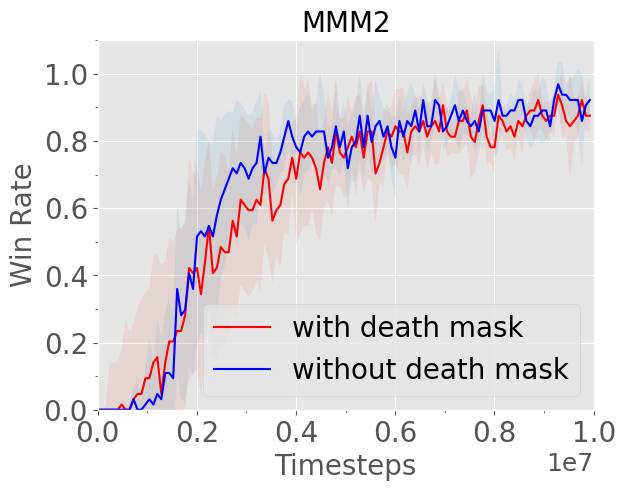}
    	}
    }
	\centering \caption{Ablation studies demonstrating the effect of death mask on MAPPO(AS)'s performance in the SMAC domain.}
\label{fig:app-Ablation-death-new}
\end{figure*}

\begin{figure*}[ht]
    \captionsetup{justification=centering}
	\centering
    \subfigure
	{\centering
        {\includegraphics[width=0.25\textwidth]{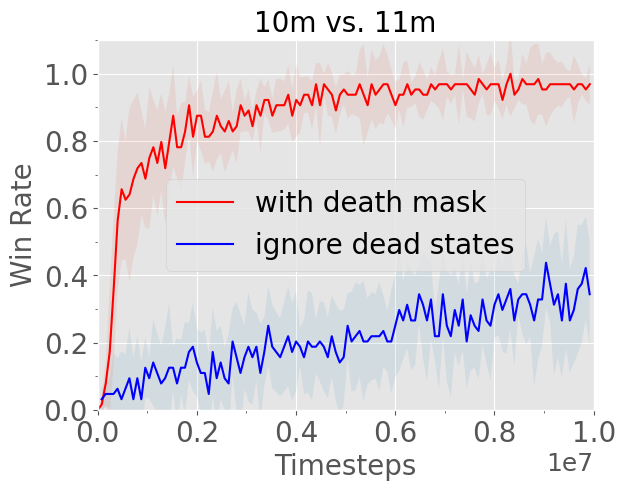}
        \includegraphics[width=0.25\textwidth]{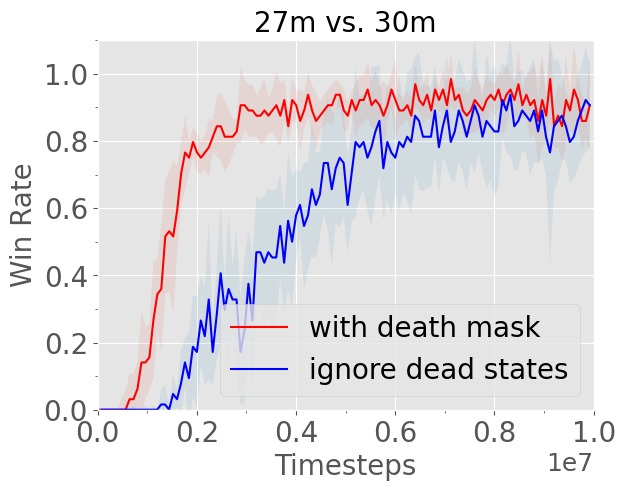}
        \includegraphics[width=0.25\textwidth]{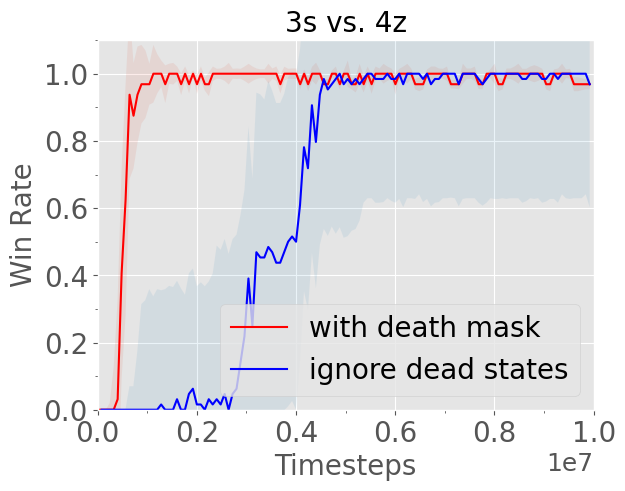}
        \includegraphics[width=0.25\textwidth]{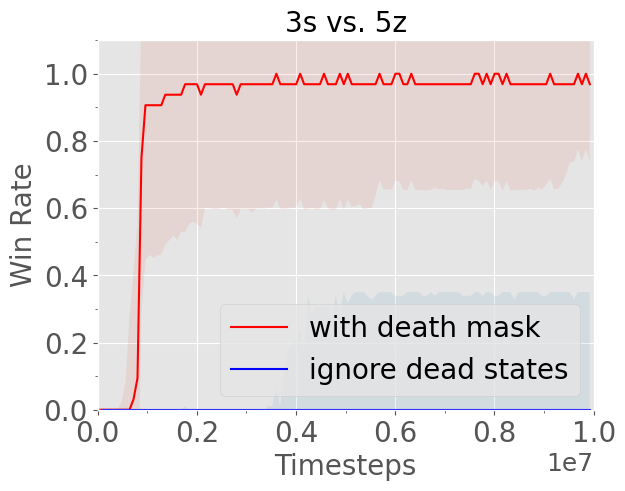}
    	}
    }
    \subfigure
	{\centering
        \includegraphics[width=0.25\textwidth]{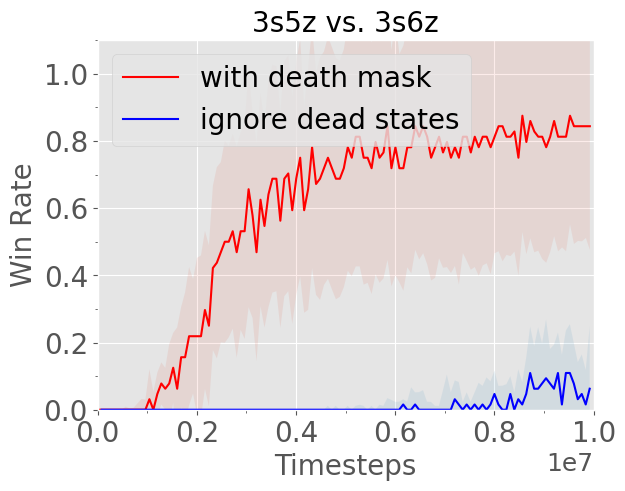}
        \includegraphics[width=0.25\textwidth]{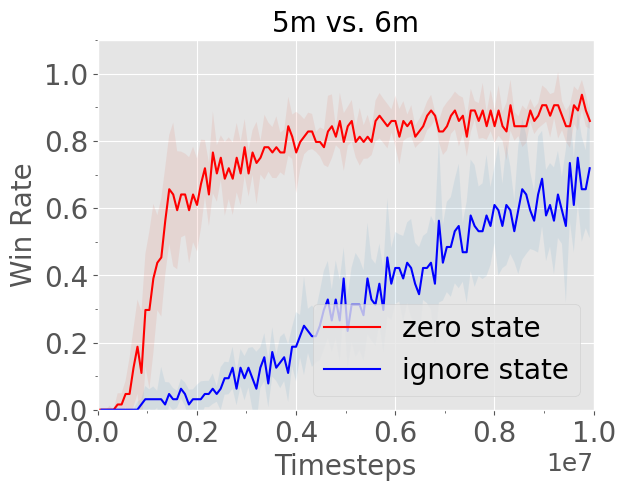}
        \includegraphics[width=0.25\textwidth]{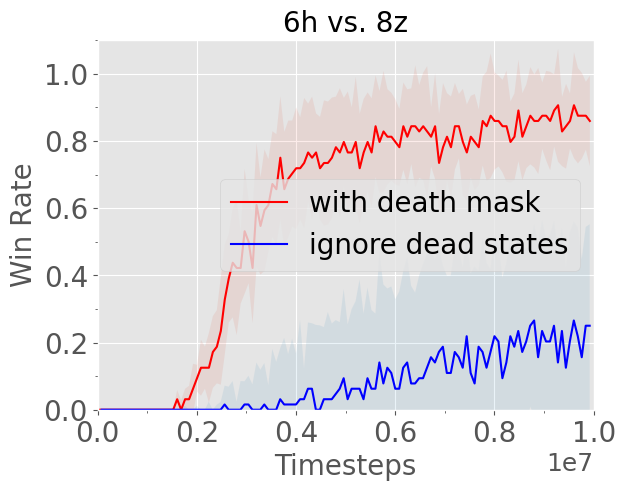}
    	\includegraphics[width=0.25\textwidth]{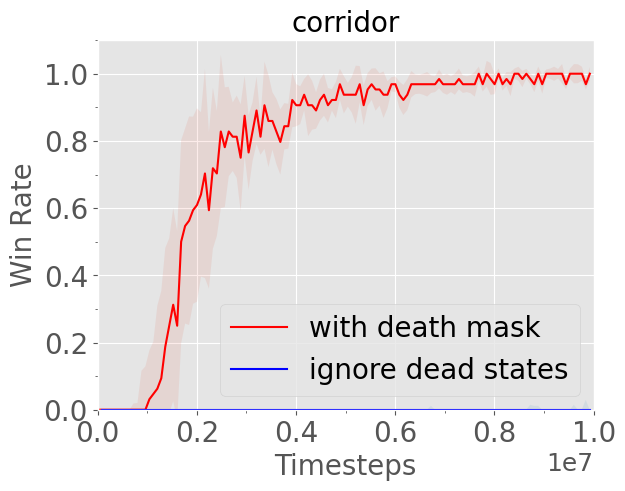}
    }
    \subfigure
	{\centering
    	\includegraphics[width=0.25\textwidth]{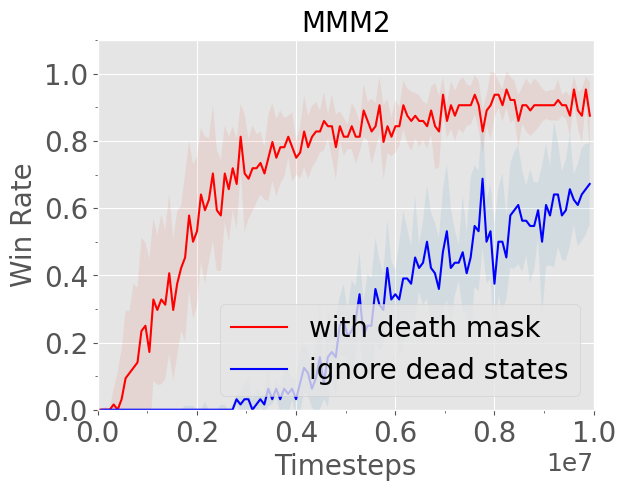}
    }
	\centering \caption{Ablation studies demonstrating the effect of death mask on MAPPO's performance in the SMAC domain.}
\label{fig:app-Ablation-ignore}
\end{figure*}

\begin{figure*}[ht]
    \captionsetup{justification=centering}
	\centering
    \subfigure
	{\centering
        {\includegraphics[width=0.25\textwidth]{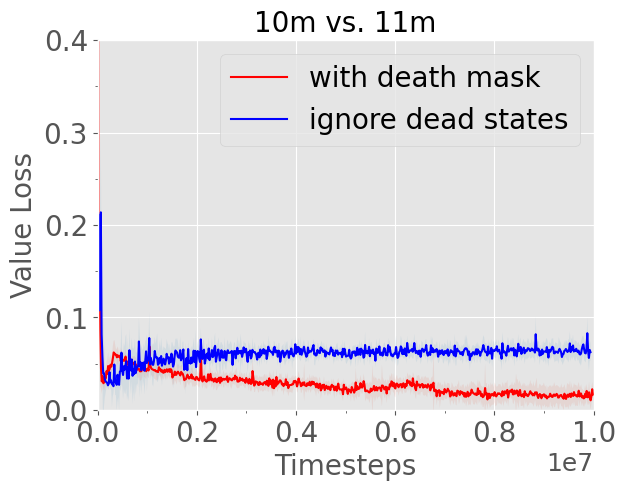}
        \includegraphics[width=0.25\textwidth]{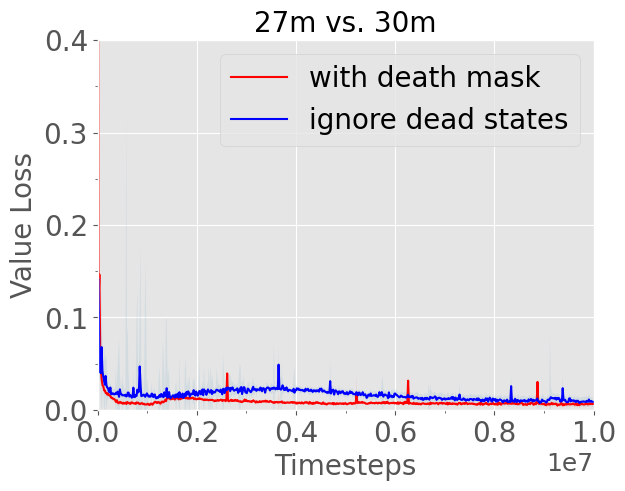}
        \includegraphics[width=0.25\textwidth]{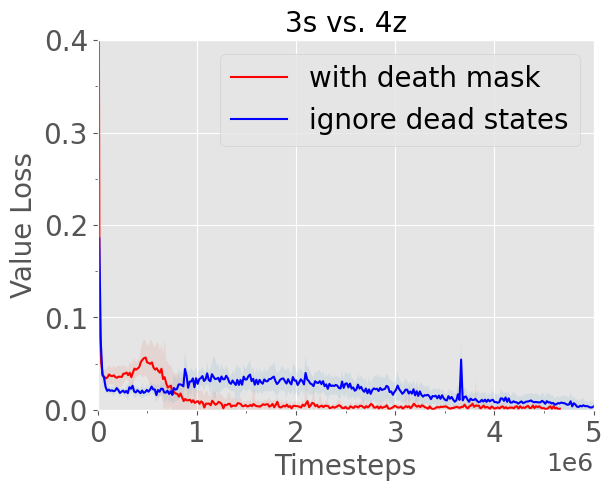}
        \includegraphics[width=0.25\textwidth]{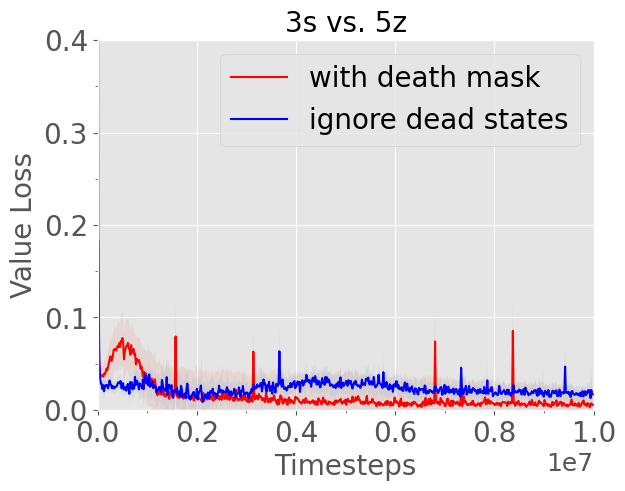}
    	}
    }
    \subfigure
	{\centering
        \includegraphics[width=0.25\textwidth]{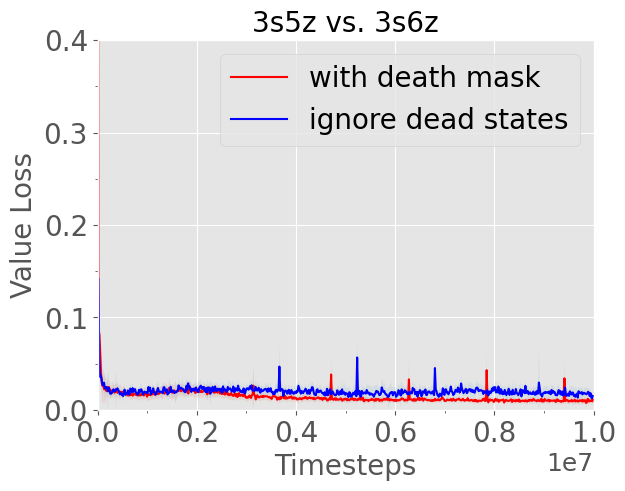}
        \includegraphics[width=0.25\textwidth]{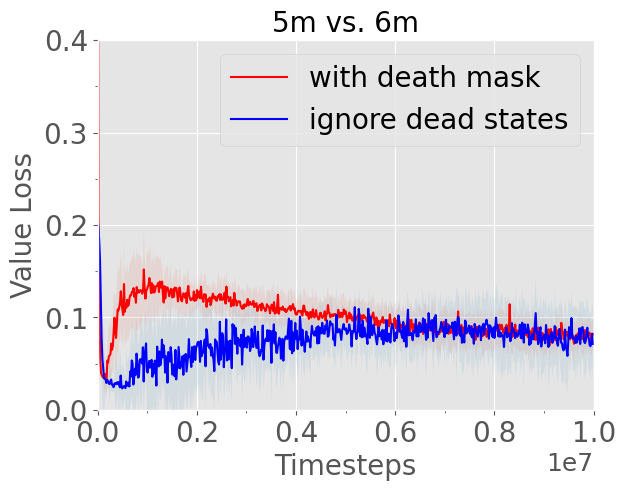}
        \includegraphics[width=0.25\textwidth]{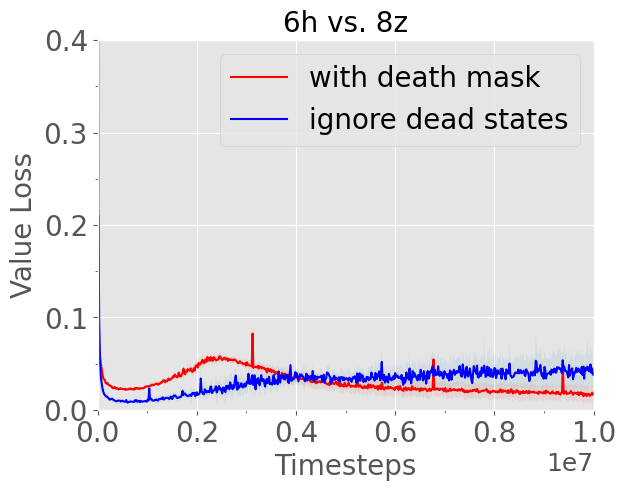}
    	\includegraphics[width=0.25\textwidth]{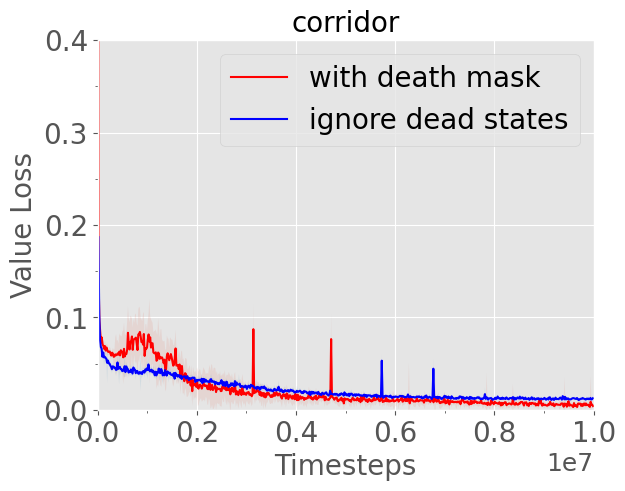}
    }
    \subfigure
	{\centering
    	\includegraphics[width=0.25\textwidth]{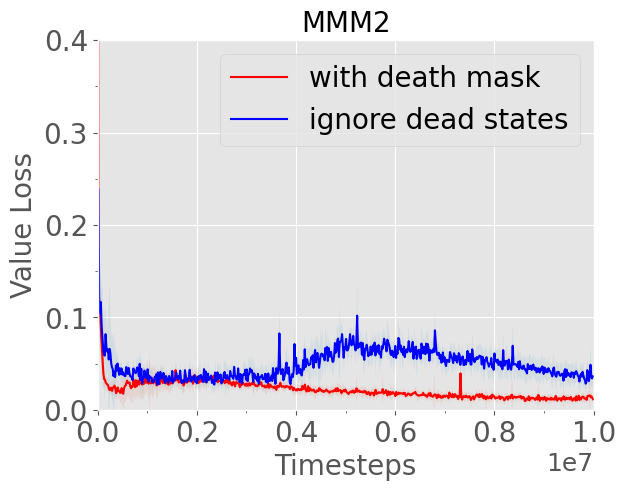}
    }
	\centering \caption{Effect of death mask on MAPPO's value loss in the SMAC domain.}
\label{fig:app-Ablation-ignore_valueloss}
\end{figure*}

\begin{figure*}[ht]
    \captionsetup{justification=centering}
	\centering
    \subfigure
	{\centering
        {\includegraphics[width=0.25\textwidth]{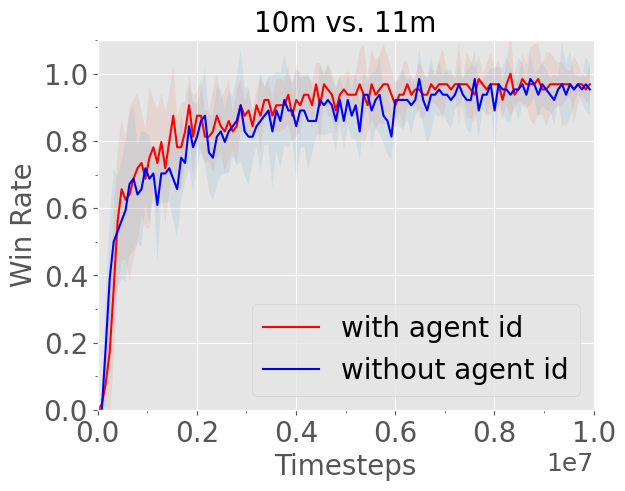}
        \includegraphics[width=0.25\textwidth]{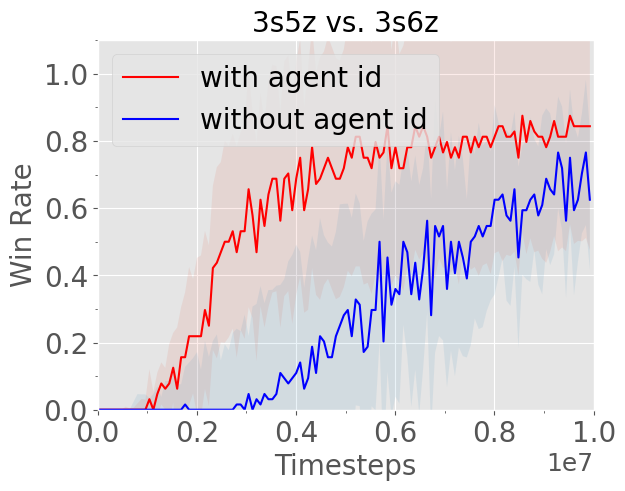}
        \includegraphics[width=0.25\textwidth]{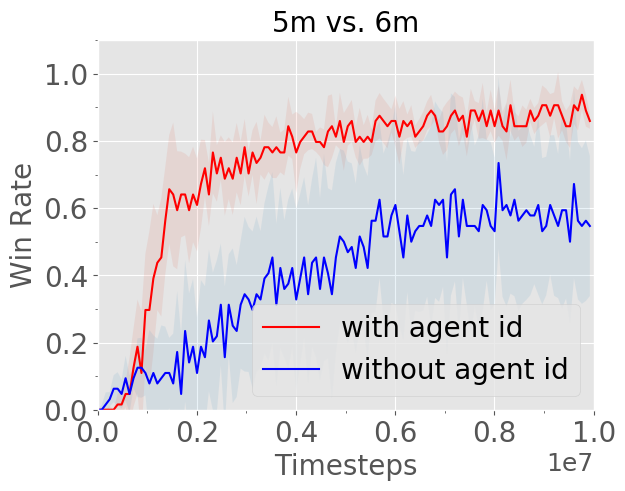}
        \includegraphics[width=0.25\textwidth]{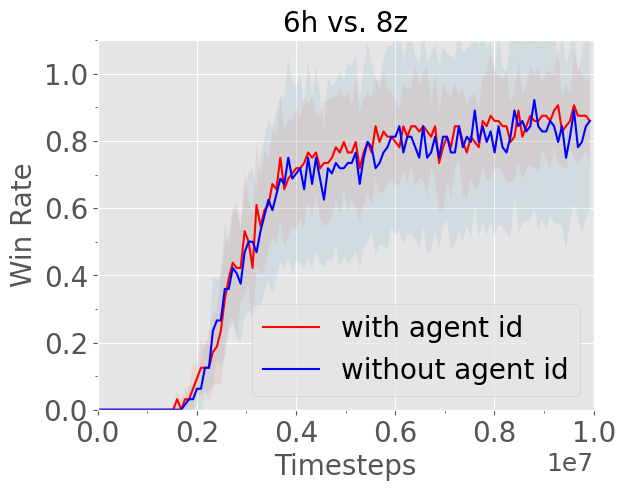}
        }
    }
    \subfigure
	{\centering
        {
        \includegraphics[width=0.25\textwidth]{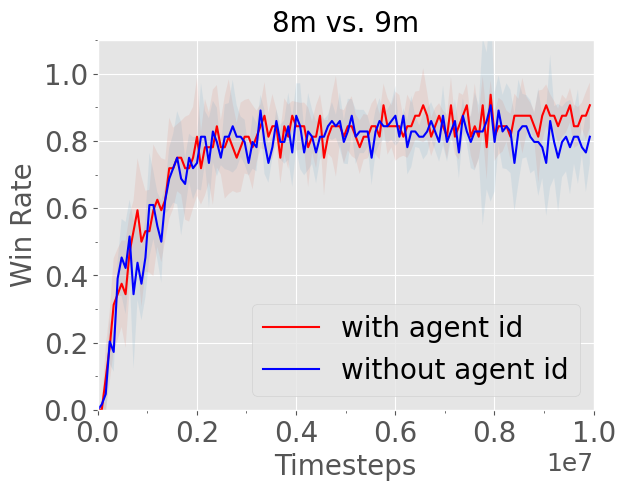}
        \includegraphics[width=0.25\textwidth]{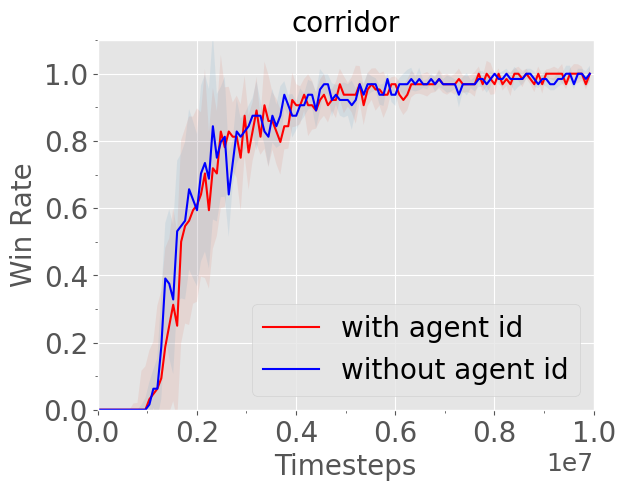}
        \includegraphics[width=0.25\textwidth]{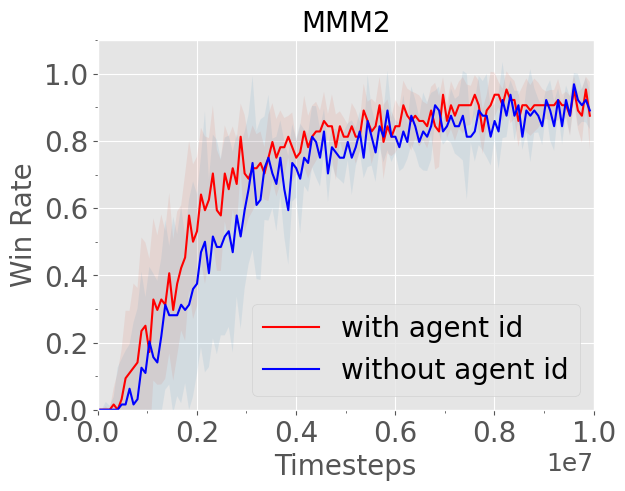}
    	}
    }
	\centering \caption{Ablation studies demonstrating the effect of agent id on MAPPO's performance in the SMAC domain. }
\label{fig:app-Ablation-agentid}
\end{figure*}

We present the learning curves for all ablation studies performed. Fig.~\ref{fig:app-Ablation-popart} demonstrates the impact of value normalization on MAPPO's performance. Fig. \ref{fig:app-Ablation-state} shows the effect of global state information on MAPPO's performance in SMAC. Fig. \ref{fig:app-Ablation-epoch} studies the influence of training epochs on MAPPO's performance. Fig. \ref{fig:app-Ablation-clip} studies the influence of clipping term on MAPPO's performance.
Fig. \ref{fig:app-Ablation-death} and Fig. \ref{fig:app-Ablation-death-new} illustrates the influence of the death mask on MAPPO(FP)'s and MAPPO(AS)'s performance. Similarly, Fig. \ref{fig:app-Ablation-ignore} compares the performance of MAPPO when ignoring states in which an agent is dead when computing GAE to using the death mask when computing the GAE. Fig.~\ref{fig:app-Ablation-ignore_valueloss} illustrates the effect of death mask on MAPPO's value loss in the SMAC domain. Lastly, Fig.~\ref{fig:app-Ablation-agentid} shows the influence of including the agent-id in the agent-specific global state.

\begin{table*}[bt]
\centering
\begin{tabular}{cc}
\toprule
common hyperparameters      & value  \\
\midrule
recurrent data chunk length & 10   \\
gradient clip norm          & 10.0 \\
gae lamda                   & 0.95    \\              
gamma                       & 0.99 \\
value loss                        & huber loss \\
huber delta                 & 10.0   \\
batch size                   & num envs $\times$ buffer length $\times$ num agents  \\
mini batch size           & batch size {/} mini-batch  \\
optimizer                & Adam       \\
optimizer epsilon            & 1e-5       \\
weight decay             & 0          \\
network initialization  & Orthogonal \\
use reward normalization   &True \\
use feature normalization   &True \\

\bottomrule
\end{tabular}
\caption{Common hyperparameters used in MAPPO across all domains.}
\label{tab:MAPPO-common-parameters}
\end{table*}

\begin{table*}[bt]
\centering
\begin{tabular}{cc}
\toprule
common hyperparameters      & value  \\
\midrule
gradient clip norm          & 10.0 \\ 
random episodes             & 5 \\
epsilon                     & $1.0 \rightarrow 0.05$ \\
epsilon anneal time         & 50000 timesteps \\
train interval              & 1 episode\\
gamma                       & 0.99 \\
critic loss                        & mse loss \\
buffer size                 & 5000 episodes   \\
batch size                  & 32 episodes  \\
optimizer                & Adam       \\
optimizer eps            & 1e-5       \\
weight decay             & 0          \\
network initialization  & Orthogonal \\ 
use reward normalization   &True \\
use feature normalization   &True \\

\bottomrule
\end{tabular}
\caption{Common hyperparameters used in QMix and MADDPG across all domains.}
\label{tab:QMixMADDPG-common-parameters}
\end{table*}

\begin{table*}[t!]
\centering
\begin{tabular}{@{}cc@{}}
\toprule
hyperparameters        & value      \\
\midrule
num envs         & \begin{tabular}[c]{@{}c@{}}MAPPO: 128 \end{tabular} \\
buffer length            & MAPPO: 25 \\
num GRU layers            & 1          \\
RNN hidden state dim      & 64         \\
fc layer dim             & 64         \\
num fc            & 2          \\
num fc after             & 1          \\
\bottomrule
\end{tabular}
\caption{Common hyperparameters used in the MPE domain for MAPPO, MADDPG, and QMix.}
\label{tab:typical-MPE}
\end{table*}

\begin{table*}[h!]
\centering
\begin{tabular}{@{}cc@{}}
\toprule
hyperparameters        & value      \\
\midrule
num envs          & \begin{tabular}[c]{@{}c@{}}MAPPO:8\end{tabular}  \\
buffer length            & \begin{tabular}[c]{@{}c@{}}MAPPO: 400 \end{tabular} \\
num GRU layers            & 1          \\
RNN hidden state dim      & 64         \\
fc layer dim             & 64         \\
num fc            & 2          \\
num fc after             & 1          \\
\bottomrule
\end{tabular}
\caption{Common hyperparameters used in the SMAC domain for MAPPO and QMix.}
\label{tab:typical-SMAC}
\end{table*}

\begin{table*}[h!]
\centering
\begin{tabular}{cc}
\toprule
hyperparameters        & value      \\
\midrule
num envs          & 1000 \\
buffer length          & 100 \\
fc layer dim             & 512         \\
num fc            & 2          \\
\bottomrule
\end{tabular}
\caption{Common hyperparameters used in the Hanabi domain for MAPPO.}
\label{tab:typical-Hanabi}
\end{table*}

\begin{table*}[h!]
\centering
\begin{tabular}{@{}cc@{}}
\toprule
hyperparameters        & value      \\
\midrule
parallel envs          & \begin{tabular}[c]{@{}c@{}}MAPPO: 50 \\ QMix: 1\end{tabular}  \\
horizon length            & 199        \\
num GRU layers            & 1          \\
RNN hidden state dim      & 64         \\
fc layer dim             & 64         \\
num fc            & 2          \\
num fc after             & 1          \\
\bottomrule
\end{tabular}
\caption{Common hyperparameters used in the GRF domain for MAPPO and QMix.}
\label{tab:typical-GRF}
\end{table*}

\begin{table*}[]
\centering
\begin{adjustwidth}{-.98in}{-.5in}
\begin{tabular}{ccccccccc}
\toprule
Domains & lr                        & epoch        & mini-batch & activation      & clip & gain         & entropy coef & network     \\ \midrule
MPE     & {[}1e-4,5e-4,7e-4,1e-3{]} & {[}5,10,15,20{]} & {[}1,2,4{]}       & {[}ReLU,Tanh{]} & {[}0.05,0.1,0.15,0.2,0.3,0.5{]} & {[}0.01,1{]} & /  & {[}mlp,rnn{]}               \\ 
SMAC    & {[}1e-4,5e-4,7e-4,1e-3{]} & {[}5,10,15{]}    & {[}1,2,4{]}       & {[}ReLU,Tanh{]} & {[}0.05,0.1,0.15,0.2,0.3,0.5{]} & {[}0.01,1{]} & /   & {[}mlp,rnn{]}               \\
Hanabi  & {[}1e-4,5e-4,7e-4,1e-3{]} & {[}5,10,15{]}    & {[}1,2,4{]}       & {[}ReLU,Tanh{]} &{[}0.05,0.1,0.15,0.2,0.3,0.5{]}& {[}0.01,1{]} & {[}0.01, 0.015{]} & {[}mlp,rnn{]} \\
Football  & {[}1e-4,5e-4,7e-4,1e-3{]} & {[}5,10,15{]}    & {[}1,2,4{]}       & {[}ReLU,Tanh{]} & {[}0.01,1{]} & {[}0.01, 0.015{]} & {[}mlp,rnn{]} \\
\bottomrule
\end{tabular}
\end{adjustwidth}
\caption{Sweeping procedure of MAPPO cross all domains.}
\label{tab:MAPPO-tune-hyper}
\end{table*}

\begin{table*}[]
\centering
\begin{tabular}{cccccc}
\toprule
Domains & lr                        & tau                   & hard interval & activation      & gain         \\\midrule
MPE     & {[}1e-4,5e-4,7e-4,1e-3{]} & {[}0.001,0.005,0.01{]} & {[}100,200,500{]} & {[}ReLU,Tanh{]} & {[}0.01,1{]} \\ 
SMAC    & {[}1e-4,5e-4,7e-4,1e-3{]} & {[}0.001,0.005,0.01{]} & {[}100,200,500{]} & {[}ReLU,Tanh{]} & {[}0.01,1{]} \\ \bottomrule
\end{tabular}
\caption{Sweeping procedure of QMix in the MPE and SMAC domains.}
\label{tab:QMix-tune-hyper}
\end{table*}

\begin{table*}[]
\centering
\begin{tabular}{cccccc}
\toprule
Domains & lr                        & tau                    & activation      & gain  & network       \\\midrule
MPE     & {[}1e-4,5e-4,7e-4,1e-3{]} & {[}0.001,0.005,0.01{]} & {[}ReLU,Tanh{]} & {[}0.01,1{]}  & {[}mlp,rnn{]}\\ \bottomrule
\end{tabular}
\caption{Sweeping procedure of MADDPG in the MPE domain.}
\label{tab:MADDPG-tune-hyper}
\end{table*}

\begin{table*}[]
\centering
\begin{adjustwidth}{-.3in}{-.5in}
\begin{tabular}{cccccccccccc}
\toprule
\multirow{2}{*}{Scenarios} & \multirow{2}{*}{lr} & \multirow{2}{*}{gain} &   \multirow{2}{*}{network} & \multicolumn{3}{c}{MAPPO}                    & \multicolumn{2}{c}{MADDPG} & \multicolumn{3}{c}{QMix}          \\ \cmidrule{5-12} 
 &      &       &     & epoch & mini-batch & activation & tau        & activation     & tau   & hard interval & activation \\ \midrule
Spread                     & 7e-4                & 0.01      & rnn            & 10        & 1          & Tanh          & 0.005      & ReLU           & /     & 100           & ReLU       \\ 
Reference                  & 7e-4                & 0.01        & rnn          & 15        & 1          & ReLU          & 0.005      & ReLU           & 0.005 & /             & ReLU       \\ 
Comm                       & 7e-4                & 0.01     & rnn             & 15        & 1          & Tanh          & 0.005      & ReLU           & 0.005 & /             & ReLU       \\ \bottomrule
\end{tabular}
\end{adjustwidth}
\caption{Adopted hyperparameters used for MAPPO, MADDPG and QMix in the MPE domain.}
\label{tab:diff-MPE}
\end{table*}

\begin{table*}[]
\centering
\begin{adjustwidth}{-.2in}{-.5in}
\begin{tabular}{ccccccccc|cc}
\toprule
\multirow{2}{*}{Maps} & \multirow{2}{*}{lr} & \multirow{2}{*}{activation} & \multicolumn{6}{c}{MAPPO}                                         & \multicolumn{2}{c}{QMix} \\ \cmidrule{4-11} 
                      &                     &                             & epoch & mini-batch & clip & gain & network & stacked frames & hard interval    & gain   \\ \midrule
2m vs. 1z        & 5e-4 & ReLU & 15 & 1 & 0.2  & 0.01 & rnn & 1 & 200 & 0.01 \\
3m               & 5e-4 & ReLU & 15 & 1 & 0.2  & 0.01 & rnn & 1 & 200 & 0.01 \\
2s vs. 1sc       & 5e-4 & ReLU & 15 & 1 & 0.2  & 0.01 & rnn & 1 & 200 & 0.01 \\
3s vs. 3z        & 5e-4 & ReLU & 15 & 1 & 0.2  & 0.01 & rnn & 1 & 200 & 0.01 \\
3s vs. 4z        & 5e-4 & ReLU & 15 & 1 & 0.2  & 0.01 & mlp & 4 & 200 & 0.01 \\
3s vs. 5z        & 5e-4 & ReLU & 15 & 1 & 0.05  & 0.01 & mlp & 4 & 200 & 0.01 \\
2c vs. 64zg      & 5e-4 & ReLU & 5  & 1 & 0.2  & 0.01 & rnn & 1 & 200 & 0.01 \\
so many baneling & 5e-4 & ReLU & 15 & 1 & 0.2  & 0.01 & rnn & 1 & 200 & 0.01 \\
8m               & 5e-4 & ReLU & 15 & 1 & 0.2  & 0.01 & rnn & 1 & 200 & 0.01 \\
MMM              & 5e-4 & ReLU & 15 & 1 & 0.2  & 0.01 & rnn & 1 & 200 & 0.01 \\
1c3s5z           & 5e-4 & ReLU & 15 & 1 & 0.2  & 0.01 & rnn & 1 & 200 & 0.01 \\
8m vs. 9m        & 5e-4 & ReLU & 15 & 1 & 0.05  & 0.01 & rnn & 1 & 200 & 0.01 \\
bane vs. bane    & 5e-4 & ReLU & 15 & 1 & 0.2  & 0.01 & rnn & 1 & 200 & 0.01 \\
25m              & 5e-4 & ReLU & 10 & 1 & 0.2  & 0.01 & rnn & 1 & 200 & 0.01 \\
5m vs. 6m        & 5e-4 & ReLU & 10 & 1 & 0.05 & 0.01 & rnn & 1 & 200 & 0.01 \\
3s5z             & 5e-4 & ReLU & 5  & 1 & 0.2  & 0.01 & rnn & 1 & 200 & 0.01 \\
MMM2             & 5e-4 & ReLU & 5  & 2 & 0.2  & 1    & rnn & 1 & 200 & 0.01 \\
10m vs. 11m      & 5e-4 & ReLU & 10 & 1 & 0.2  & 0.01 & rnn & 1 & 200 & 0.01 \\
3s5z vs. 3s6z    & 5e-4 & ReLU & 5  & 1 & 0.2  & 0.01 & rnn & 1 & 200 & 1    \\
27m vs. 30m      & 5e-4 & ReLU & 5  & 1 & 0.2  & 0.01 & rnn & 1 & 200 & 1    \\
6h vs. 8z        & 5e-4 & ReLU & 5  & 1 & 0.2  & 0.01 & mlp & 1 & 200 & 1    \\
corridor         & 5e-4 & ReLU & 5  & 1 & 0.2  & 0.01 & mlp & 1 & 200 & 1    \\ 
\bottomrule
\end{tabular}
\end{adjustwidth}
\caption{Adopted hyperparameters used for MAPPO and QMix in the SMAC domain.}
\label{tab:diff-SMAC}
\end{table*}

\begin{table*}[]
\centering
\begin{tabular}{cccccccc}
\toprule
\multirow{2}{*}{Tasks} & \multicolumn{7}{c}{MAPPO}                                                                                                                          \\ \cmidrule{2-8} 
                       & lr                                                               & epoch & mini-batch & activation & gain & entropy coef & network \\ \midrule
2-player                 & \begin{tabular}[c]{@{}c@{}}actor:7e-4\\ critic:1e-3\end{tabular} & 15    & 1          & ReLU       & 0.01 & 0.015        & mlp                   \\ \bottomrule
\end{tabular}
\caption{Adopted hyperparameters used for MAPPO in the Hanabi domain.}
\label{tab:diff-Hanabi}
\end{table*}

\begin{table*}[]
\centering
\begin{tabular}{cccccccc|cc}
\toprule
\multirow{2}{*}{Scenarios} & \multirow{2}{*}{lr} & \multirow{2}{*}{activation} & \multirow{2}{*}{buffer length} & \multicolumn{4}{c}{MAPPO}                             & \multicolumn{2}{c}{QMix} \\ \cmidrule{5-10}
                      &                     &             &                & epoch & mini-batch & gain & network  & hard interval   & gain   \\
\midrule
3v.1             & 5e-4                & ReLU        & 200                & 15    & 2          & 0.01 & rnn                   & 200             & 0.01   \\
Corner                    & 5e-4                & ReLU    & 1000                    & 15    & 2          & 0.01 & rnn                   & 200             & 0.01   \\
CA(easy)            & 5e-4                & ReLU     & 200                    & 15    & 2          & 0.01 & rnn                 & 200             & 0.01   \\
CA(hard)            & 5e-4                & ReLU      & 1000                  & 15    & 2          & 0.01 & rnn                  & 200             & 0.01   \\
PS             & 5e-4                & ReLU        & 200                 & 15    & 2          & 0.01 & rnn                  & 200             & 0.01   \\
RPS             & 5e-4                & ReLU       & 200                  & 15    & 2          & 0.01 & rnn                 & 200             & 0.01   \\
\bottomrule
\end{tabular}
\caption{Adopted hyperparameters used for MAPPO and QMix in the Football domain.}
\label{tab:diff-Football}
\end{table*}

